%% file: main.tex
\documentclass[conference]{IEEEtran}
\IEEEoverridecommandlockouts
\pdfoutput=1 
\usepackage{multirow}
\usepackage{amsmath,amssymb,amsfonts,amsbsy}
\usepackage{hhline}
\usepackage{graphicx}
\usepackage{float}
\usepackage{multicol}
\usepackage{textcomp}
\usepackage{booktabs}
\usepackage{float}
\usepackage{enumitem}
\usepackage[font=small,labelfont=bf]{caption}
\usepackage{subcaption}
\usepackage{xcolor}
\usepackage{graphicx}
\usepackage[margin=1in]{geometry}
\usepackage[utf8]{inputenc}
\usepackage{balance}
\usepackage{hyperref}

\DeclareMathOperator*{\argmax}{arg\,max}
\newcommand{\floor}[1]{\lfloor #1 \rfloor}

\DeclareMathAlphabet{\mathpzc}{OT1}{pzc}{m}{it}

\renewcommand{\footnoterule}{%
	\kern -3pt
	\hrule
	\kern 2pt
}
\renewcommand{\figurename}{Fig.}
\renewcommand{\tablename}{Table}
\renewcommand{\baselinestretch}{0.832}
\usepackage{microtype}

\usepackage[mincitenames=1,sorting=none,backend=biber,doi=false,isbn=false,url=false]{biblatex} 
\def\BibTeX{{\rm B\kern-.05em{\sc i\kern-.025em b}\kern-.08em
		T\kern-.1667em\lower.7ex\hbox{E}\kern-.125emX}}

\title{Logic Synthesis Meets Machine Learning:\\ Trading Exactness for Generalization
\thanks{This work was supported in part by the Semiconductor Research Corporation under Contract 2867.001.}
}
\author{\IEEEauthorblockN{
Shubham Rai\textsuperscript{f,6,$\dagger$},
Walter Lau Neto\textsuperscript{n,10,$\dagger$},
Yukio Miyasaka\textsuperscript{o,1},
Xinpei Zhang\textsuperscript{a,1},
Mingfei Yu\textsuperscript{a,1},
Qingyang Yi\textsuperscript{a,1},\\ 
Masahiro Fujita\textsuperscript{a,1},
Guilherme B. Manske\textsuperscript{b,2},
Matheus F. Pontes\textsuperscript{b,2},
Leomar S. da Rosa\ Junior\textsuperscript{b,2},\\
Marilton S. de Aguiar\textsuperscript{b,2},
Paulo F. Butzen\textsuperscript{e,2},
Po-Chun Chien\textsuperscript{c,3},
Yu-Shan Huang\textsuperscript{c,3},
Hoa-Ren Wang\textsuperscript{c,3},\\
Jie-Hong R. Jiang\textsuperscript{c,3},
Jiaqi Gu\textsuperscript{d,4}, 
Zheng Zhao\textsuperscript{d,4},
Zixuan Jiang\textsuperscript{d,4},
David Z. Pan\textsuperscript{d,4},
Brunno A. de Abreu\textsuperscript{e,5,9},\\
Isac de Souza Campos\textsuperscript{m,5,9},
Augusto Berndt\textsuperscript{m,5,9},
Cristina Meinhardt\textsuperscript{m,5,9},
Jonata T. Carvalho\textsuperscript{m,5,9},\\
Mateus Grellert\textsuperscript{m,5,9},
Sergio Bampi\textsuperscript{e,5},
Aditya Lohana\textsuperscript{f,6},
Akash Kumar\textsuperscript{f,6},
Wei Zeng\textsuperscript{j,7},
Azadeh Davoodi\textsuperscript{j,7},\\
Rasit O. Topaloglu\textsuperscript{k,7},
Yuan Zhou\textsuperscript{l,8},
Jordan Dotzel\textsuperscript{l,8},
Yichi Zhang\textsuperscript{l,8},
Hanyu Wang\textsuperscript{l,8},
Zhiru Zhang\textsuperscript{l,8},\\
Valerio Tenace\textsuperscript{n,10},
Pierre-Emmanuel Gaillardon\textsuperscript{n,10},
Alan Mishchenko\textsuperscript{o,$\dagger$}, and
Satrajit Chatterjee\textsuperscript{p,$\dagger$}
}
\\[-0.0cm]
\IEEEauthorblockA{
\textsuperscript{a}University of Tokyo, Japan,  
\textsuperscript{b}Universidade Federal de Pelotas, Brazil, 
\textsuperscript{c}National Taiwan University,\\ Taiwan, 
\textsuperscript{d}University of Texas at Austin, USA, 
\textsuperscript{e}Universidade Federal do Rio Grande do Sul, Brazil,   \\
\textsuperscript{f}Technische Universitaet Dresden, Germany,  
\textsuperscript{j}University of Wisconsin--Madison, USA, \textsuperscript{k}IBM, USA, \\
\textsuperscript{l}Cornell University, USA,  
\textsuperscript{m}Universidade Federal de Santa Catarina, Brazil,
\textsuperscript{n}University of Utah, USA,\\ 
\textsuperscript{o}UC Berkeley, USA,~\textsuperscript{p}Google AI, USA\\
\IEEEauthorblockN{\small The alphabetic characters in the superscript represent the affiliations while the digits represent the team numbers}
\IEEEauthorblockN{\small \textsuperscript{$\dagger$}Equal contribution. Email: shubham.rai@tu-dresden.de, walter.launeto@utah.edu, alanmi@berkeley.edu,  schatter@google.com}}

\vspace{-1em}

}

\begin{document}

\maketitle

\begin{abstract}
\emph{Logic synthesis} is a fundamental step in hardware design whose goal is to find structural representations of Boolean functions while minimizing delay and area. If the function is completely-specified, the implementation accurately represents the function. If the function is incompletely-specified, the implementation has to be true only on the care set. While most of the algorithms in logic synthesis rely on SAT and Boolean methods to exactly implement the care set, we investigate learning in logic synthesis, attempting to trade exactness for generalization. This work is directly related to \emph{machine learning} where the care set is the training set and the implementation is expected to generalize on a validation set. We present learning incompletely-specified functions based on the results of a competition conducted at IWLS 2020. The goal of the competition was to implement 100 functions given by a set of care minterms for training, while testing the implementation using a set of validation minterms sampled from the same function. We make this benchmark suite available and offer a detailed comparative analysis of the different approaches to learning.

\end{abstract}

\section{Introduction}
\label{intro}
\input{introduction.tex}
\section{Background and Preliminaries}
\label{background}
\input{background.tex}

\section{Benchmarks}
\label{benchmarks}
\input{benchmarks}


\input{approach_small.tex}

\section{Results}
\label{results}
\input{results}

\balance
\section{Conclusion}
\label{conc}
\input{conclusion.tex}

\renewcommand{\bibfont}{\footnotesize}
\printbibliography

\begin{refsection}
\section*{\LARGE Appendix}
\label{Appendix }
\setcounter{section}{0}
\input{approach}

\printbibliography
\end{refsection}

\end{document}

%% file: introduction.tex


Logic synthesis is a key ingredient in modern electronic design automation flows. A central problem in logic synthesis is the following: 
Given a Boolean function $f : {\mathbb B}^n \to {\mathbb B}$ (where ${\mathbb B}$ denotes the set $\{0, 1\}$), construct a logic circuit that implements $f$ with the minimum number of logic gates.
The function $f$ may be completely specified, i.e., we are given $f(x)$ for all $x \in {\mathbb B}^n$, or it may be incompletely specified, i.e., we are only given $f(x)$ for a subset of ${\mathbb B}^n$ called the {\em careset}. 
An incompletely specified function provides more flexibility for optimizing the circuit since the values produced by the circuit outside the careset are not of interest.

Recently, machine learning has emerged as a key enabling technology for a variety of breakthroughs in artificial intelligence. A central problem in machine learning is that of supervised learning: Given a class $\mathcal{H}$ of functions from a domain $X$ to a co-domain $Y$, find a member $h \in \mathcal{H}$ that best fits a given set of training examples of the form $(x, y) \in X \times Y$.
The quality of the fit is judged by how well $h$ generalizes, i.e., how well $h$ fits examples that were {\em not} seen during training.

Thus, logic synthesis and machine learning are closely related. 
Supervised machine learning can be seen as logic synthesis of an incompletely specified function with a different constraint (or objective): the circuit must also generalize well outside the careset (i.e., to the test set) possibly at the expense of reduced accuracy on the careset (i.e., on the training set). Conversely, logic synthesis may be seen as a machine learning problem where in addition to generalization, we care about finding an element of $\mathcal{H}$ that has small size, and the sets $X$ and $Y$ are not smooth but discrete.

To explore this connection between the two fields, the two last authors of this paper organized a programming contest at the 2020 International Workshop in Logic Synthesis.
The goal of this contest was to come up with an algorithm to synthesize a small circuit for a Boolean function $f : {\mathbb B}^n \to {\mathbb B}$ learnt from a training set of examples. Each example $(x, y)$ in the training set is an input-output pair, i.e., $x \in {\mathbb B}^n$ and $y \in {\mathbb B}$.
The training set was chosen at random from the
$2^n$ possible inputs of the function (and in most cases was much smaller than $2^n$).
The quality of the solution was evaluated by measuring accuracy on a test set not provided to the participants. 

The synthesized circuit for $f$ had to be in the form of an \emph{And-Inverter Graph} (AIG)~\cite{Chatterjee07, aiger} with no more than 5000 nodes. An AIG is a standard data structure used in logic synthesis to represent Boolean functions where a node corresponds to a 2-input And gate and edges represent direct or inverted connections.
Since an AIG can represent any Boolean function, in this problem $\mathcal{H}$ is the full set of Boolean functions on $n$ variables.
 
To evaluate the algorithms proposed by the participants, we created a set of 100 benchmarks drawn from a mix of standard problems in logic synthesis such as synthesis of arithmetic circuits and random logic from standard logic synthesis benchmarks. We also included some tasks from standard machine learning  benchmarks. For each benchmark the participants were provided with the training set (which was sub-divided into a training set proper of 6400 examples and a validation set of another 6400 examples though the participants were free to use these subsets as they saw fit), and the circuits returned by their algorithms were evaluated on the corresponding test set (again with 6400 examples) that was kept private until the competition was over. The training, validation and test sets were created in the PLA format~\cite{PLA}. 
The score assigned to each participant was the average test accuracy over all the benchmarks with possible ties being broken by the circuit size.

Ten teams spanning 6 countries took part in the contest. They explored many different techniques to solve this problem. 
In this paper we present short overviews of the techniques used by the different teams (the superscript for an author indicates their team number), as well a comparative analysis of these techniques.
The following are our main findings from the analysis:
\begin{itemize}
    \item No one technique dominated across all the benchmarks, and most teams including the winning team used an ensemble of techniques.
    \item Random forests (and decision trees) were very popular and form a strong baseline, and may be a useful technique for approximate logic synthesis.
    \item Sacrificing a little accuracy allows for a significant reduction in the size of the circuit.
\end{itemize}
These findings suggest an interesting direction for future work: {\em Can machine learning algorithms be used for approximate logic synthesis to greatly reduce power and area when exactness is not needed?}

Finally, we believe that the set of benchmarks used in this contest along with the solutions provided by the participants (based on the methods described in this paper) provide an interesting framework to evaluate further advances in this area. 
To that end we are making these available at \url{https://github.com/iwls2020-lsml-contest/}. 

%% file: background.tex
We review briefly the more popular techniques used.

\emph{\textbf{\textit{Sum-of-Products}}} (SOP), or disjunctive normal form, is a two-level logic representation commonly used in logic synthesis. Minimizing the SOP representation of an incompletely specified Boolean function is a well-studied problem with a number of exact approaches \cite{COUDERT199497, coudertcovering, negativethinking} as well as heuristics \cite{brayton1984logic,Rudell1987,McGeer1993,Hlavicka2001} with ESPRESSO~\cite{brayton1984logic} being the most popular. 

\emph{\textbf{\textit{Decision Trees (DT) and Random Forests (RF)}}} are very popular techniques in machine learning and they were used by many of the teams. In the contest scope, the decision trees were applied as a classification tree, where the internal nodes were associated to the function input variables, and terminal nodes classify the function as 1 or 0, given the association of internal nodes. Thus, each internal node has two outgoing-edges: a \textit{true} edge if the variable value exceeds a threshold value, and a \textit{false} value otherwise. The threshold value is defined during training. Hence, each internal node can be seen as a multiplexer, with the selector given by the threshold value. 
Random forests are composed by multiple decision trees, where each tree is trained over a distinct feature, so that trees are not very similar. The output is given by the combination of individual predictions. 

\emph{\textbf{\textit{Look-up Table (LUT) Network}}} is a network of randomly connected \textit{k-}input LUTs, where each \textit{k}-input LUT can implement any function with up to \textit{k} variables. LUT networks were first employed in a theoretical study to understand if pure memorization (i.e., fitting without any explicit search or optimization) could lead to generalization~\cite{chatterjee2018learning}.

%% file: benchmarks.tex

\begin{table}[t]
\centering
\caption{An overview of different types of functions in the benchmark set. They are selected from three domains: {\color[HTML]{6434FC} {Arithmetic}}, {\color[HTML]{FE0000}{Random Logic}}, and {\color[HTML]{3166FF}{Machine Learning}}.}
\label{table:benchmarks}
 	\def\arraystretch{1.5}
 	\resizebox{\columnwidth}{!}{
\begin{tabular}{|c|l|}
\hline
{\color[HTML]{6434FC} 00-09} & {\color[HTML]{6434FC} 2 MSBs of $k$-bit adders for $k \in \{16, 32, 64, 128, 256\}$}                             \\ \hline
{\color[HTML]{6434FC} 10-19}  & {\color[HTML]{6434FC} MSB of $k$-bit dividers and remainder circuits for $k \in \{16, 32, 64, 128, 256\}$}       \\ \hline
{\color[HTML]{6434FC} 20-29}  & {\color[HTML]{6434FC} MSB and middle bit of $k$-bit multipliers for $k \in \{8, 16, 32, 64, 128\}$}              \\ \hline
{\color[HTML]{6434FC} 30-39}  & {\color[HTML]{6434FC} $k$-bit comparators for $k \in \{10, 20, \dots, 100\}$}                                     \\ \hline
{\color[HTML]{6434FC} 40-49}  & {\color[HTML]{6434FC} LSB and middle bit of $k$-bit square-rooters with $k \in \{16, 32, 64, 128, 256\}$}        \\ \hline
{\color[HTML]{FE0000} 50-59}  & {\color[HTML]{FE0000} 10 outputs of PicoJava design with 16-200 inputs and roughly balanced onset \& offset} \\ \hline
{\color[HTML]{FE0000} 60-69}  & {\color[HTML]{FE0000} 10 outputs of MCNC i10 design with 16-200 inputs and roughly balanced onset \& offset}  \\ \hline
{\color[HTML]{FE0000} 70-79}  & {\color[HTML]{FE0000} 5 other outputs from MCNC benchmarks + 5 symmetric functions of 16 inputs}            \\ \hline
{\color[HTML]{3166FF} 80-89}  & {\color[HTML]{3166FF} 10 binary classification problems from MNIST group comparisons}                       \\ \hline
{\color[HTML]{3166FF} 90-99}  & {\color[HTML]{3166FF} 10 binary classification problems from CIFAR-10 group comparisons}                    \\ \hline
\end{tabular}}
\end{table}

The set of 100 benchmarks used in the contest can be broadly divided into 10 categories, each with 10 test-cases. The summary of categories is shown in~\tablename{~\ref{table:benchmarks}}. 
For example, the first 10 test-cases are created by considering the two most-significant bits (MSBs) of $k$-input adders for $k \in \{16, 32, 64, 128, 256\}$. 

Test-cases ex60 through ex69 were derived from MCNC benchmark~\cite{yang1991logic} i10 by extracting outputs 91, 128, 150, 159, 161, 163, 179, 182, 187, and 209 (zero-based indexing). For example, ex60 was derived using the ABC command line: \textit{\&read} i10.aig; \textit{\&cone} -O 91. 

Five test-cases ex70 through ex74 were similarly derived from MCNC benchmarks \textit{cordic} (both outputs), \textit{too\_large} (zero-based output 2), \textit{t481}, and \textit{parity}.  

Five 16-input symmetric functions used in ex75 through ex79 have the following signatures:\\
\phantom{and} 00000000111111111, 11111100000111111,\\
\phantom{and} 00011110001111000, 00001110101110000, and\\
\phantom{and} 00000011111000000. \\
They were generated by ABC using command \textit{symfun}~$\langle$\textit{signature}$\rangle$.

\tablename{~\ref{table:mlbench}} shows the rules used to generate the last 20 benchmarks.  Each of the 10 rows of the table contains two groups of labels, which were compared to generate one test-case. Group A results in value 0 at the output, while Group B results in value 1. The same groups were used for MNIST~\cite{lecun-mnisthandwrittendigit-2010} and CIFAR-10~\cite{krizhevsky2009learning}.  For example, benchmark ex81 compares odd and even labels in MNIST, while benchmark ex91 compares the same labels in CIFAR-10.

In generating the benchmarks, the goal was to fulfill the following requirements:
(1) Create problems, which are non-trivial to solve.
(2) Consider practical functions, such as arithmetic logic and symmetric functions, extract logic cones from the available benchmarks, and derive binary classification problems from the MNIST and CIFAR-10 machine learning challenges.
(3) Limit the number of AIG nodes in the solution to 5000 to prevent the participants from generating large AIGs and rather concentrate on algorithmic improvements aiming at high solution quality using fewer nodes.

There was also an effort to discourage the participants from developing strategies for reverse-engineering the test-cases based on their functionality, for example, detecting that some test-cases are outputs of arithmetic circuits, such as adders or multipliers. Instead, the participants were encouraged to look for algorithmic solutions to handle arbitrary functions and produce consistently good solutions for every one independently of its origin.

\begin{table}
\centering
\caption{Group comparisons for MNIST and CIFAR10}
\begin{tabular}{lrrrr}
\toprule
ex & Group A & Group B \\
\midrule
0  & 0-4 & 5-9    \\
1  & odd & even   \\
2  & 0-2 & 3-5    \\
3  & 01  & 23     \\
4  & 45  & 67     \\
5  & 67  & 89     \\
6  & 17  & 38     \\
7  & 09  & 38     \\
8  & 13  & 78     \\
9  & 03  & 89     \\ \bottomrule
\end{tabular}
\vspace{-2em}
\label{table:mlbench}
\end{table}

%% file: approach_small.tex

\section{Overview of the Various Approaches}
\textbf{Team 1's}
\label{s_team1}
solution is to take the best one among ESPRESSO, LUT network, RF, and pre-defined standard function matching (with some arithmetic functions). If the AIG size exceeds the limit, a simple approximation method is applied to the AIG.

ESPRESSO is used with an option to finish optimization after the first irredundant operation. LUT network has some parameters: the number of levels, the number of LUTs in each level, and the size of each LUT. These parameters are incremented like a beam search as long as the accuracy is improved. The number of estimators in random forest is explored from 4 to 16.

A simple approximation method is used if the number of AIG nodes is more than 5000. The AIG is simulated with thousands of random input patterns, and the node which most frequently outputs 0 is replaced by constant-0 while taking the negation (replacing with constant-1) into account. This is repeated until the AIG size meets the condition. The nodes near the outputs are excluded from the candidates by setting a threshold on levels. The threshold is explored through try and error. It was observed that the accuracy drops 5\% when reducing 3000-5000 nodes.


\textbf{Team 2's} solution uses J48 and PART AI classifiers to learn the unknown 
Boolean function from a single training set that combines the training and validation sets. The algorithm first transforms the PLA file in an ARFF (Attribute-Relation File Format) description to handle the WEKA tool~\cite{weka}. We used the WEKA tool to run five different configurations to the J48 classifier and five configurations to the PART classifier, varying the confidence factor. The J48 classifier creates a decision tree that the developed software converts in a PLA file.  In the sequence, the ABC tool transforms the PLA file into an AIG file. The PART classifier creates a set of rules that the developed software converts in an AAG file. After, the AIGER transforms the AAG file into an AIG file
to decide the best configuration for each classifier. Also, we use the minimum number of objects to determine the best classifier. Finally, the ABC tool checks the size of the generated AIGs to match the contest requirements.

\label{s_team2}

\textbf{Team 3's}
\label{s_team3}
solution consists of decision tree based and neural network (NN) based methods.
For each benchmark, multiple models are trained and 3 are selected for ensemble.
%
For the DT-based method, the fringe feature extraction process proposed in \cite{Pagallo:1990, Arlindo:1993} is adopted.
The DT is trained and modified for multiple iterations.
In each iteration, the patterns near the fringes (leave nodes) of the DT are identified as the composite features of 2 decision variables.
These newly detected features are then added to the list of decision variables for the DT training in the next iteration.
The procedure terminates when there are no new features found or the number of the extracted features exceeds the preset limit.

For the NN-based method, a 3-layer network is employed, where each layer is fully-connected and uses {\it sigmoid} as the activation function.
As the synthesized circuit size of a typical NN could be quite large, the connection pruning technique proposed in \cite{Han:2015} is adopted 
to meet the stringent size restriction.
The NN is pruned until the number of fanins of each neuron is at most 12.
Each neuron is then synthesized into a LUT by rounding 
its activation~\cite{chatterjee2018learning}.
The overall dataset, training and validation set combined, for each benchmark is re-divided into 3 partitions before training.
Two partitions are selected as the new training set, and the remaining one as the new validation set, resulting in 3 different grouping configurations.
Under each configuration, multiple models are trained with different methods and hyper-parameters, and the one with the highest validation accuracy is chosen for ensemble.

\textbf{Team 4's}
\label{s_team4}
solution is based on multi-level ensemble-based feature selection, recommendation-network-based model training, subspace-expansion-based prediction, and accuracy-node joint exploration during synthesis.

Given the high sparsity in the high-dimensional boolean space, a multi-level feature importance ranking is adopted to reduce the learning space.
Level 1: a 100-\texttt{ExtraTree} based \texttt{AdaBoost}~\cite{NN_JCSS1997_Freund} ensemble classifier is used with 10-repeat permutation importance~\cite{NN_ML2001_Breiman} ranking to select the top-$k$ important features, where $k\in[10,16]$.
Level 2: a 100-\texttt{ExtraTree} based \texttt{AdaBoost} classifier and an \texttt{XGB} classifier with 200 trees are used with stratified 10-fold cross-validation to select top-$k$ important features, where $k$ ranges from 10 to 16, given the 5,000 node constraints.

Based on the above 14 groups of selected features, 14 state-of-the-art recommendation models, Adaptive Factorization Network (\texttt{AFN})~\cite{NN_AAAI2020_Cheng}, are independently learned as DNN-based boolean function approximators.
A 128-dimensional logarithmic neural network is used to learn sparse boolean feature interaction, and a 4-layer MLP is used to combine the formed cross features with overfitting being handled by fine-tuned dropout.
After training, a $k$-feature trained model will predict the output for $2^k$ input combinations to expand the full $k$-dimensional hypercube, where other pruned features are set to DON'T CARE type in the predicted \texttt{.pla} file to allow enough smoothness in the Boolean hypercube.
Such a subspace expansion technique can fully-leverage the prediction capability of our model to maximize the accuracy on the validation/test dataset while constraining the maximum number of product terms for node minimization during synthesis.

\textbf{Team 5's}
\label{s_team5}
solution explores the use of DTs and RFs, along with NNs, to learn the required Boolean functions. DTs/RFs are easy to convert into SOP expressions. To evaluate this proposal, the implementation obtains the models using the Scikit-learn Python library \cite{sklearn}. The solution is chosen from simulations using {\it DecisionTreeClassifier} for the DTs, and an ensemble of {\it DecisionTreeClassifier } for the RFs -- the {\it RandomForestClassifier} structure would be inconvenient, considering the 5000-gate limit, given that it employs a weighted average of each tree.

The simulations are performed using different tree depths and feature selection methods ({\it SelectKBest} and {\it SelectPercentile}). NNs are also employed to enhance our exploration capabilities, using the {\it MLPClassifier} structure. Given that SOPs cannot be directly obtained from the output of the NN employed, the NN is used as a feature selection method to obtain the importance of each input based on their weight values. With a small sub-set of weights obtained from this method, the proposed solution performs a small exhaustive search by applying combinations of functions on the four features with the highest importance, considering OR, XOR, AND, and NOT functions. The SOP with the highest accuracy (respecting the 5001-gate limit) out of the DTs/RFs and NNs tested was chosen to be converted to an AIG file. The data sets were split into an 80\%-20\% ratio, preserving the original data set's target distribution. The simulations were run using half of the newly obtained training set (40\%) and the whole training set to increase our exploration.

\textbf{Team 6's}
\label{s_team6}
solution learns the unknown Boolean function using the method as mentioned in~\cite{chatterjee2018learning}. 
In order to construct the LUT network, we use the minterms as input features to construct layers of LUTs with connections starting from the input layer. We then carry out two schemes of connections between the layers: `random set of input' and `unique but random set of inputs'. By `random set of inputs', we imply that we just randomly select the outputs of preceding layer and feed it to the next layer. This is the default flow. By `unique but random set of inputs', we mean that we ensure that all outputs from a preceding layer is used before duplication of connection. 

We carry out experiments with four hyper parameters to achieve accuracy-- number of inputs per LUT, number of LUTS per layers, selection of connecting edges from the preceding layer to the next layer and the depth (number of LUT layers) of the model. We experiment with varying number of inputs for each LUT in order to get the maximum accuracy. We notice from our experiments that 4-input LUTs returns the best average numbers across the benchmark suite. 

Once the network is created, we convert the network into an SOP form using \emph{sympy} package in python. This is done from reverse topological order starting from the outputs back to the inputs. Using the SOP form, we generate the verilog file which is then used with ABC to calculate the accuracy. 

\textbf{Team 7's}
\label{s_team7}
solution is a mix of conventional ML and pre-defined standard function matching. If a training set matches a pre-defined standard function, a custom AIG of the identified function is written out. Otherwise, an ML model is trained and translated to an AIG.

Team 7 adopts tree-based ML models for the straightforward conversion from tree nodes to SOP terms. The model is either a decision tree with unlimited depth, or an extreme gradient boosting (XGBoost) of 125 trees with a maximum depth of five, depending on the results of a 10-fold cross validation on training data.

With the learned model, all underlying tree leaves are converted to SOP terms, which are minimized and compiled to AIGs with ESPRESSO and \texttt{ABC}, respectively. If the model is a decision tree, the converted AIG is final. If the model is XGBoost, the value of each tree leaf is first quantized to one bit, and then aggregated with a 3-layer network of 5-input majority gates for efficient implementation of AIGs.

Tree-based models may not perform well in symmetric functions or complex arithmetic functions. However, patterns in the importance of input bits can be observed for some pre-defined standard functions such as adders, comparators, outputs of XOR or MUX. Before ML, Team 7 checks if the training data come from a symmetric function, and compares training data with each identified special function. In case of a match, an AIG of the identified function is constructed directly without ML.

\textbf{Team 8's}
\label{s_team8}
solution is an ensemble drawing from multiple classes of models. It includes a multi-layer perceptron (MLP), binary decision tree (BDT) augmented with functional decomposition, and a RF.
These models are selected to capture various types of circuits.  For all benchmarks, all models are trained independently, and the model with the best validation accuracy that results in a circuit with under 5000 gates is selected. The MLP uses a periodic activation instead of the traditional ReLU to learn additional periodic features in the input. It has three layers, with the number of neurons divided in half between each layer. The BDT is a customized implementation of the C4.5 tree that has been modified with functional decomposition in the cases where the information gain is below a threshold. The RF is a collection of 17 trees limited to a maximum depth of 8. RF helps especially in the cases where BDT overfits.

After training, the AIGs of the trained models are generated to ensure they are under 5000 gates. 
In all cases, the generated AIGs are simplified using the Berkeley ABC tool to produce the final AIG graph.

\textbf{Team 9's}
\label{s_team9}
proposes a Bootstrapped flow that explores the search algorithm Cartesian Genetic Programming (CGP). CGP is an evolutionary approach proposed as a generalization of Genetic Programming used in the digital circuit’s domain. It is called Cartesian because the candidate solutions are composed of a two-dimensional network of nodes. 
CGP is a population-based approach often using the evolution strategies (1+$\lambda$)-ES algorithm for searching the parameter space. Each individual is a circuit, represented by a two-dimensional integer matrix describing the functions and the connections among nodes. 

The proposed flow decides between two initialization: 1) starts the CGP search from random (unbiased) individuals seeking for optimal circuits; or, 2) exploring a bootstrapped initialization with individuals generated by previously optimized SOPs created by decision trees or ESPRESSO when they provide AIGs with more than 55\% of accuracy. 
This flow restricts the node functions to XORs, ANDs, and Inverters; in other words, we may use AIG or XAIG to learn the circuits.
Variation is added to the individuals through mutations seeking to find a circuit that optimizes a given fitness function. The mutation rate is adaptive, according to the $1/5^{th}$ rule \cite{doerr:15}.  
When the population is bootstrapped with DTs or SOP, the circuit is fine-tuned with the whole training set. 
When the random initialization is used, it was tested with multiple configurations of sizes and mini-batches of the training set that change based on the number of generations processed.  

\textbf{Team 10's}
\label{s_team10}
solution learns Boolean function representations, using DTs. We developed a Python program using the \emph{Scikit-learn} library where the parameter \emph{max\_depth} serves as an upper-bound to the growth of the trees, and is set to be 8. The training set PLA, treated as a numpy matrix, is used to train the DT. On the other hand, the validation set PLA is then used to test whether the obtained DT meets the minimum validation accuracy, which we empirically set to be 70\%. If such a condition is not met, the validation set is merged with the training set. According to empirical evaluations, most of the benchmarks with accuracy $< 70\%$ showed a validation accuracy fluctuating around 50\%, regardless of the size and shapes of the DTs. This 
suggests that the training sets were not able to provide enough representative cases to effectively exploit the adopted technique, thus leading to DTs
with very high training accuracy, but completely negligible performances. For DTs having a validation
accuracy $\geq$ 70\%, 
the tree structure is annotated as a Verilog netlist, where each DT node is replaced with a
multiplexer. The obtained
Verilog netlist is then processed with the ABC Synthesis Tool in order to generate a compact and optimized
AIG structure.
This approach has shown an average accuracy over the validation set of 84\%, with an average size of AIG of 140 nodes (and no AIG with more than 300 nodes). More detailed information about the adopted technique can be found in \cite{rizzo}.



%% file: results.tex


\subsection{Accuracy}

\begin{table}[!t]
\centering
\caption{Performance of the different teams}
\renewcommand{\arraystretch}{1}
 	\def\arraystretch{}
 	\resizebox{\columnwidth}{!}{
\begin{tabular}{rrrrr}
\toprule
team & $\downarrow$ test accuracy & And gates  & levels  & overfit \\
\midrule
1    & {\bf 88.69}     & 2517.66 & 39.96      & 1.86             \\
7    & 87.50     & 1167.50  & 32.02     & {\bf 0.05}             \\
8    & 87.32     & 1293.92 & 21.49     & 0.14             \\
3    & 87.25     & 1550.33 & 21.08    
   & 5.76             \\2    & 85.95     & 731.92  & 80.63    
   & 8.70             \\9    & 84.65     & 991.89  & 103.42   
  & 1.75             \\4    & 84.64     & 1795.31 & 21.00    
    & 0.48             \\5    & 84.08     & 1142.83 & 145.87   
   & 4.17             \\10   & 80.25     & {\bf 140.25}  & 10.90    
   & 3.86             \\6    & 62.40     & 356.26  & {\bf 8.73}     
 & 0.88            \\ \bottomrule
\end{tabular}}
\label{table:accuracy}
\end{table}

\begin{figure}[t]
	\centering
	\scalebox{0.8}{\includegraphics[width=\columnwidth]{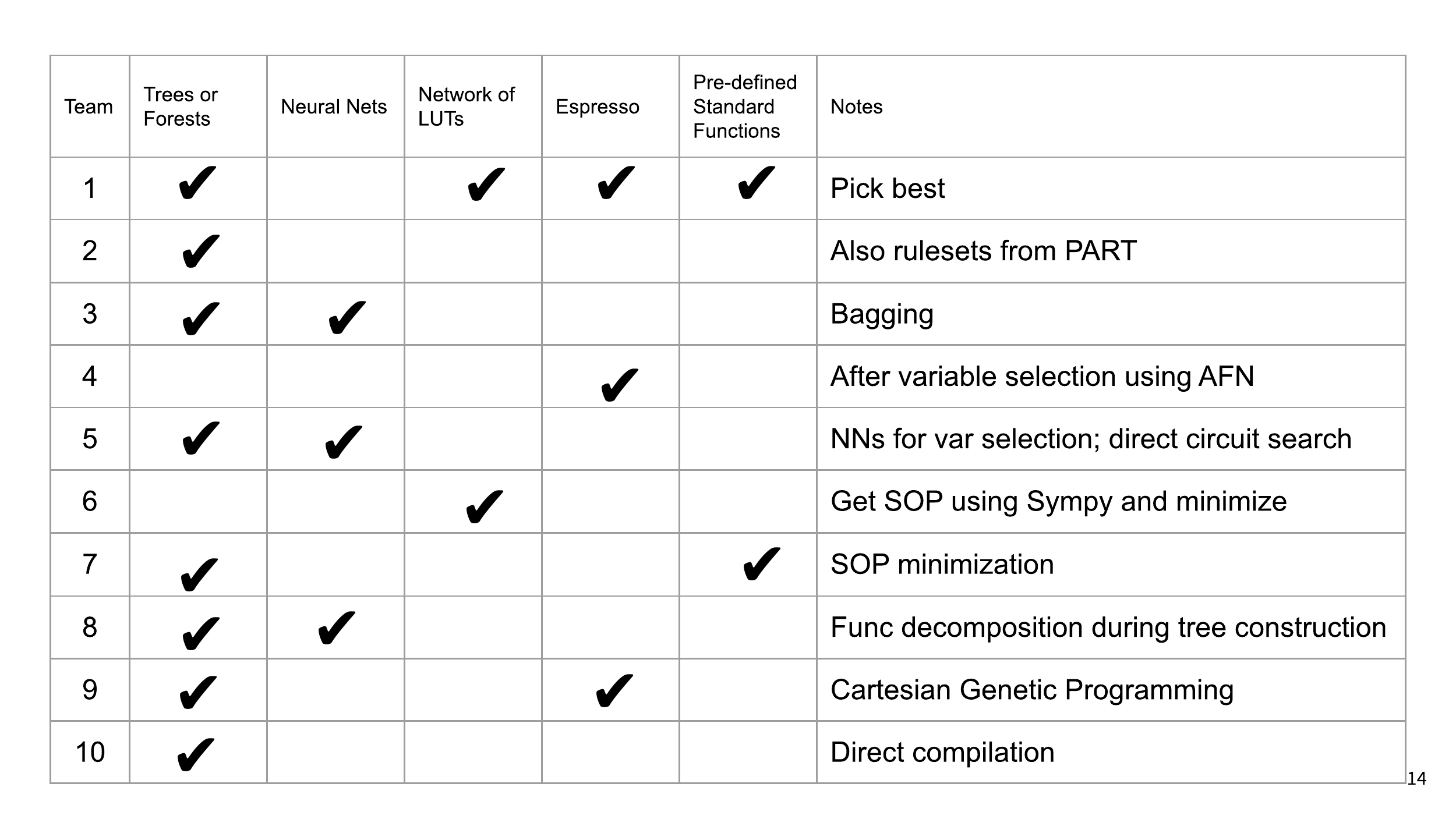}}
	\caption{Representation used by various teams}
	\label{fig:techniques}
\end{figure}

\tablename{~\ref{table:accuracy}} shows the average accuracy of the solutions found by all the 10 teams, along with the average circuit size, the average number of levels in the circuit, and the overfit measured as the average difference between the accuracy on the validation set and the test set. The following interesting observations can be made: \textbf{(i)} most of the teams achieved more than $80\%$ accuracy. \textbf{(ii)} the teams were able to find circuits with much fewer gates than the specification.

\begin{figure}[!t]
	\centering
	\scalebox{0.8}{\includegraphics[width=\columnwidth]{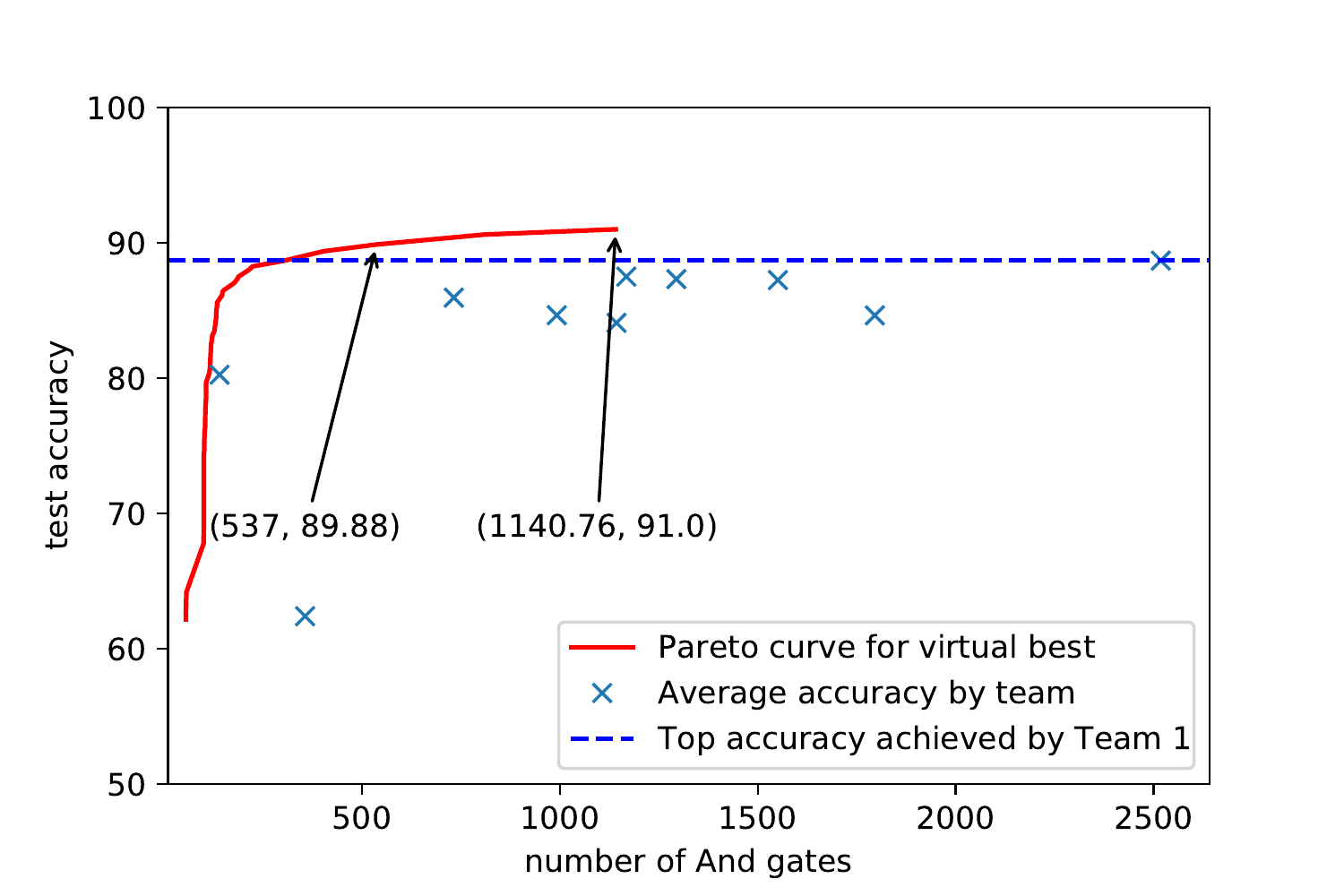}}
	\caption{Acc-size trade-off across teams and for virtual best}
	\vspace{-2em}
	\label{fig:area}
\end{figure}

\begin{figure}
	\centering
	\scalebox{0.8}{\includegraphics[width=\columnwidth]{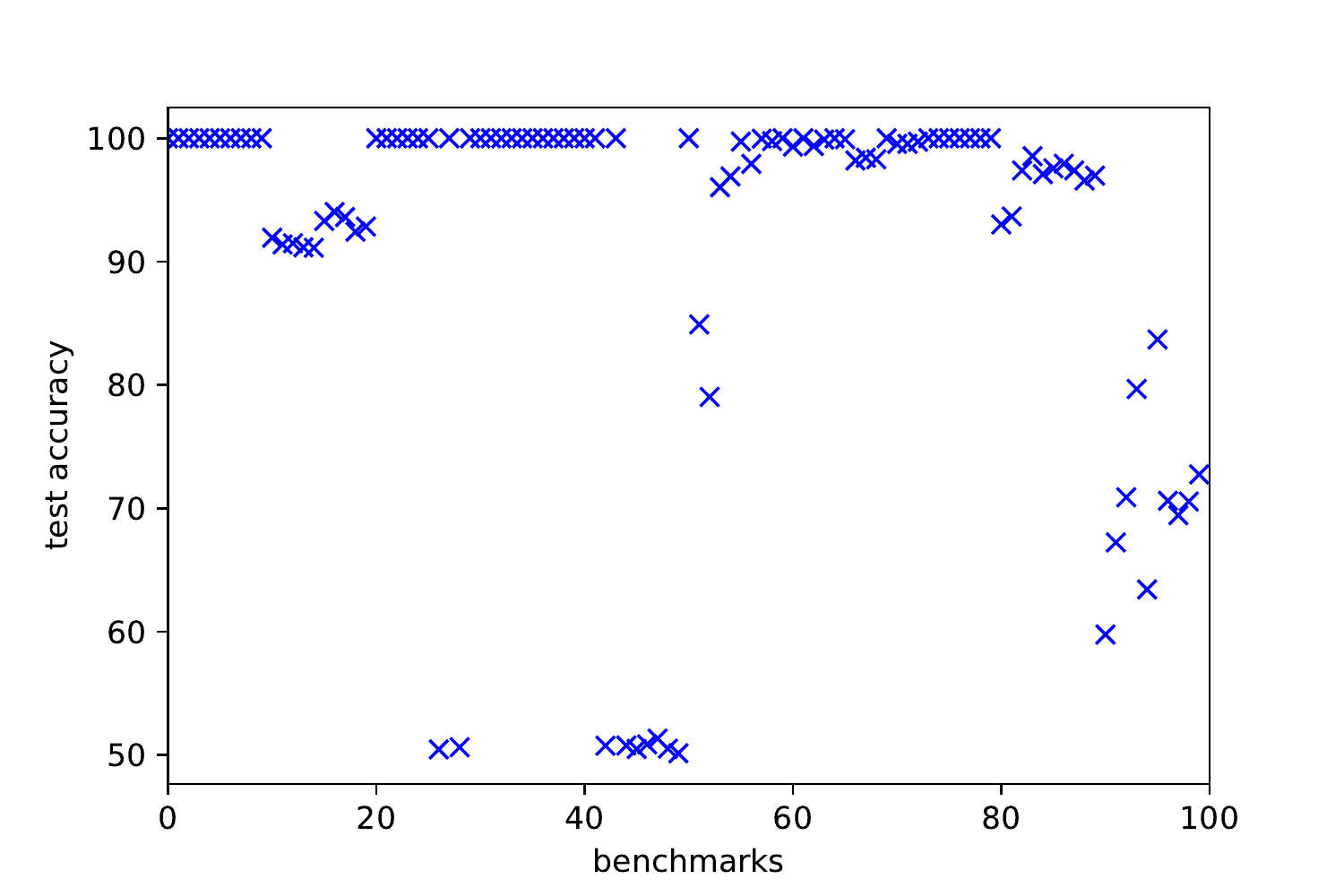}}
	\caption{Maximum accuracy achieved for each example}
	\vspace{-2em}
	\label{fig:max-acc-example}
\end{figure}



When it comes to comparing network size \emph{vs} accuracy, there is no clear trend. For instance, teams 1 and 7 have similar accuracy, with very divergent number of nodes, as seen in~\tablename{~\ref{table:accuracy}}. For teams that have relied on just one approach, such as Team 10 and 2 who used only decision trees, it seems that more AND nodes might lead to better accuracy. 
Most of the teams, however, use a portfolio approach, and for each benchmark choose an appropriate technique. Here it is worth pointing out that there is no approach, which is consistently better across all the considered benchmarks.
Thus, applying several approaches and deciding which one to use, depending on the target Boolean functions, seems to be the best strategy.~\figurename{~\ref{fig:techniques}} presents the approaches used by each team. 




While the size of the network was not one of the optimization criteria in the contest, it is an important parameter considering the hardware implementation, as it impacts area, delay, and power. The average area reported by individual teams are shown in~\figurename{~\ref{fig:area}} as `$\times$'. 
Certain interesting observations can be made from~\figurename{~\ref{fig:area}}. Apart from showing the average size reached by various teams, it also shows the \emph{Pareto-curve} between the average accuracy across all benchmarks and their size in terms of number of AND gates. It can be observed that while 91\% accuracy constraint requires about 1141 gates, a reduction in accuracy constraint of merely 2\%, requires a circuit of only half that size.
This is an insightful observation which strongly suggests that with a slight compromise in the accuracy, much smaller size requirements can be satisfied.


Besides area, it is also worth to look for the number of logic-levels in the generated implementation, as it correlates with circuit delay. Similar to the number of nodes, there is no clear distinction on how the number of levels impacts the final accuracy. Team 6 has delivered the networks with the smallest depth, often at the cost of accuracy. In practice, the winning team has just the 4th larger depth among the 10 teams. 

Finally, ~\figurename{~\ref{fig:max-acc-example}} shows the maximum accuracy achieved for each benchmarks. While most of the benchmarks achieved a 100\% accuracy, several benchmarks only achieved close to 50\%. That gives an insight on which benchmarks are harder to generalize, and these benchmarks might be used as test-case for further developments on this research area. 


\subsection{Generalization gap} 

The generalization gap for each team is presented in the last column of~\tablename{~\ref{table:accuracy}}. This value presents how well the learnt model can generalize on an unknown set. Usually, a generalization gap ranging from 1\% to 2\% is considered to be good. It is possible to note that most of the teams have reached this range, with team 7 having a very small gap of 0.05\%. Furthermore, given that the benchmark functions are incompletely-specified, with a very small subset of minterms available for training, reaching generalization in small networks is a challenge. Therefore, a special mention must be given to Team 10, who reached a high level of accuracy with extremely small network sizes. 

\subsection{Win-rate for different teams}
\begin{figure}[t]
	\centering
	\scalebox{0.8}{\includegraphics[width=\columnwidth]{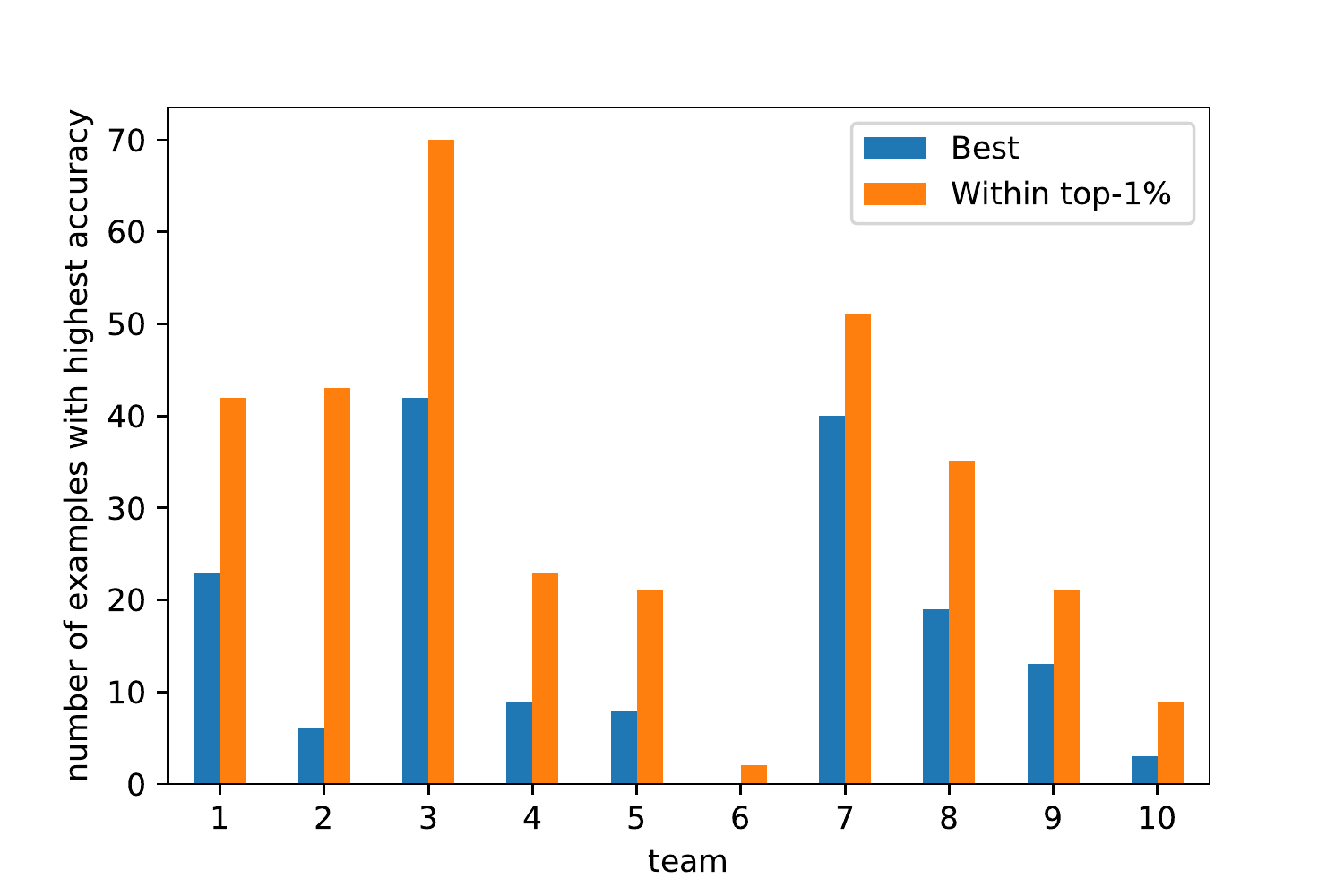}}
	\caption{Top-accuracy results achieved by different teams}
	\vspace{-2ex}
	\label{fig:team-rate}
	\vspace{-4pt}
\end{figure}

~\figurename{~\ref{fig:team-rate}} shows a bar chart showing which team achieved the best accuracy and the top-1\% for the largest number of benchmarks. Team 3 is the winner in terms of both of these criteria, achieving the best accuracy among all the teams for 42 benchmark. Following Team 3, is Team 7, and then the winning Team 1. Still, when it comes to the best average accuracy, Team 1 has won the contest. This figure gives insights on 
which approaches have been in the top of achieved accuracy more frequently, and give a pointer to what could be the ideal composition of techniques to achieve high-accuracy. As shown in ~\figurename{~\ref{fig:techniques}}, indeed a portfolio of techniques needs to be employed to achieve high-accuracy, since there is no single technique that dominates. 

%

%% file: conclusion.tex
In this work, we explored the connection between logic synthesis of incompletely specified functions and supervised learning. This was done via a programming contest held at the 2020 International Workshop on Logic and Synthesis where the objective was to synthesize small circuits that generalize well from input-output samples.

The solutions submitted to the contest used a variety of techniques spanning logic synthesis and machine learning. Portfolio approaches ended up working better than individual techniques, though random forests formed a strong baseline. Furthermore, by sacrificing a little accuracy, the size of the circuit could be greatly reduced.
These findings suggest an interesting direction for future work: When exactness is not needed, can synthesis be done using machine learning algorithms to greatly reduce area and power?

Future extensions of this contest could  target circuits with multiple outputs and algorithms generating an optimal trade-off between accuracy and area (instead of a single solution).

 

%% file: approach.tex
\renewcommand{\baselinestretch}{1}

The detailed version of approaches adopted by individual teams are described below.
\section{\large team 1}
\label{team1}
{Authors: Yukio Miyasaka, Xinpei Zhang, Mingfei Yu, Qingyang Yi, Masahiro Fujita, \emph{The University of Tokyo, Japan and University of California, Berkeley, USA} \\}
\input{Texfiles_from_collaborator/team1/main.tex}

\section{\large team 2}
\label{team2}
{Authors: Guilherme Barbosa Manske, Matheus Ferreira Pontes, Leomar Soares da Rosa Junior, Marilton Sanchotene de Aguiar, Paulo Francisco Butzen, \emph{Universidade Federal de Pelotas, Universidade Federal do Rio Grande do Sul, Brazil}\\ }

\input{Texfiles_from_collaborator/team2/main.tex}


\section{\large team 3}
\label{team3}
{Authors: Po-Chun Chien, Yu-Shan Huang, Hoa-Ren Wang, and Jie-Hong R. Jiang, 
\emph{Graduate Institute of Electronics Engineering, Department of Electrical Engineering,
National Taiwan University,
Taipei, Taiwan}	\\
}
\input{Texfiles_from_collaborator/team3/report.tex}

\section{\large team4}
\label{team4}
{Authors: Jiaqi Gu, Zheng Zhao, Zixuan Jiang, David Z. Pan,
\emph{Department of Electrical and Computer Engineering, University of Texas at Austin, USA} \\}
\input{Texfiles_from_collaborator/team4/main.tex}

\section{\large team 5}
\label{team5}
{Authors: Brunno Alves de Abreu, Isac de Souza Campos, Augusto Berndt,  Cristina Meinhardt, Jonata Tyska Carvalho, Mateus Grellert and Sergio Bampi,
\emph{Universidade Federal do Rio Grande do Sul, Universidade Federal de Santa Catarina, Brazil}\\}

\input{Texfiles_from_collaborator/team5/main.tex}

\section{\large team 6}
\label{team6}
{Authors: Aditya Lohana, Shubham Rai and Akash Kumar,
\emph{Chair for Processor Design, 
Technische Universitaet Dresden, Germany} \\}
\input{Texfiles_from_collaborator/team6/main.tex}

\vspace{3mm}
\section{\large team7: Learning with Tree-Based Models and Explanatory Analysis}
\label{team7}
{Authors: Wei Zeng, Azadeh Davoodi, and Rasit Onur Topaloglu,
\emph{University of Wisconsin--Madison, IBM, USA} \\
}
\input{Texfiles_from_collaborator/team7/main.tex}

%
%
%
\section{\large team 8: Learning Boolean Circuits with ML Model Ensemble}
\label{team8}
{Authors: Yuan Zhou, Jordan Dotzel, Yichi Zhang, Hanyu Wang, Zhiru Zhang,
\emph{School of Electrical and Computer Engineering, Cornell University, Ithaca, NY, USA} \\
}
\input{Texfiles_from_collaborator/team8/main.tex}

\section{\large team 9:Bootstrapped CGP Learning Flow
}
\label{team9}
{Authors: Augusto Berndt, Brunno Alves de Abreu, Isac de Souza Campos, Cristina Meinhardt, Mateus Grellert, Jonata Tyska Carvalho, 
\emph{Universidade Federal de Santa Catarina,
Universidade Federal do Rio Grande do Sul, Brazil
} \\
}
\input{Texfiles_from_collaborator/team9/long.tex}

\section{\large team 10}
\label{team10}
{Authors: Valerio Tenace, Walter Lau Neto, and Pierre-Emmanuel Gaillardon,
\emph{University of Utah, USA} \\}

\begin{figure}[h]
    \centering
    \includegraphics[width=0.4\textwidth]{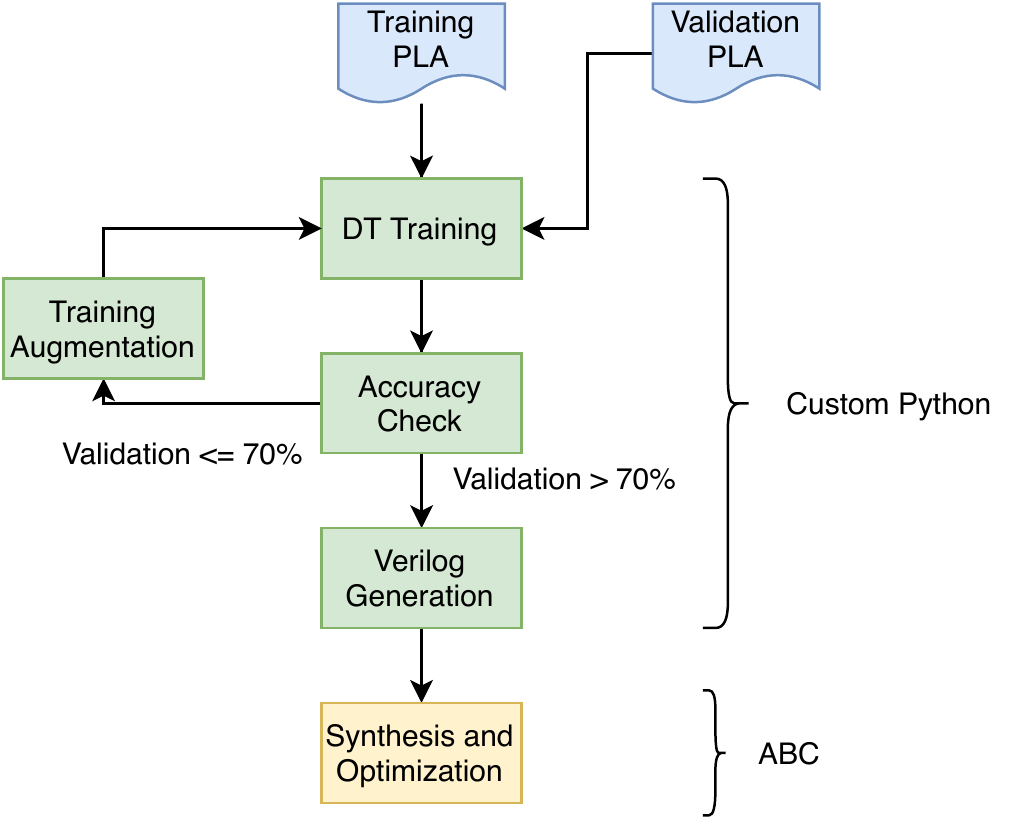}
    \caption{Overview of adopted flow.}
    \label{fig:team10}
\end{figure}

In order to learn incompletely specified Boolean functions for the contest, we decided to resort to decision trees (DTs). ~\figurename{~\ref{fig:team10}} presents an overview of the adopted design flow.

\begin{figure}[h]
    \centering
    \includegraphics[width=0.5\textwidth]{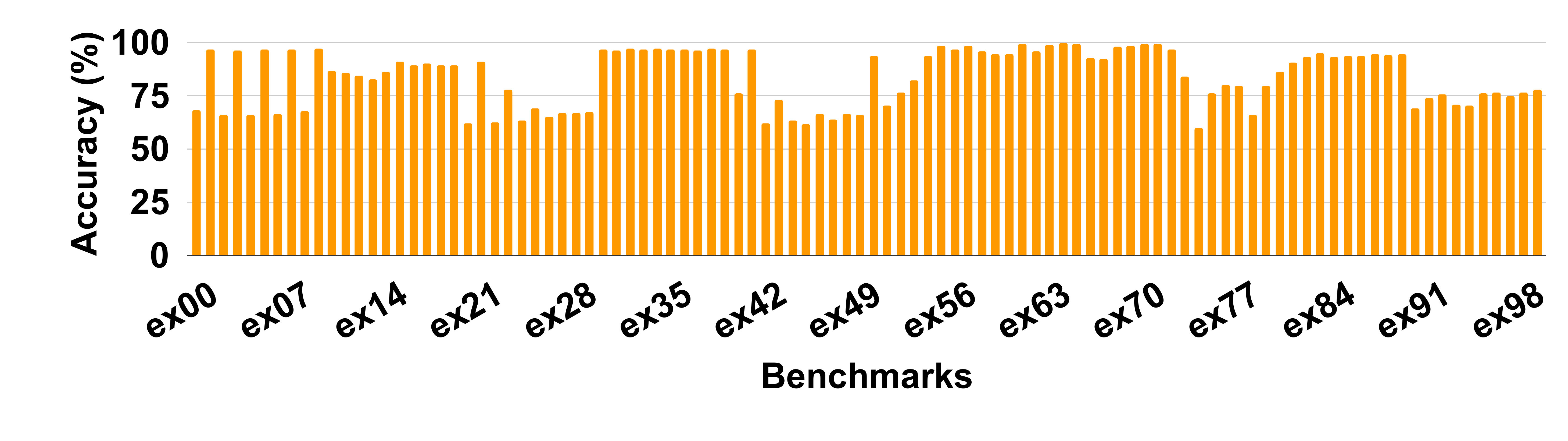}
    \caption{Accuracy for different benchmarks.}
    \label{fig:team10-acc}
\end{figure}

\begin{figure}[h]
    \centering
    \includegraphics[width=0.5\textwidth]{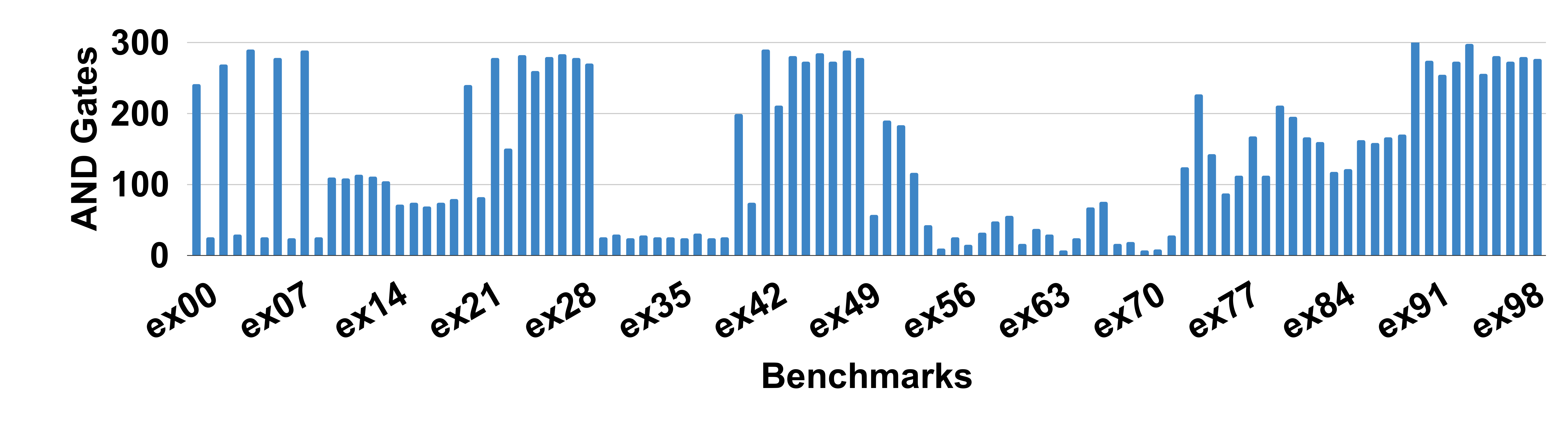}
    \caption{Number of AIG nodes for different benchmarks.}
    \label{fig:team10-size}
\end{figure}

We developed a Python program using the \emph{Scikit-learn} library where the parameter \emph{max\_depth} serves as an upper-bound to the growth of the trees. Through many rounds of evaluation, we opted to set the \emph{max\_depth} to 8, as it gives small AIG networks without sacrificing the accuracy too much. 
The training set PLAs, provided by the contest, are read into a numpy matrix, which is used to train the DT. On the other hand, the validation set PLA is used to test whether the obtained DT meets the minimum validation accuracy, which is empirically set to be 70\%. If such a condition is not met, the training set is augmented by merging it with the validation set. According to empirical evaluations, most of these benchmarks with performance $< 70\%$ showed a validation accuracy fluctuating around 50\%, regardless of the size and shapes of the DTs. This suggests that the training sets were not able to provide enough representative cases to effectively exploit the adopted technique, thus leading to DTs with very high training accuracy, but completely negligible performances. For DTs having a validation accuracy $\geq 70\%$, the tree structure is annotated as a Verilog netlist, where each DT node is replaced with a multiplexer. In cases where the accuracy does not achieve the lower bound of 70\%, we re-train the DT with the augmented set, and then annotate it in Verilog. The obtained netlist is then processed with the ABC Synthesis Tool in order to generate a compact and optimized
AIG structure.
More detailed information about the adopted technique can be found in \cite{rizzo}. Overall, our approach was capable of achieving very good accuracy for most of the cases, without exceeding 300 AIG nodes in any benchmark, thus yielding the smallest average network sizes among all the participants. In many cases, we achieved an accuracy greater than 90\% with less than 50 nodes, and a mean accuracy of $84\%$ with only 140 AND gates on average. \figurename{~\ref{fig:team10-acc}} presents the accuracy achieved for all the adopted benchmarks, whereas ~\figurename{~\ref{fig:team10-size}} shows the AIG size for the same set of circuits. These results clearly show that DTs are a viable technique to learn Boolean functions efficiently. 

%% file: Texfiles_from_collaborator/team1/main.tex
\subsection{Learning Methods}

We tried ESPRESSO, LUT network, and Random forests. ESPRESSO will work well if the underlying function is small as a 2-level logic. On the other hand, LUT network has a multi-level structure and seems good for the function which is small as a multi-level logic. Random forests also support multi-level logic based on a tree structure.

ESPRESSO is used with an option to finish optimization after the first irredundant operation. LUT network has some parameters: the number of levels, the number of LUTs in each level, and the size of each LUT. These parameters are incremented like a beam search as long as the accuracy is improved. The number of estimators in Random forests is explored from 4 to 16.

If the number of AIG nodes exceeds the limit (5000), a simple approximation method is applied to the AIG. The AIG is simulated with thousands of random input patterns, and the node which most frequently outputs 0 is replaced by constant-0 while taking the negation (replacing with constant-1) into account. This is repeated until the AIG size meets the condition. To avoid the result being constant 0 or 1, the nodes near the outputs are excluded from the candidates by setting a threshold on level. The threshold is explored through try and error.

\subsection{Preliminary Experiment}

\begin{figure}[!ht]
	\centering
	\scalebox{1}{\includegraphics[width=\columnwidth]{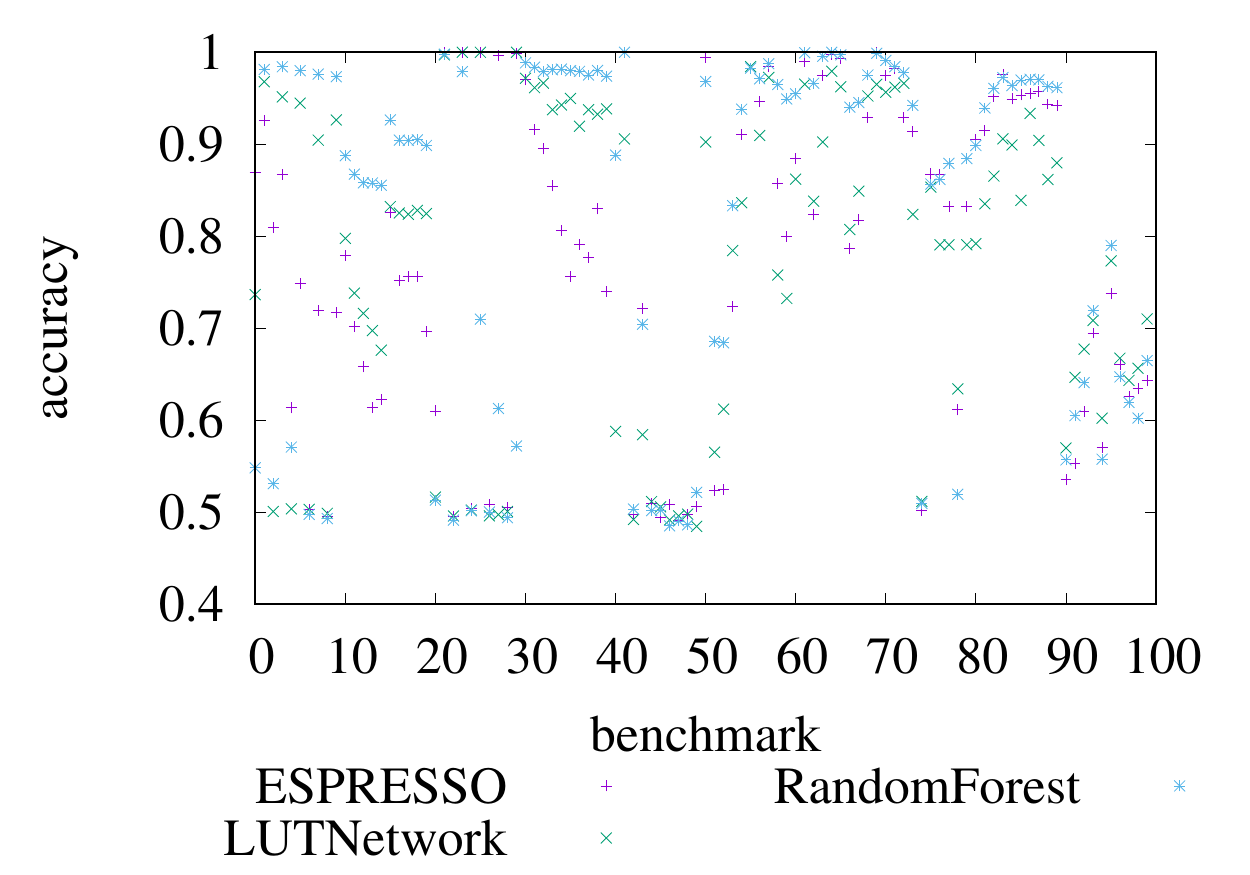}}
	\caption{The test accuracy of the methods}
	\label{fig:preex_acc}
\end{figure}

\begin{figure}[ht]
	\centering
	\scalebox{1.0}{\includegraphics[width=\columnwidth]{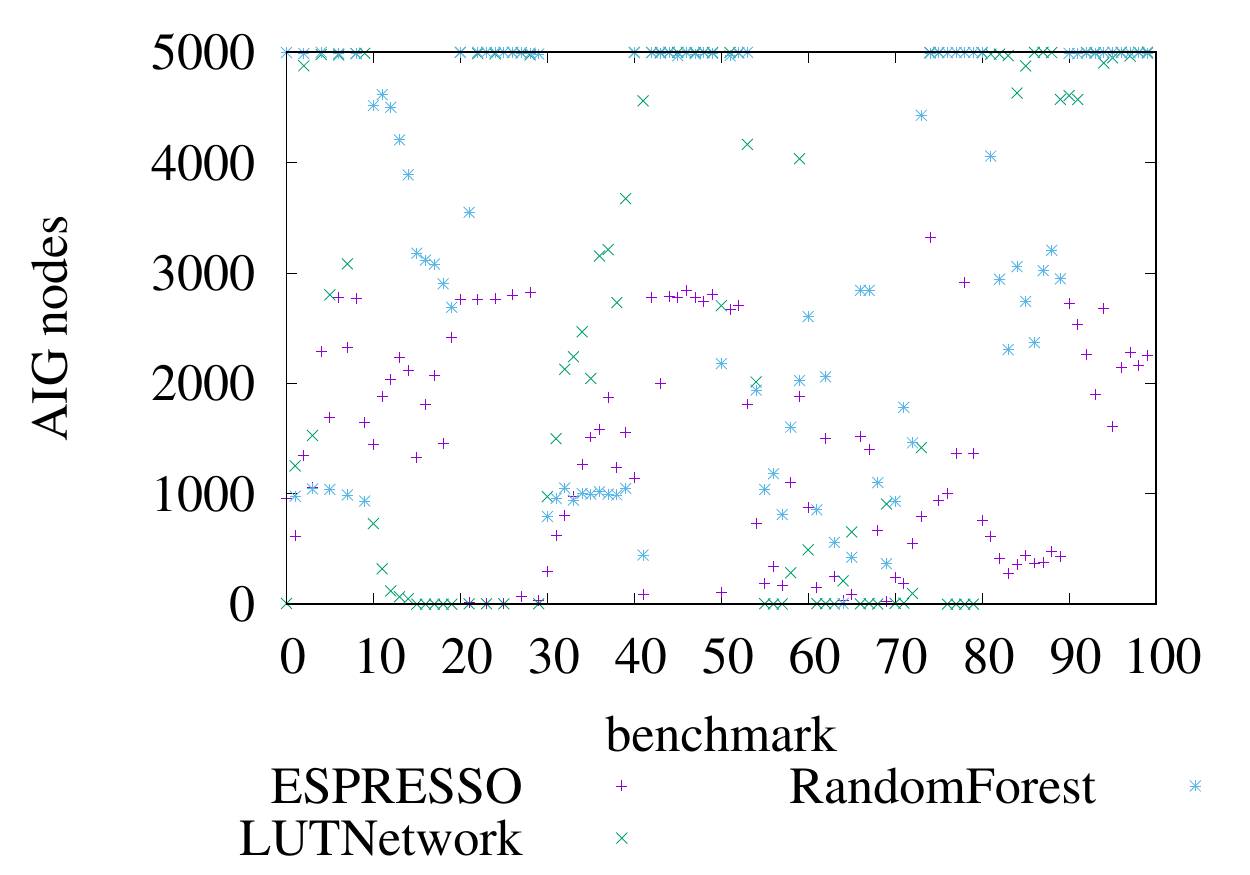}}
	\caption{The resulting AIG size of the methods}
	\label{fig:preex_size}
\end{figure}

We conducted a simple experiment. The parameters of LUT was fixed as follows: the number of levels was 8, the number of LUTs in a level was 1028, and the size of each LUT was 4. The number of estimators in Random forests was 8. The test accuracy and the AIG size of the methods is shown at~\figurename{~\ref{fig:preex_acc}} and~\figurename{~\ref{fig:preex_size}}. Generally Random forests works best, but LUT network works better in a few cases among case 90-99. All methods failed to learn case 0, 2, 4, 6, 8, 20, 22, 24, 26, 28, and 40-49. The approximation was applied to AIGs generated by LUT network and Random forests for these difficult cases and case 80-99. ESPRESSO always generates a small AIG with less than 5000 nodes as it conforms to less than ten thousands of min-terms.

\begin{figure}[ht]
	\centering
	\scalebox{1.0}{\includegraphics[width=\columnwidth]{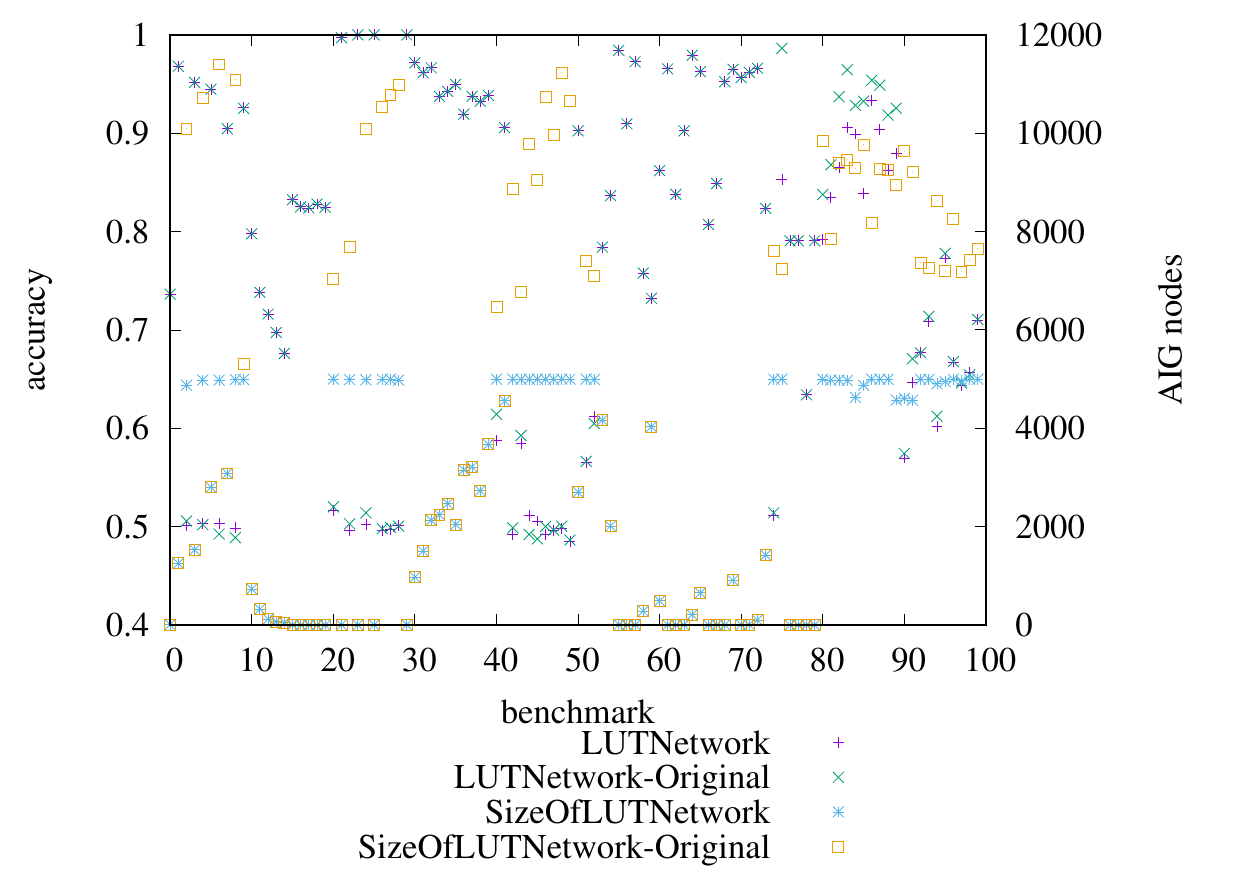}}
	\caption{The test accuracy and AIG size of LUT network before and after approximation}
	\label{fig:preex_LUTapprox}
\end{figure}

The effect of approximation of the AIGs generated by LUT network is shown at~\figurename{~\ref{fig:preex_LUTapprox}}. For difficult cases, the accuracy was originally around 50\%. For case 80-99, the accuracy drops at most 5\% while reducing 3000-5000 nodes. Similar thing was observed in the AIGs generated by Random forests.

\subsection{Pre-defined standard function matching}

The most important method in the contest was actually matching with a pre-defined standard functions. There are difficult cases where all of the methods above fail to get meaningful accuracy. We analyzed these test cases by sorting input patterns and was able to find adders, multipliers, and square rooters fortunately because the inputs of test cases are ordered regularly from LSB to MSB for each word. Nevertheless, it seems almost impossible to realize significantly accurate multiplier and square rooters with more than 100 inputs within 5000 AIG nodes.

\subsection{Exploration After The Contest}

\subsubsection{Binary Decision Tree}

We examined BDT (Binary Decision Tree), inspired by the methods of other teams. Our target is the second MSB of adder because only BDT was able to learn the second MSB of adder with more than 90\% accuracy according to the presentation of the 3rd place team. In normal BDT, case-splitting by adding a new node is performed based on the entropy gain. On the other hand, in their method, when the best entropy gain is less than a threshold, the number of patterns where the negation of the input causes the output to be negated too is counted for each input, and case-splitting is performed at the input which has such patterns the most.

First of all, the 3rd team's tree construction highly depends on the order of inputs. Even in the smallest adder (16-bit adder, case 0), there is no pattern such that the pattern with an input negated is also included in the given set of patterns. Their program just chooses the last input in such case and fortunately works in the contest benchmark, where the last input is the MSB of one input word. However, if the MSB is not selected at first, the accuracy dramatically drops. When we sort the inputs at random, the accuracy was 59\% on average of 10 times.

Next, we tried SHAP analysis \cite{shap-tree} on top of XGBoost, based on the method of the 2nd place team, to find out the MSBs of input-words. The SHAP analysis successfully identifies the MSBs for a 16-bit adder, but not that for a larger adder (32-bit adder, case 2).

In conclusion, it is almost impossible for BDT to learn 32-bit or larger adders with random input order. If the problem provides a black box simulator as in ICCAD contest 2019 problem A \cite{Chen2020}, we may be able to know the MSBs by simulating one-bit flipped patterns, such as one-hot patterns following an all-zero pattern. Another mention is that BDT cannot learn a large XOR (16-XOR, case 74). This is because the patterns are divided into two parts after each case-splitting and the entropy becomes zero at a shallow level. So, BDT cannot learn a deep adder tree (adder with more than 2 input-words), much less multipliers.

\subsubsection{Binary Decision Diagram}

We also tried BDD (Binary Decision Diagram) to learn adder. BDD minimization using don't cares \cite{Shiple1994} is applied to the BDD of the given on-set. Given an on-set and a care-set, we traverse the BDD of on-set while replacing a node by its child if the other child is don't care (one-sided matching), by an intersection of two children if possible (two-sided matching), or by an intersection between a complemented child and the other child if possible (complemented two-sided matching). Unlike BDT, BDD can learn a large XOR up to 24-XOR (using 6400 patterns) because patterns are shared where nodes are shared.

BDD was able to learn the second MSB of adder tree only if the inputs are sorted from MSB to LSB mixing all input-words (the MSB of the first word, the MSB of the second word, the MSB of the third word, ..., the MSB of the last word, the second MSB of the first word, ...). For normal adder (2 words), one-sided matching achieved 98\% accuracy. The accuracy was almost the same among any bit-width because the top dozens of bits control the output. For 4-word adder tree, one-sided matching got around 80\% accuracy, while two-sided matching using a threshold on the gain of substitution achieved around 90\% accuracy. Note that naive two-sided matching fails (gets 50\% accuracy). Furthermore, a level-based minimization, where nodes in the same level are merged if the gain does not exceed the threshold, achieved more than 95\% accuracy. These accuracy on 4-word adder tree is high compared to BDT, whose accuracy was only 60\% even with the best ordering.

For 6-word adder tree, the accuracy of the level-based method was around 80\%. We came up with another heuristic that if both straight two-sided matching and complemented two-sided matching are available, the one with the smaller gain is used, under a bias of 100 nodes on the complemented matching This heuristic increased the accuracy of the level-based method to be 85-90\%. However, none of the methods above obtained meaningful (more than 50\%) accuracy for 8-word adder tree.

We conclude that BDD can learn a function if the BDD of its underlying function is small under some input order and we know that order. The main reason for minimization failure is that merging inappropriate nodes is mistakenly performed due to a lack of contradictory patterns. Our heuristics prevent it to some degree. If we have a black box simulator, simulating patterns to distinguish straight and complemented two-sided matching would be helpful. Reordering using don't cares is another topic to explore.

%% file: Texfiles_from_collaborator/team2/main.tex
\subsection{Proposed solution}


Our solution is a mix of two machine learning techniques, J48~\cite{c45} and PART~\cite{part}. We will first present a general overview of our solution. Then we will focus on individual machine learning classifier techniques. Finally, we present the largest difference that we have found between both strategies, showing the importance of exploring more than only one solution. 

\begin{figure}[htbp]
\centerline{\includegraphics[scale=.25]{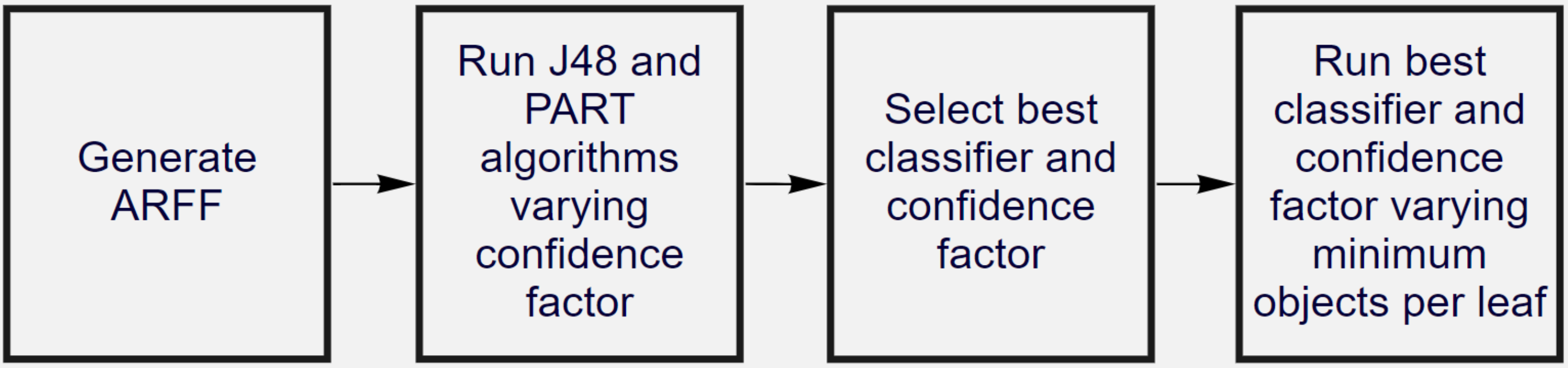}}
\caption{Solution's flowchart.}
\label{flowchart}
\end{figure}

The flowchart in~\figurename{~\ref{flowchart}} illustrates our overall solution. The first step was to convert the PLA structure into one that Weka~\cite{weka} could interpret. We decided to use the ARFF structure due to how the structure of attributes and classes is arranged.

After converting to ARFF, as shown in the flowchart's second block, the  J48 and PART algorithms are executed. In this stage of the process, the confidence factor is varied in five values (0.001, 0.01, 0.1, 0.25, and 0.5) for each algorithm. In total, this step will generate ten results. The statistics extracted from cross-validation were used to determine the best classifier and the best confidence factor.


This dynamic selection between the two classifiers and confidence factors was necessary since a common configuration was not found for all the examples provided in the IWLS Contest. After selecting the best classifier and the best confidence factor, six new configurations are performed. At this point, the parameter to be varied is the minimum number of instances per sheet, that is, the Weka parameter "-M". The minimum number of instances per sheet was defined (0, 1, 3, 4, 5, and 10). Again, the selection criterion was the result of the cross-validation statistic.

\subsubsection{J48}
Algorithms for constructing decision trees are among the most well known and widely used machine learning methods.  With decision trees, we can classify instances by sorting them based on feature values. We classify the samples starting at the root node and sorted based on their feature values so that in each node, we represent a feature in an example to be classified, and each branch represents a value that the node can assume.  In the machine learning community, J. Ross Quinlan's ID3 and its successor, C4.5~\cite{c45}, are the most used decision tree algorithms. J48 is an open-source Java implementation of the C4.5 decision tree algorithm in Weka.

The J48 classifier output is a decision tree, which we transform into a new PLA file.  First, we go through the tree until we reach a leaf node, saving all internal nodes' values in a vector. When we get to the leaf node, we use the vector data to write a line of the PLA file. After, we read the next node, and the amount of data that we keep in the vector depends on the height of this new node. 

Our software keeps going through the tree until the end of the J48 file, and then it finishes the PLA file writing the metadata. Finally, we use the ABC tool to create the AIG file, using the PLA file that our software has created.



\begin{figure}[htbp]
\centerline{\includegraphics[scale=.4]{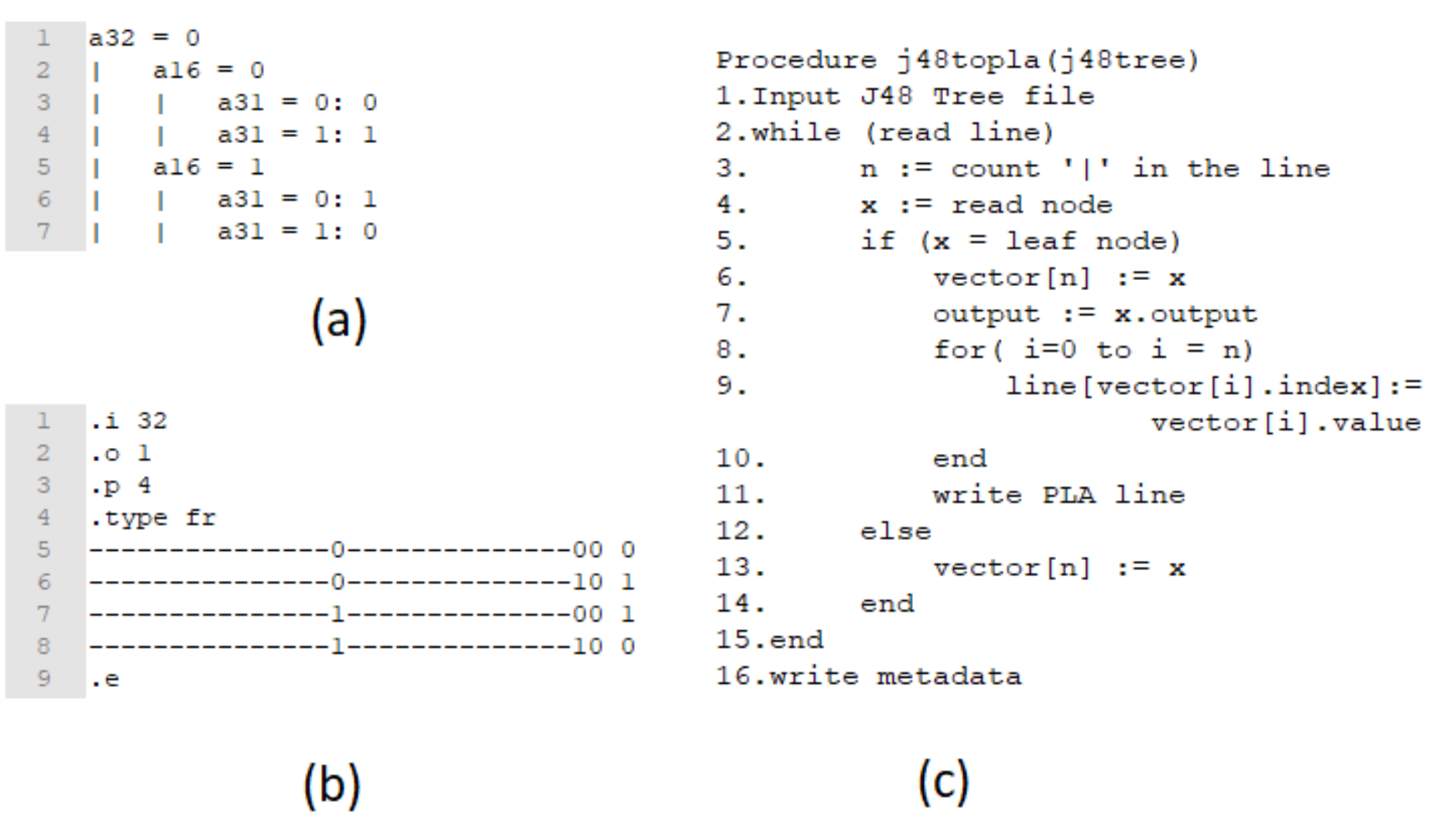}}
\caption{(a) Decision tree (J48 output), (b) PLA file generated by our software and (c) pseudocode j48topla.}
\label{j482}
\end{figure}

In~\figurename{~\ref{j482}}, we show an example of how our software works.~\figurename{~\ref{j482}}~(a) shows a decision tree (J48 output) with 7 nodes, 4 of which are leaves.~\figurename{~\ref{j482}}~(b) shows the PLA file generated by our software. The PLA file has 1 line for every leaf in the tree.~\figurename{~\ref{j482}}~(c) shows the pseudocode j48topla, where \textit{n} control the data in the vector and \textit{x} is the node read in the line.

\subsubsection{PART}




In the PART algorithm~\cite{part}, we can infer rules generating partial decision trees. Thus two major paradigms for rule generation are combined: creating rules from decision trees and the separate-and-conquer rule learning technique. Once we build a partial tree, we extract a single rule from it, and for this reason, the PART algorithm avoids post-processing.

The PART classifier's output is a set of rules, which checks from the first to the last rule to define the output for a given input. We transform this set of rules in an AAG file, and to follow the order of the rules, we have created a circuit that guarantees this order.  Each rule is an AND logic gate with a varied number of inputs. 

First, we go through the PART file and create all the rules (ANDs), inverting the inputs that are 0. We need to save the values and positions of all rules in a data structure. After, we read this data structure, connecting all the outputs. If a rule makes the output goes to 1, we add an OR logic gate to connect with the rest of the rules. If a rule makes the output goes to 0, we add an AND logic gate, with an inverter in the first input. These guarantees that the first correct rule will define the output. Finally, we use the AIGER~\cite{aiger} library to convert the created AAG to AIG.

\begin{figure}[htbp]
\centerline{\includegraphics[scale=.6]{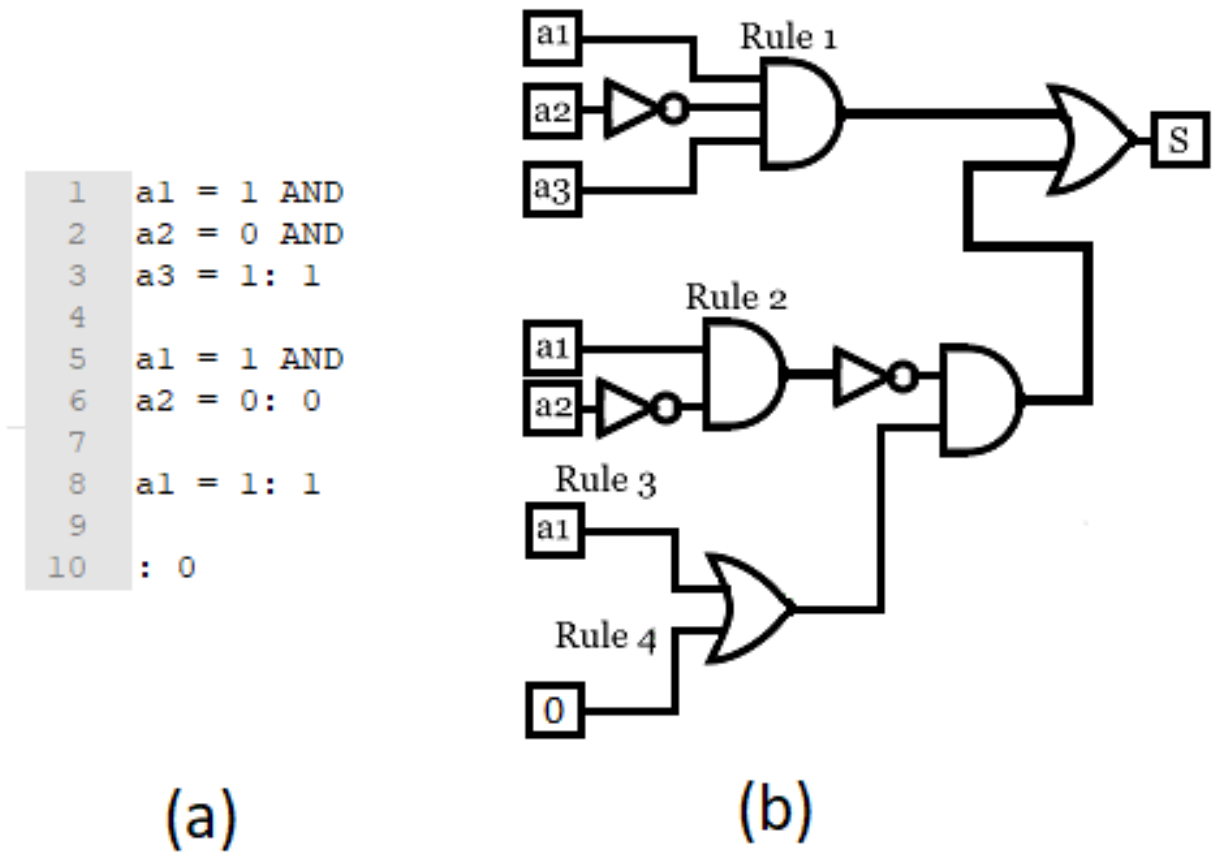}}
\caption{(a) Set of rules (PART file) and (b) circuit created with this set of rules.}
\label{part}
\end{figure}

In~\figurename{~\ref{part}}, we can see how this circuit is created.~\figurename{~\ref{part}}(a) shows a set of rules (PART file) with four rules, where a blank line separates each rule.~\figurename{~\ref{part}}~(b) shows the circuit created with this set of rules.

\subsection{Results}
\begin{figure}[htbp]
\centerline{\includegraphics[scale=.5]{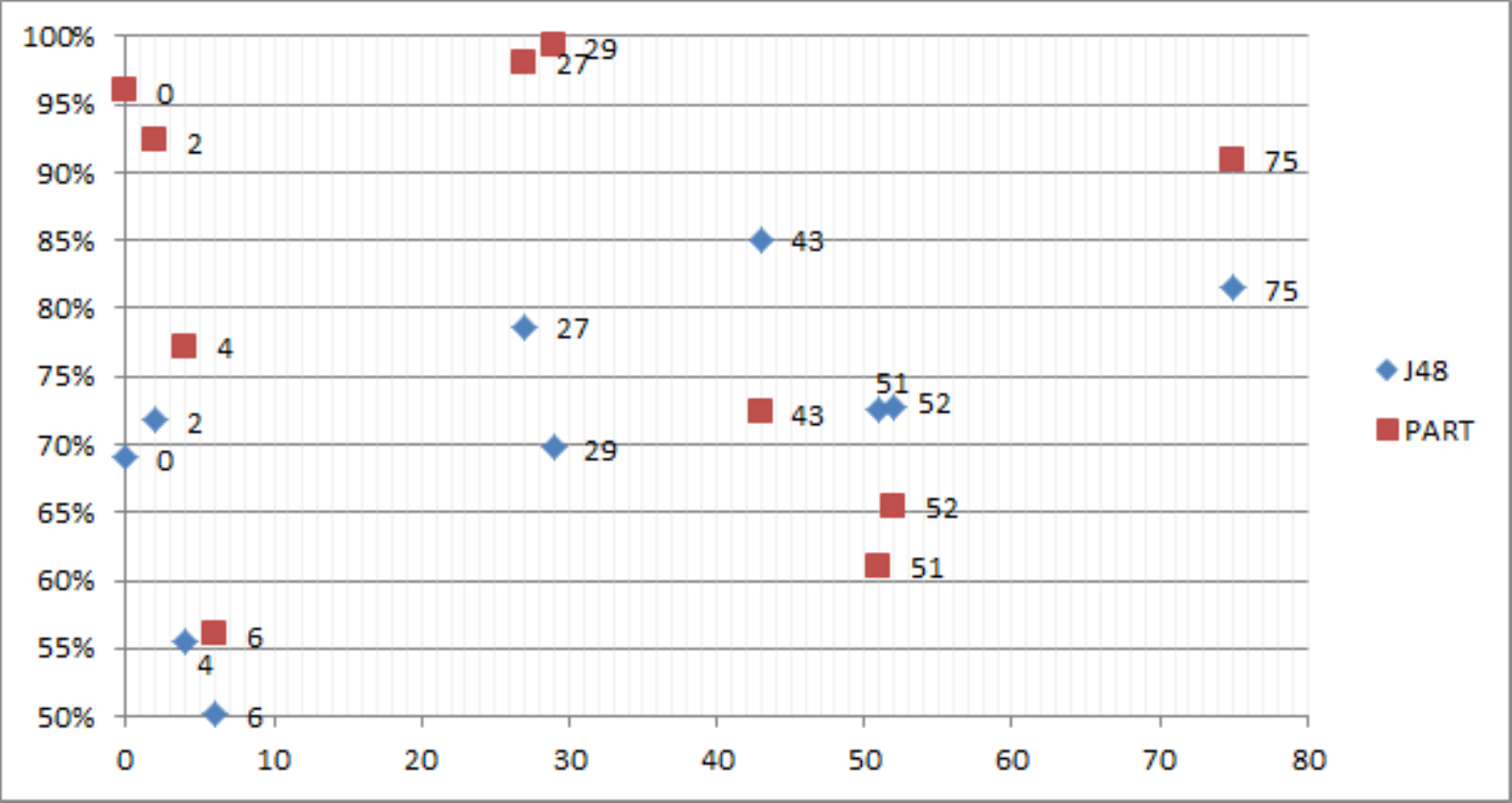}}
\caption{Accuracy of the ten functions that had the biggest difference in accuracy between J48 and PART classifiers.The 10 functions are 0, 2, 4, 6, 27, 29, 43, 51, 52 and 75}
\label{resultsacur}
\end{figure}

\figurename{~\ref{resultsacur}} shows the accuracy of the ten functions that varied the most between the J48 and PART classifiers. We compared the best result of J48 and the best result of PART with the Weka parameter "-M" fixed in 2. The biggest difference happened in circuit 29, with J48 getting 69.74\% of accuracy and PART 99.27\%, resulting in a difference of 29.52\%. 

Most of the functions got similar accuracy for both classifiers. The average accuracy of the J48 classifier was 83.50\%, while the average accuracy of the PART classifier was 84.53\%, a difference of a little over 1\%. After optimizing all the parameters in the Weka tool, we got an average accuracy of 85.73\%. All accuracy values were obtained with cross-validation.

\begin{figure}[htbp]
\centerline{\includegraphics[scale=.5]{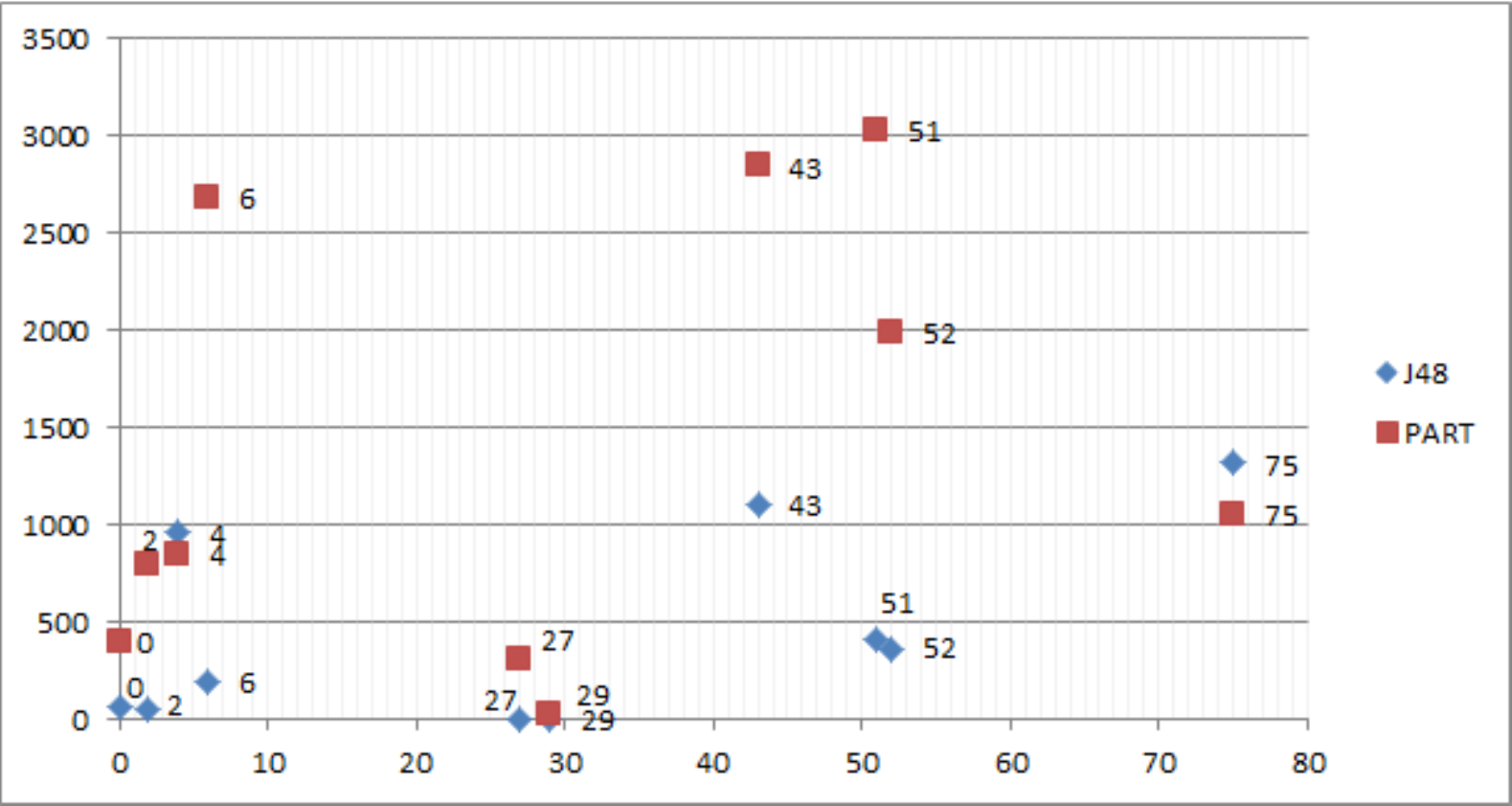}}
\caption{Number of ANDs in ten functions used in~\figurename{~\ref{resultsacur}}.}
\label{resultsands}
\end{figure}

In~\figurename{~\ref{resultsands}}, we compare the number of ANDs in the AIG in the same ten functions. The interesting point observed in this plot refers to circuits 43, 51, and 52. The better accuracy for these circuits is obtained through the J48 classifier, while the resulting AIG is smaller than the ones created from PART solution.  The complementary behavior is observed in circuits 4, and 75. This behavior reinforces the needed for diversification in machine learning classifiers.

%% file: Texfiles_from_collaborator/team3/report.tex
Team 3's solution consists of decision tree (DT) based and neural network (NN) based methods.
For each benchmark, multiple models are generated and 3 are selected for ensemble.

\subsection{DT-based method}
\begin{figure}[ht]
\begin{center}
  \includegraphics[width=0.85\columnwidth]{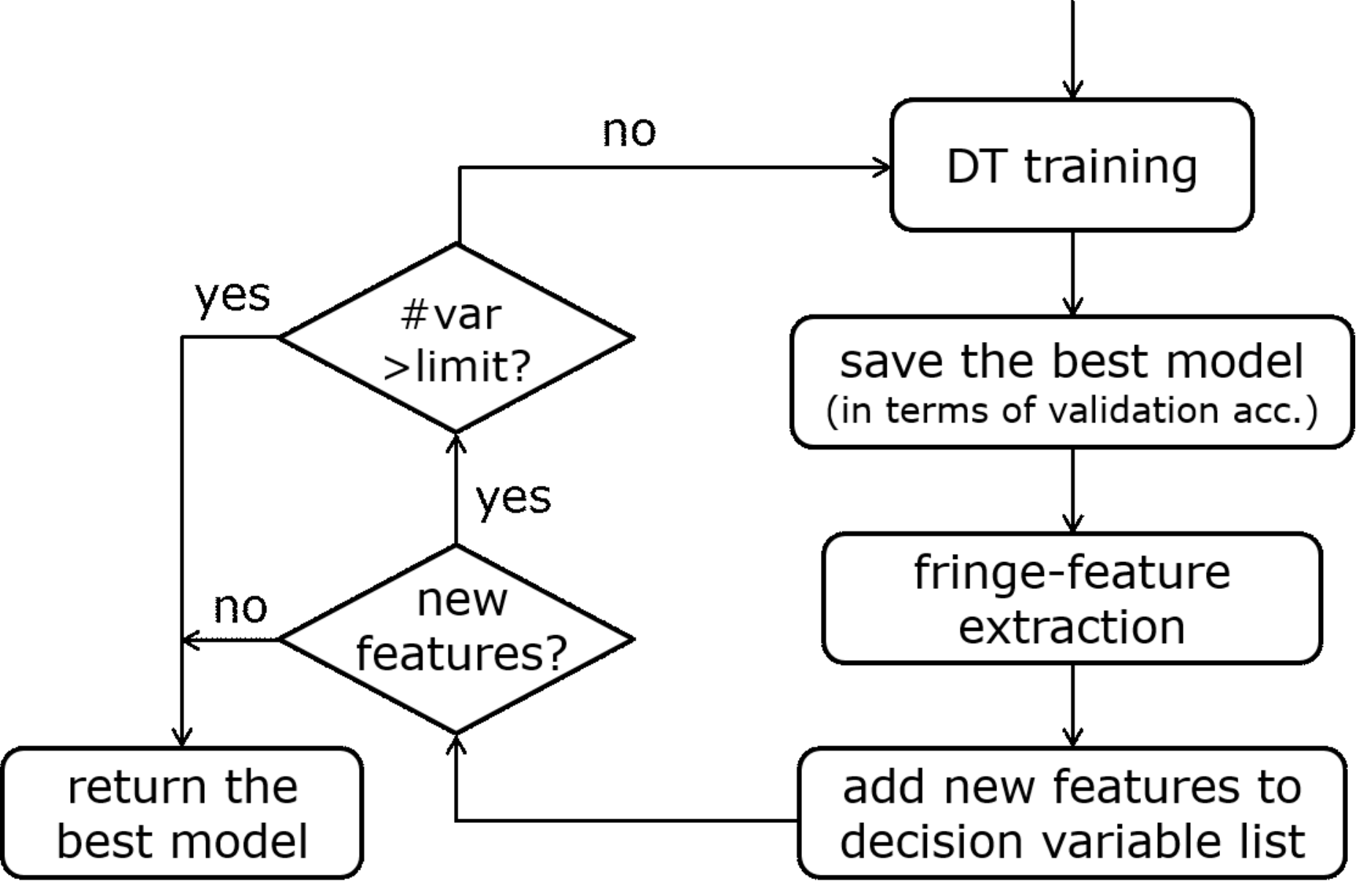}
  \caption{Fringe DT learning flow.}
  \label{fig:dt-flow}
\end{center}
\end{figure}

\begin{figure}[ht]
\begin{center}
  \includegraphics[width=\columnwidth]{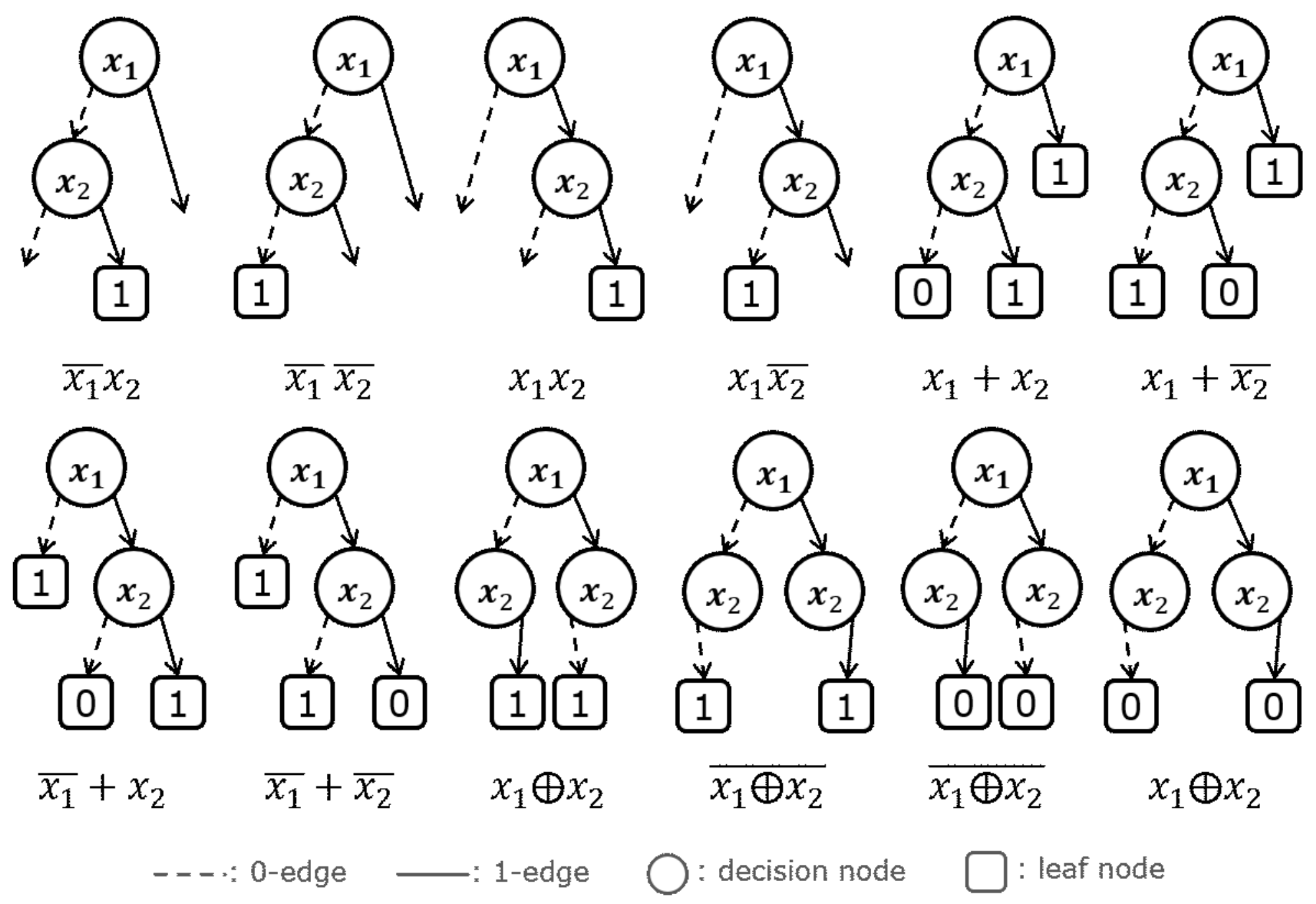}
  \caption{12 fringe patterns.}
  \label{fig:fr-pats}
\end{center}
\end{figure}

For the DT-based method, the fringe feature extraction process proposed in \cite{Pagallo:1990, Arlindo:1993} is adopted.
The overall learning procedure is depicted in ~\figurename{~\ref{fig:dt-flow}}.
The DT is trained and modified for multiple iterations.
In each iteration, the patterns near the fringes (leave nodes) of the DT are identified as the composite features of 2 decision variables.
As illustrated in~\figurename{~\ref{fig:fr-pats}}, 12 different fringe patterns can be recognized, each of which is the combination of 2 decision variables under various Boolean operations.
These newly detected features are then added to the list of decision variables for the DT training in the next iteration.
The procedure terminates when there are no new features found or the number of the extracted features exceeds the preset limit.
After training, the DT model can be synthesized into a MUX-tree in a straightforward manner, which will not be covered in detail in the paper.

\subsection{NN-based method}
\begin{figure}[ht]
\begin{center}
  \includegraphics[width=\columnwidth]{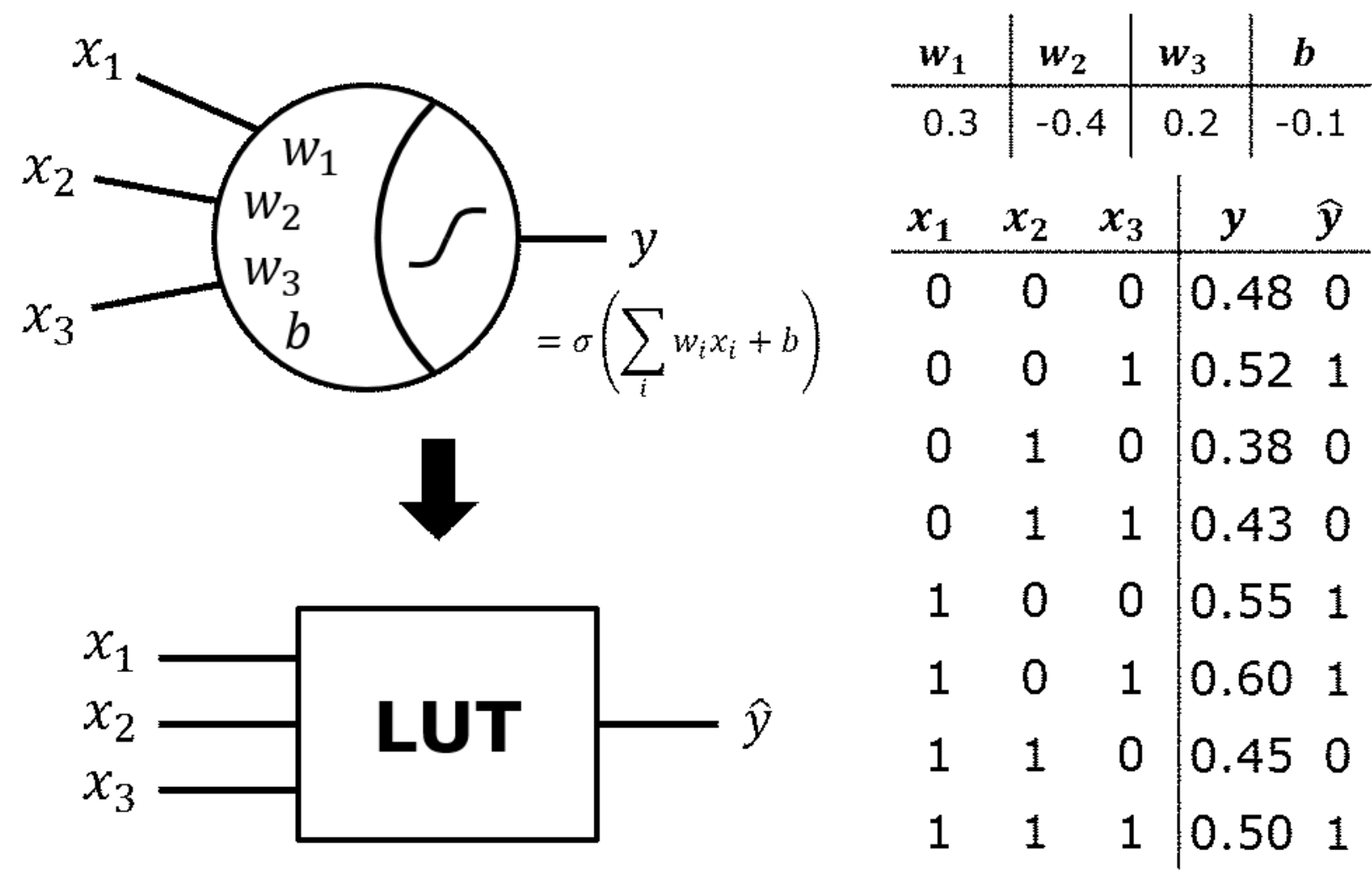}
  \caption{Neuron to LUT transformation.}
  \label{fig:nn-lut}
\end{center}
\end{figure}

For the NN-based method, a 3-layer network is employed, where each layer is fully-connected and uses $sigmoid (\sigma)$ as the activation function.
As the synthesized circuit size of a typical NN could be quite large, the connection pruning technique proposed in \cite{Han:2015} is adopted in order to meet the stringent size restriction.
Network pruning is an iterative process.
In each iteration, a portion of unimportant connections (the ones with weights close to 0) are discarded and the network is then retrained to recover its accuracy.
The NN is pruned until the number of fanins of each neuron is at most 12.
To synthesize the network into a logic circuit, an alternative can be done by utilizing the arithmetic modules, such as adders and multipliers, for the intended computation.
Nevertheless, the synthesized circuit size can easily exceed the limit due to the high complexity of the arithmetic units.
Instead, each neuron in the NN is converted into a LUT by rounding and quantizing its activation.
~\figurename{~\ref{fig:nn-lut}} shows an example transformation of a neuron into a LUT, where all possible input assignments are enumerated, and the neuron output is quantized to 0 or 1 as the LUT output under each assignment.
The synthesis of the network can be done quite fast, despite the exponential growth of the enumeration step, since the number of fanins of each neuron has been restricted to a reasonable size during the previous pruning step.
The resulting NN after pruning and synthesis has a structure similar to the LUT-network in~\cite{chatterjee2018learning}, where, however, the connections were assigned randomly instead of learned iteratively.

\subsection{Model ensemble}
The overall dataset, training and validation set combined, of each benchmark is re-divided into 3 partitions before training.
Two partitions are selected as the new training set, and the remaining one as the new validation set, resulting in 3 different grouping configurations.
Under each configuration, multiple models are trained with different methods and hyper-parameters, and the one with the highest validation accuracy is chosen for ensemble.
Therefore, the obtained circuit is a voting ensemble of 3 distinct models.
If the circuit size exceeds the limit, the largest model is then removed and re-selected from its corresponding configuration.

\subsection{Experimental results}
The DT-based and NN-based methods were implemented with \texttt{scikit-learn}~\cite{sklearn} and \texttt{PyTorch}~\cite{pytorch}, respectively.
After synthesis of each model, the circuit was then passed down to \texttt{ABC}~\cite{abc} for optimization.
Both methods were evaluated on the 100 benchmarks provided by the contest.

\begin{figure}[ht]
\begin{center}
  \includegraphics[width=\columnwidth]{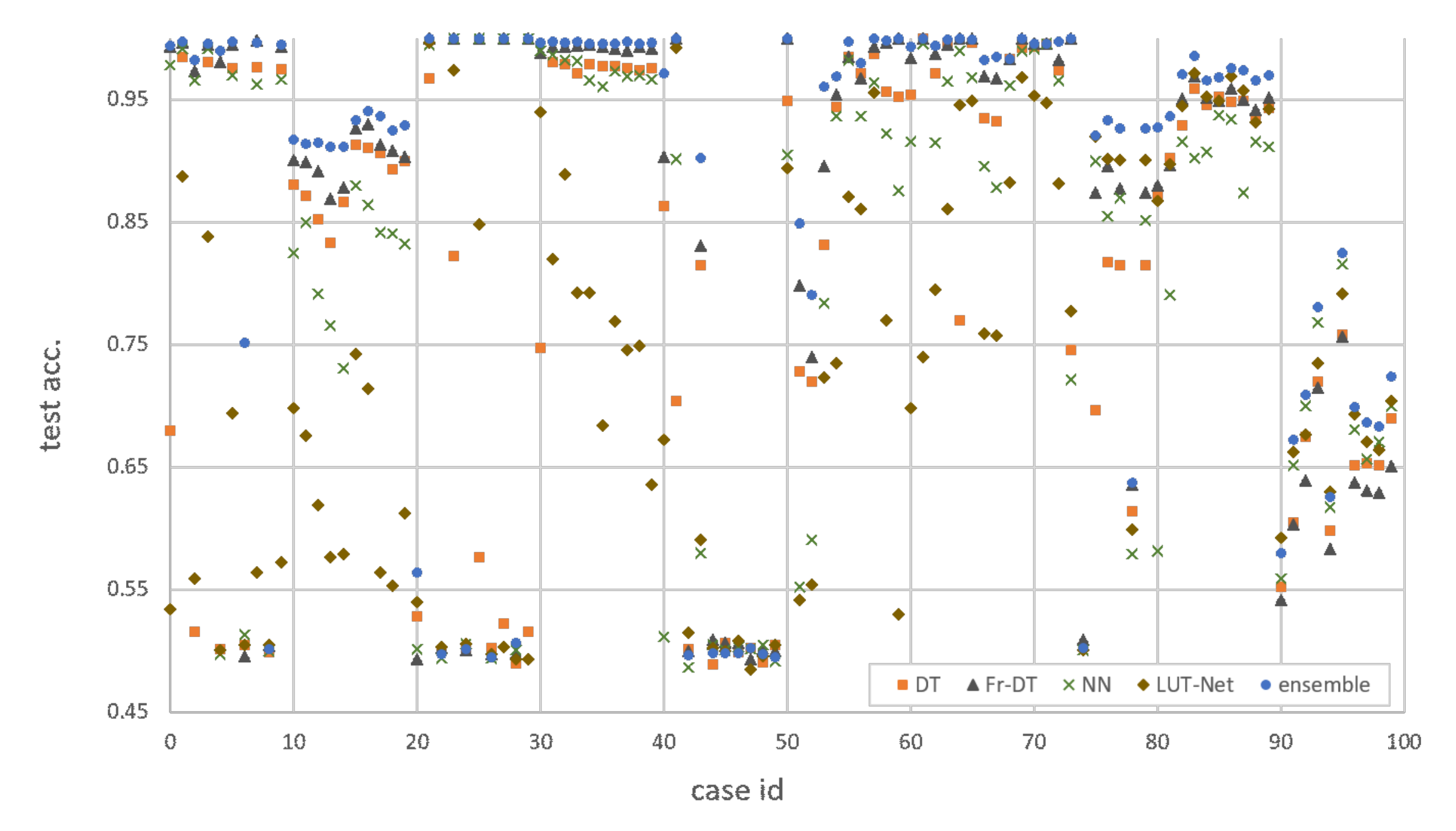}
  \caption{Test accuracy of each benchmark by different methods.}
  \label{fig:acc}
\end{center}
\end{figure}

\begin{figure}[ht]
\begin{center}
  \includegraphics[width=\columnwidth]{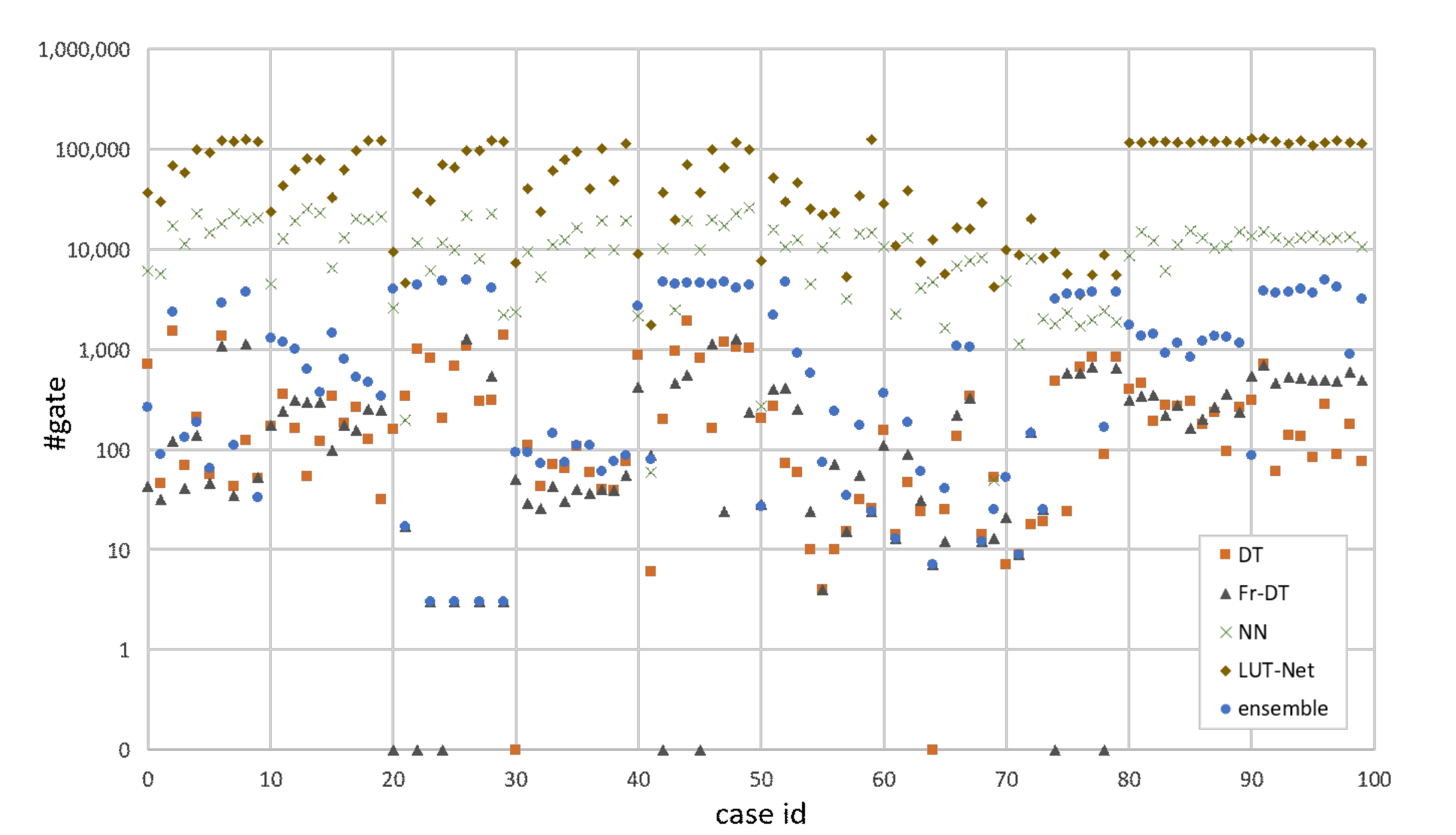}
  \caption{Circuit size of each benchmark by different methods.}
  \label{fig:size}
\end{center}
\end{figure}

\tablename{~\ref{tab:sum}} summarizes the experimental results.
The first column lists various methods under examination, where \textit{Fr-DT} and \textit{DT} correspond to the DT-based method with or without fringe feature extraction, \textit{NN} correspond to the NN-based method,  \textit{LUT-Net} is the learning procedure proposed in \cite{chatterjee2018learning}, and \textit{ensemble} is the combination of \textit{Fr-DT}, \textit{DT} and \textit{NN}.
The remaining columns of~\tablename{~\ref{tab:sum}} specify the average training, validation and testing accuracies along with the circuit sizes (in terms of the number of AIG nodes) of the 100 benchmarks.
~\figurename{~\ref{fig:acc} and~\ref{fig:size}} plot the testing accuracy and circuit size of each case, where different methods are marked in different colors.

\begin{table}[ht]
\centering
\caption{Summary of experimental results.}
\label{tab:sum}
\fontsize{6.5}{7.8}\selectfont
\def\arraystretch{1}
\resizebox{\linewidth}{!}{
\begin{tabular}{|c|c|c|c|r|}
\hline\hline
method   & avg. train acc. & avg. valid acc. & avg. test acc. & \multicolumn{1}{c|}{avg. size} \\ \hline
\textit{DT}       & 90.41\%         & 80.33\%         & 80.15\%        & 303.90                        \\
\textit{Fr-DT}    & 92.47\%         & 85.37\%         & 85.23\%        & 241.47                        \\
\textit{NN}       & 82.64\%         & 80.91\%         & 80.90\%        & 10981.38                      \\
\textit{LUT-Net}~\cite{chatterjee2018learning}  & 98.37\%         & 72.78\%         & 72.68\%        & 64004.39                      \\\hline
\textit{ensemble} & -               & -               & 87.25\%        & 1550.33                      \\ \hline\hline
\end{tabular}}
\end{table}

\textit{Fr-DT} performed the best of the 4 methods under comparison.
It increased the average testing accuracy by over 5\% when compared to the ordinary \textit{DT}, and even attained circuit size reduction.
\textit{Fr-DT} could successfully identify and discover the important features of each benchmark.
Therefore, by adding the composite features into the variable list, more branching options are provided and thus the tree has the potential to create a better split during the training step.
On the other hand, even though \textit{NN} could achieved a slightly higher accuracy than \textit{DT} on average, its circuit size exceeded the limit in 75 cases, which is undesirable.
When comparing \textit{NN} to \textit{LUT-Net}, which was built in a way so that it had the same number of LUTs and average number of connections as \textit{NN}, \textit{NN} clearly has an edge.
The major difference of \textit{NN} and \textit{LUT-Net} exists in the way they connect LUTs from consecutive layers, the former learn the connections iteratively from a fully-connected network, whereas the latter assign them randomly.
Moreover,~\tablename{~\ref{tab:nn}} shows the accuracy degradation of \textit{NN} after connection pruning and neuron-to-LUT conversion.
It remains future work to mitigate this non-negligible $\sim$2\% accuracy drop.
Finally, by ensemble, models with the highest testing accuracies could be obtained.

\begin{table}[ht]
\centering
\caption{Accuracy degradation of \textit{NN} after pruning and synthesis.}
\label{tab:nn}
\begin{tabular}{lrrr}
\toprule
\textit{NN} config. & avg. train acc. & avg. valid acc. & avg. test acc. \\ \midrule
initial         & 87.30\%         & 83.14\%         & 82.87\%        \\
after pruning   & 89.06\%         & 82.60\%         & 81.88\%        \\
after synthesis & 82.64\%         & 80.91\%         & 80.90\%        \\ \bottomrule
\end{tabular}
\end{table}

Of the 300 selected models during ensemble, \textit{Fr-DT} and ordinary \textit{DT} account for 80.3\% and 16.0\%, respectively, with \textit{NN} taking up the remaining 3.7\%. 
As expected, the best-performing \textit{Fr-DT} models are in the majority.
It seems that the DT-based method is better-suited for this learning task.
However, there were several cases, such as case 75 where \textit{NN} achieved 89.97\% testing accuracy over \textit{Fr-DT}'s 87.38\%, with an acceptable circuit size (2320 AIG nodes).

\subsection{Take-Away}
Team 3 adopted DT-based and NN-based methods to tackle the problem of learning an unknown Boolean function. 
The team ranked $4$ in terms of testing accuracy among all the contestants of the IWLS 2020 programming contest.
From the experimental evaluation, the approach that utilized decision tree with fringe feature extraction could achieve the highest accuracy with the lowest circuit size in average.
This approach is well-suited for this problem and can generate a good solution for almost every benchmark, regardless of its origin.

%
%

%% file: Texfiles_from_collaborator/team4/main.tex



%

\subsection{Deep-Learning-Based Boolean Function Approximation}
\label{sec:Introduction}
\begin{figure}[h]
    \centering
    \includegraphics[width=0.35\textwidth]{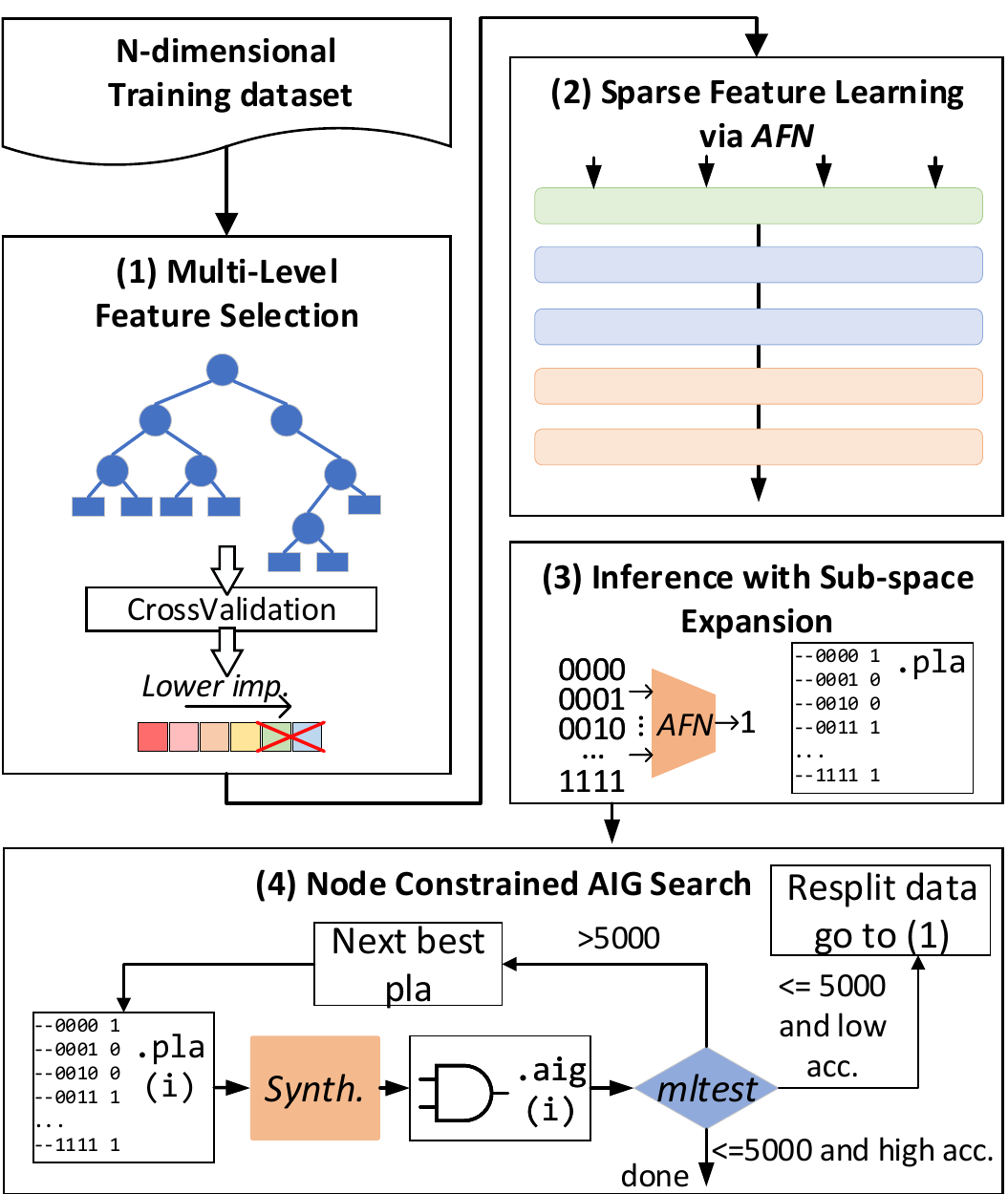}
    \caption{Deep-learning-based Boolean function approximation framework.}
    \label{fig:Flow}
\end{figure}

Given the universal approximation theorem, neural networks can be used to fit arbitrarily complex functions.
For example, multi-layer perceptrons (MLPs) with enough width and depth are capable to learn the most non-smooth binary function in the high-dimensional hypercube, i.e., XNOR or XOR~\cite{NN_NN1989_Hornik}.
For this contest, we adopt a deep-learning-based method to learn an unknown high-dimensional Boolean function with 5k node constraints and extremely-limited training data, formulated as follows,
\begin{align}
    \small
    \label{eq:Formulation}
        \min&~\mathcal{L}(W;\mathcal{D}^{trn}),\\
        \text{s.t.}~~&\mathcal{N}(\texttt{AIG}(W))\leq5,000, \notag
\end{align}
where $\mathcal{L}(W,\mathcal{D}^{trn})$ is the binary classification loss function on the training set, $\mathcal{N}(\texttt{AIG}(W))$ is the number of node of the synthesized AIG representation based on the learned model.
To solve this constrained stochastic optimization, we adopt the following techniques, 1) multi-level ensemble-based feature selection, 2) recommendation-network-based model training, 3) subspace-expansion-based prediction, and 4) accuracy-node joint exploration during synthesis.
The framework is shown in~\figurename{~\ref{fig:Flow}}.
We demonstrate the test result and give an analysis to show our performance on the 100 public benchmarks.

\subsection{Feature Selection with Multi-Level Model Ensemble}
\label{sec:FeatureSelection}
\begin{figure}[t]
    \centering
    \includegraphics[width=0.48\textwidth]{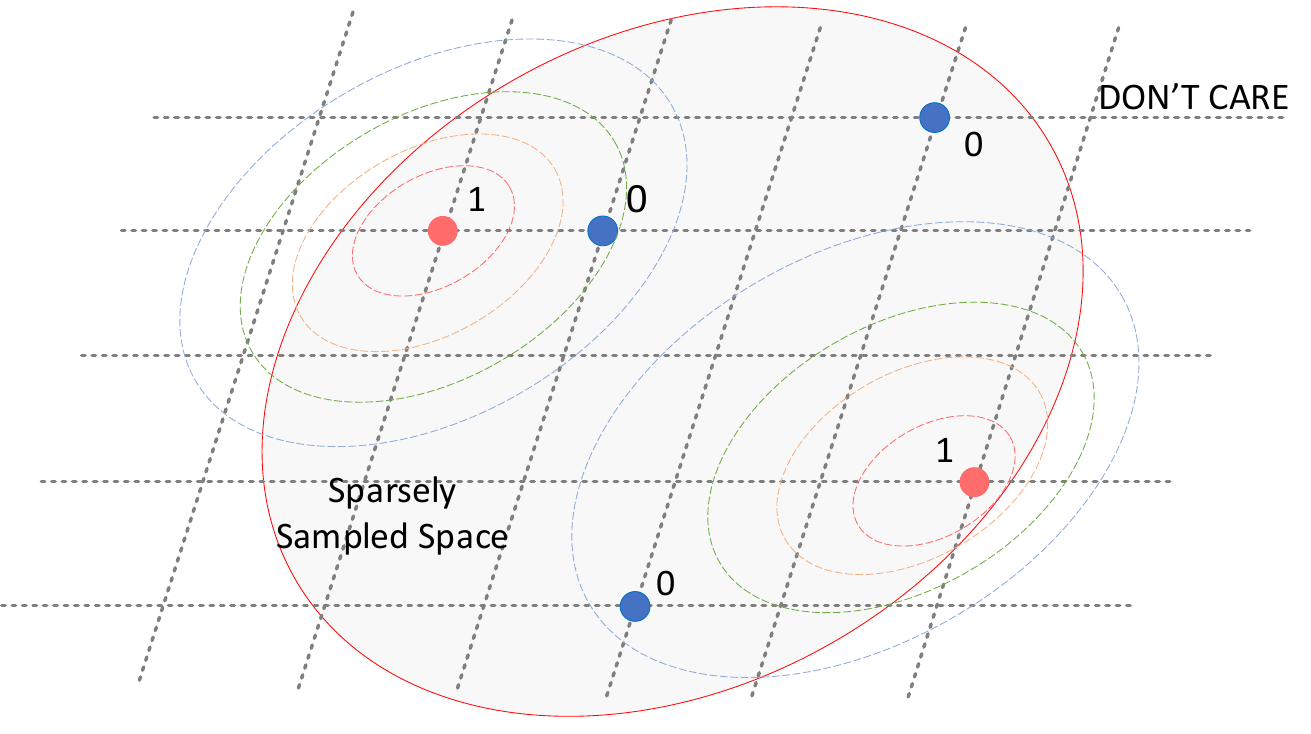}
    \caption{Smoothness assumption in the sparse high-dimensional Boolean space.}
    \label{fig:HypercubeSpace}
    \vspace{-10pt}
\end{figure}

The first critical step for this learning task is to perform data pre-processing and feature engineering. 
The public 100 benchmarks have very unbalanced input dimensions, ranging from 10 to over 700, but the training dataset merely has 6,400 examples per benchmark, which gives an inadequate sampling of the true distribution.
We also observe that a naive AIG representation directly synthesized from the given product terms has orders-of-magnitude more AND gates than the 5,000 node constraint.
The number of AND gate after node optimization of the synthesizer highly depends on the selection of don't-care set.
Therefore, we make a smoothness assumption in the sparse high-dimensional hypercube that any binary input combinations that are not explicitly described in the PLA file are set to don't-care state, such that the synthesizer has enough optimization space to cut down the circuit scale, shown in~\figurename{~\ref{fig:HypercubeSpace}}.

Based on this smoothness assumption, we perform input dimension reduction to prune the Boolean space by multi-level feature selection.
Since we have 6,400 randomly sampled examples for each benchmark, we assume the training set is enough to recover the true functionality of circuits with less than $\floor{\log_2{6,400}}=12$ inputs and do not perform dimension reduction on those benchmarks. For benchmarks with more than 13 inputs, we empirically pre-define a feature dimension $d$ ranging from 10 to 16, which is appropriate to cover enough optimization space under accuracy and node constraints.
For each dimension $d$, we first perform the first level of feature selection by pre-training a machine learning model ensemble.

Given the good interpretability and generalization, traditional machine learning models are widely used in feature engineering to evaluate the feature importance.
We pre-train an \texttt{AdaBoost}~\cite{NN_JCSS1997_Freund} ensemble classifier with 100 \texttt{ExtraTree} sub-classifier on the training set to generate the importance score for all features.
Then we perform permutation importance ranking~\cite{NN_ML2001_Breiman} for 10 times to select the top-$d$ important features as the \emph{care set} variables $F^1(d)$.
Given that the ultimate accuracy is sensitive to the feature selection, we generate another feature group at the second level to expand the search space.
At the second level, we train two classifier ensembles, one is an \texttt{XGB} classifier with 200 sub-trees and another is a 100-\texttt{ExtraTree} based \texttt{AdaBoost} classifier.
Besides, a stratified 10-fold cross-validation is used to select top-$d$ important features $F^2(d)$ based on the average scores from the above two models.
The entire 14 candidates of input feature groups for each benchmark are ${F}=\{F^1(d),F^2(d)\}_{d=10}^{16}$.

\subsubsection{Deep Learning in the Sparse High-Dimensional Boolean Space}
\label{sec:LearnModel}
\begin{figure}
    \centering
    \includegraphics[width=0.49\textwidth]{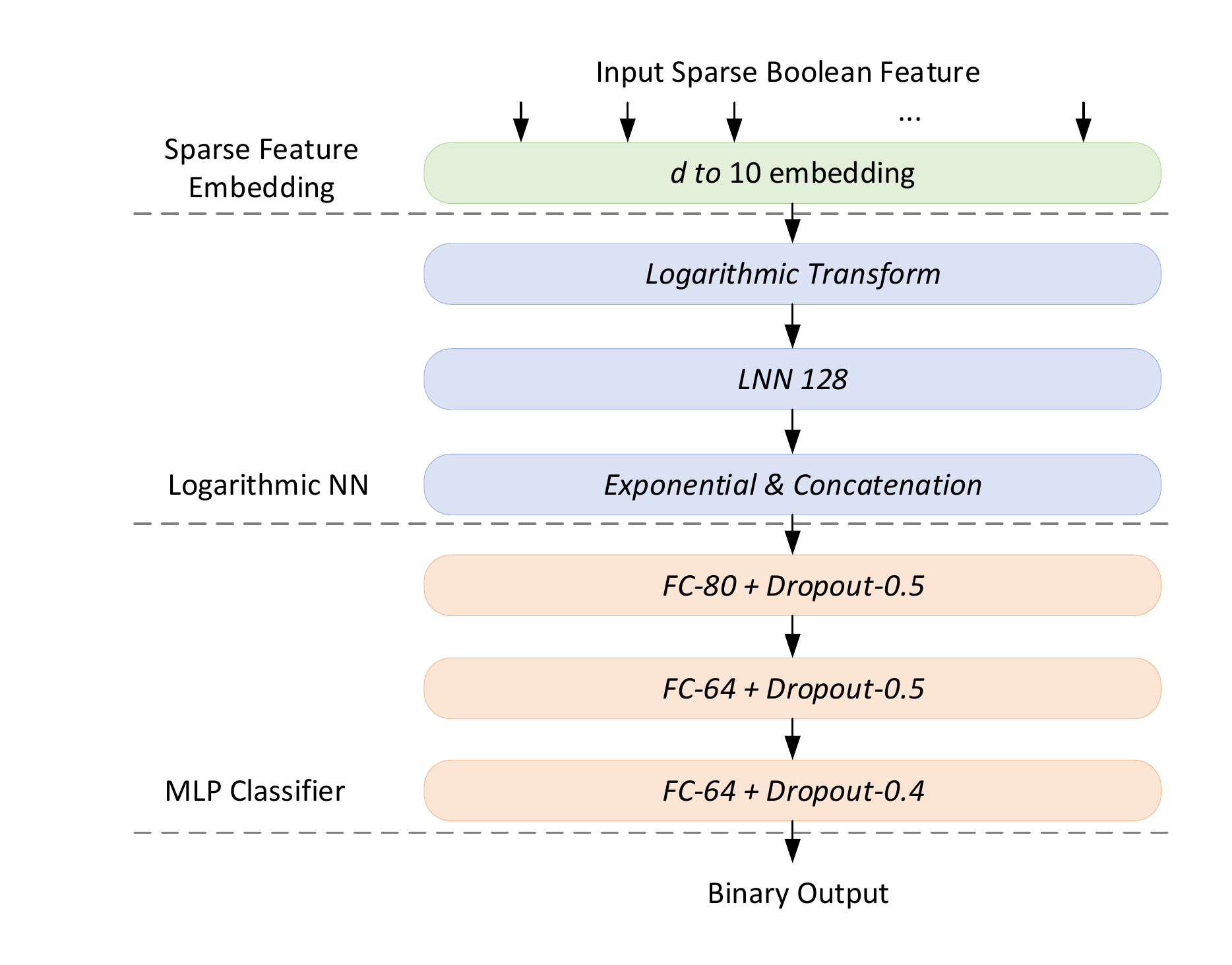}
    \caption{\texttt{AFN}~\cite{NN_AAAI2020_Cheng} and configurations used for Boolean function approximation.}
    \label{fig:AFNArch}
\end{figure}

This learning problem is different from continuous-space learning tasks, e.g., time-sequence-prediction, computer vision-related tasks, since its inputs are binarized with poor smoothness, which means high-frequency patterns in the input features are important to the model prediction, e.g., XNOR and XOR.
Besides, the extremely-limited training set gives an under-sampling of the real distribution, such that a simple multi-layer perceptron is barely capable of fitting the dataset while still having good generalization.
Therefore, motivated by a similar problem, the recommendation system design which targets at predicting the click rate based on dense and sparse features, we adopt a state-of-the-art recommendation model, adaptive factorization network (\texttt{AFN})~\cite{NN_AAAI2020_Cheng}, to fit the sparse Boolean dataset.
~\figurename{~\ref{fig:AFNArch}} demonstrates the \texttt{AFN} structure and network configuration we use to fit the Boolean dataset.
The embedding layer maps the high-dimensional Boolean feature to a 10-$d$ space and transform the sparse feature with a logarithmic transformation layer.
In the logarithmic neural network, multiple vector-wise logarithmic neurons are constructed to represent any cross features to obtain different higher-order input combinations~\cite{NN_AAAI2020_Cheng},
\begin{equation}
    \small
    \label{eq:LogNeuron}
    y_j=\exp\big(\sum_{i=1}^mw_{ij}\ln(\texttt{Embed}(F(d)))\big).
\end{equation}
Three small-scale fully-connected layers are used to combine and transform the crossed features after logarithmic transformation layers.
Dropout layers after each hidden layers are used to during training to improve the generalization on the unknown don't-care set.

\subsubsection{Inference with Sub-Space Expansion}
\label{sec:SubSpaceExpansion}
After training the \texttt{AFN}-based function approximator, we need to generate the dot-product terms in the PLA file and synthesize the AIG representation.
Since we ignore all pruned input dimension, we only assume our model can generalize in the reduced $d$-dimensional hypercube.
Hence, we predict all $2^d$ input combinations with our trained approximator, and set all other pruned inputs to don't-care state.
On 14 different feature groups ${F}$, we trained 14 different models $\{\texttt{AFN}_0,\cdots,\texttt{AFN}_d,\cdots\}$, sorted in a descending order in terms of validation accuracy.
With the above 14 models, we predict 14 corresponding PLA files $\{\mathcal{P}_0,\cdots,{\mathcal{P}_d}.\cdots\}$ with sub-space expansion to maximize the accuracy in our target space while minimizing the node count by pruning all other product terms, shown in~\figurename{~\ref{fig:Flow}}.
In the \texttt{ABC}~\cite{PD_CAV2010_Brayton} tool, we use the node optimization command sequence as \texttt{resyn2}, \texttt{resyn2a}, \texttt{resyn3}, \texttt{resyn2rs}, and \texttt{compress2rs}.

\subsubsection{Accuracy-Node Joint Search with \texttt{ABC}}
\label{sec:Search}
For each benchmark, we obtain multiple predicted PLA files to be selected based on the node constraints.
We search the PLA with the best accuracy that meets the node constraints,
\begin{align}
    \small
    \label{eq:Selection}
    \mathcal{P}^*=\argmax_{\mathcal{P}\sim\{\cdots,\mathcal{P}_d,\cdots\}}&\texttt{Acc}(\texttt{AIG}(\mathcal{P}),\mathcal{D}^{val}),\\
    \text{s.t.}~~ &\mathcal{N}(\texttt{AIG}(\mathcal{P}))\leq5,000. \notag
\end{align}
If the accuracy is still very low, e.g., 60\%, we resplit the dataset $\mathcal{D}^{trn}$ and $\mathcal{D}^{val}$ and go to step (1) again in~\figurename{~\ref{fig:Flow}}.

\subsubsection{Results and Analysis}
\label{sec:Results}
\begin{figure}
    \centering
    \includegraphics[width=0.48\textwidth]{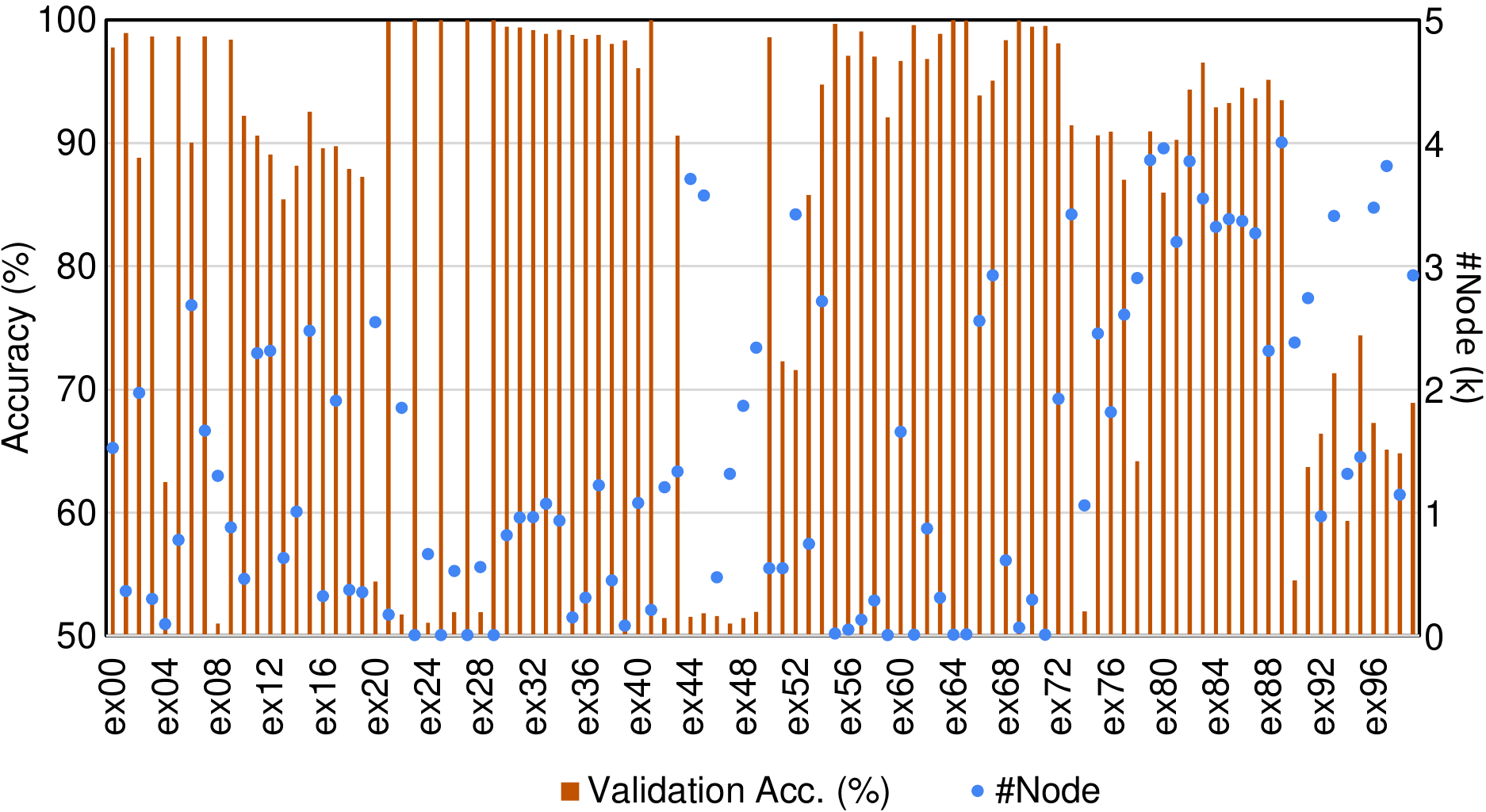}
    \caption{Evaluation results on IWLS 2020 benchmarks.}
    \label{fig:Results}
    \vspace{-10pt}
\end{figure}
\figurename{~\ref{fig:Results}} shows our validation accuracy and number of node after node optimization.
Our model achieves high accuracy on most benchmarks.
While on certain cases regardless of the input count, our model fails to achieve a desired accuracy.
An intuitive explanation is that the feature-pruning-based dimension reduction is sensitive on the feature selection.
A repeating procedure by re-splitting the dataset may help find a good feature combination to improve accuracy.

\subsection{Conclusion and Future Work}
\label{sec:Conclusion}
We introduce the detailed framework we use to learn the high-dimensional unknown Boolean function for IWLS'20 contest.
Our recommendation system based model achieves a top 3 smallest generalization gap on the test set (0.48\%), which is a suitable selection for this task.
A future direction is to combine more networks and explore the unique characteristic of various benchmarks.

%% file: Texfiles_from_collaborator/team5/main.tex
\begin{figure}[ht!]
\centering
\includegraphics[width=1\linewidth]{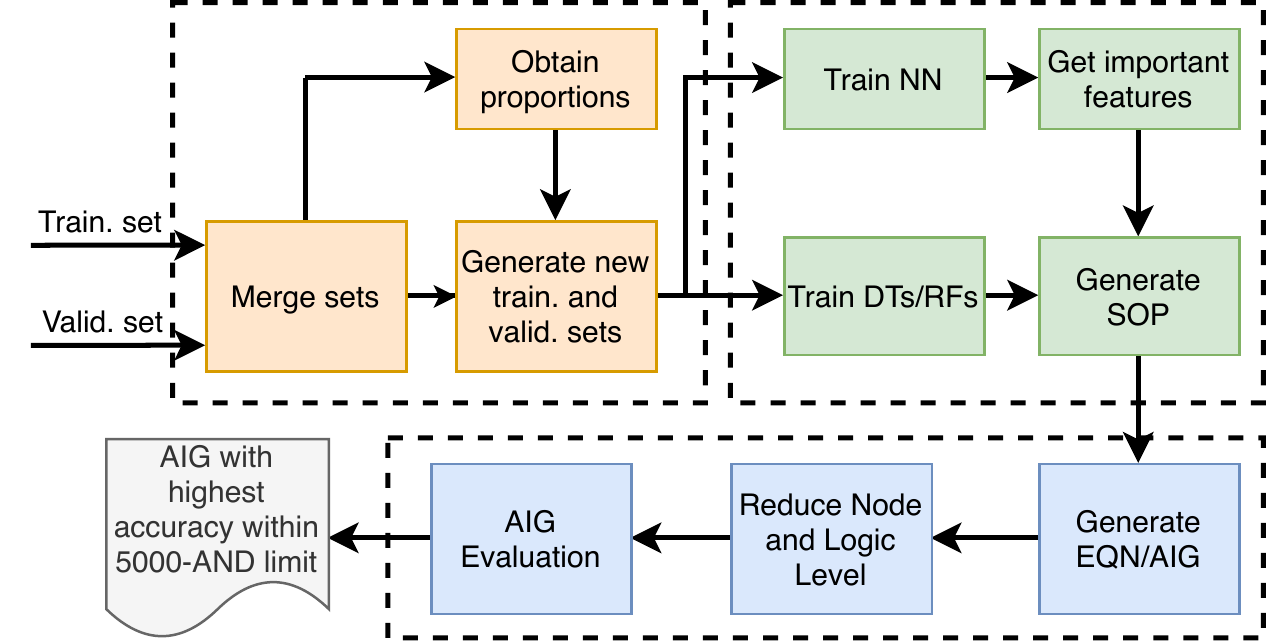}
\caption{Design flow employed in this proposal.}
\label{fig:flow}
\end{figure}

\figurename{~\ref{fig:flow}} presents the process employed in this proposal. In the first stage, the training and valid sets provided in the problem description are merged. The ratios of 0's and 1's in the output of the newly merged set are calculated, and we split the set into two new training and validation sets, considering an 80\%-20\% ratio, respectively, preserving the output variable distribution. Additionally, a second training set is generated, equivalent to half of the previously generated training set. These two training sets are used separately in our training process, and their accuracy results are calculated using the same validation set to enhance our search for the best models. This was done because the model with the training set containing 80\% of the entire set could potentially lead to models with overfitting issues, so using another set with 40\% of the entire provided set could serve as an alternative.

After the data sets are prepared, we train the DTs and RFs models. Every decision model in this proposal uses the structures and methods from the Scikit-learn Python library. The DTs and RFs from this library use the Classification and Regression Trees (CART) algorithm to assemble the tree-based decision tools \cite{cart_algorithm}. To train the DT, we use the $DecisionTreeClassifier$ structure, limiting its $max\_depth$ hyper-parameter to values of 10 and 20 due to the 5000-gate limitation of the contest. The RF model was trained similarly, but with an ensemble of $DecisionTreeClassifier$: we opted not to employ the $RandomForestClassifier$ structure given that it applied a weighted average of the preliminary decisions of each DT within it. Considering that this would require the use of multipliers, and this would not be suitable to obtain a Sum-Of-Products (SOP) equivalent expression, using several instances of the $DecisionTreeClassifier$ with a simple majority voter in the output was the choice we adopted. In this case, each DT was trained with a random subset of the total number of features. Based on preliminary testing, we found that RFs could not scale due to the contest's 5000-gate limitation. This was mainly due to the use of the majority voting, considering that the preliminaries expressions for that are too large to be combined between each other. Therefore, for this proposal, we opted to limit the number of trees used in the RFs to a value of three.

Other parameters in the $DecisionTreeClassifier$ could be varied as well, such as the split metric, by changing the Gini metric to Entropy. However, preliminary analyses showed that both metrics led to very similar results. Since Gini was slightly better in most scenarios and is also less computationally expensive, it was chosen.

Even though timing was not a limitation described by the contest, we still had to provide a solution that we could verify that was yielding the same AIGs as the ones submitted. Therefore, for every configuration tested, we had to test every example given in the problem. Hence, even though higher depths could be used without surpassing the 5000-gate limitation, we opted for 10 and 20 only, so that we could evaluate every example in a feasible time.

Besides training the DTs and RFs with varying depths, we also considered using feature selection methods from the Scikit-learn library. We opted for the use of the $SelectKBest$ and $SelectPercentile$ methods. These methods perform a pre-processing in the features, eliminating some of them prior to the training stage. The $SelectKBest$ method selects features according to the $k$ highest scores, based on a score function, namely $f\_classif$, $mutual\_info\_classif$ or $chi2$, according to the Scikit-learn library~\cite{sklearn}. The $SelectPercentile$ is similar but selects features within a percentile range given as parameter, based on the same score functions~\cite{sklearn}. We used the values of 0.25, 0.5, and 0.75 for $k$, and the percentiles considered were 25\%, 50\%, and 75\%.

The solution employing neural networks (NNs) was considered after obtaining the accuracy results for the DTs/RFs configurations. In this case, we train the model using the $MLPClassifier$ structure, to which we used the default values for every parameter. Considering that NNs present an activation function in the output, which is non-linear, the translation to a SOP would not be possible using conventional NNs. Therefore, this solution only uses the NNs to obtain a subset of features based on their importance, i.e., select the set of features with the corresponding highest weights. With this subset of features obtained, we evaluate combinations of functions, using "OR," "AND," "XOR," and "NOT" operations among them. Due to the fact that the combinations of functions would not scale well, in terms of time, with the number of features, we limit the sub-set to contain only four features. The number of expressions evaluated for each NN model trained was 792. This part of the proposal was mainly considered given the difficulty of DTs/RFs in finding trivial solutions for XOR problems. Despite solving the problems from specific examples whose solution was a XOR2 between two of the inputs, with a 100\% accuracy, we were able to slightly increase the maximum accuracy of other examples through this scan of functions. The parameters used by the $MLPClassifier$ were the default ones: 100 hidden layers and ReLu activation function~\cite{sklearn}.

The translation from the DT/RF to SOP was implemented as follows: the program recursively passes through every tree path, concatenating every comparison. When it moves to the left child, this is equivalent to a "true" result in the comparison. In the right child, we need a "NOT" operator in the expression as well. In a single path to a leaf node, the comparisons are joined through an "AND" operation, given that the leaf node result will only be true when all the comparisons conditions are true. However, given that this is a binary problem, we only consider the "AND" expression of a path when the leaf leads to a value of 1. After that, we perform an "OR" operation between the "AND" expressions obtained for each path, which yields the final expression of that DT.

The RF scenario works the same way, but it considers the expression required for the majority gate as well, whose inputs are the "OR" expressions of each DT.

From the SOP, we obtain the AIG file. The generated AIG file is then optimized using commands of the ABC tool \cite{abc}  attempting to reduce the number of AIG nodes and the number of logic levels by performing iterative collapsing and refactoring of logic cones in the AIG, and rewrite operations.
Even if these commands could be iteratively executed, we decided to run them only once given that the 5000-gate limitation was not a significant issue for our best solutions, and a single sequence of them was enough for our solutions to adhere to the restriction.

Finally, we run the AIG file using the AIG evaluation commands provided by ABC to collect the desired results of accuracy, number of nodes, and number of logic levels for both validation sets generated at the beginning of our flow. 

All experiments were performed three times with different seed values using the Numpy random seed method. This was necessary given that the $DecisionTreeClassifier$ structure has a degree of randomness. Considering it was used for both DTs and RFs, this would yield different results at each execution. The $MLPClassifier$ also inserts randomnesses in the initialization of weights and biases. Therefore, to ensure that the contest organizers could perfectly replicate the code with the same generated AIGs, we fixed the seeds to values of 0, 1, and 2. Therefore, we evaluated two classifiers (DTs/RFs), with two maximum depths, two different proportions, and three different seeds, which leads to 24 configurations. For each of the $SelectKBest$ and $SelectPercentile$ methods, considering that we analyzed three values of K and percentage each, respectively, along with three scoring functions, we have a total of 18 feature selection methods. Given that we also tested the models without any feature selection method, we have 19 possible combinations. By multiplying the number of configurations (24) with the number of combinations with and without feature selection methods (adding up to 19), we obtain a total of 456 different DT/RF models being evaluated. Additionally, each NN, as mentioned, evaluated 792 expressions. These were also tested with two different training set proportions and three different seeds, leading to a total of 4752 expressions for each of the 100 examples of the contest.

\tablename{~\ref{tab:summary}} presents some additional information on the configurations that presented the best accuracy for the examples from the contest. As it can be seen, when we split the 100 examples by the decision tool employed, most of them obtained the best accuracy results when using DTs, followed by RFs and NNs, respectively. The use of RFs was not as significant due to the 5000-gate limitation, as it restrained us from using a higher number of trees. The NNs were mainly useful to solve XOR2 problems and in problems whose best accuracy results from DTs/RFs were close to 50\%. It can also be observed that the use of the feature selection methods along with DTs/RFs was helpful; therefore, cropping a sub-set of features based on these methods can definitely improve the model obtained by the classifiers. The number of best examples based on the different scoring functions used in the feature selection methods shows that the chi2 function was the most useful. This is understandable given that this is the default function employed by the Scikit-learn library. Lastly, even though the proportions from 80\%-20\% represented the majority of the best results, it can be seen that running every model with a 40\%-20\% was also a useful approach.

\begin{table}[!ht]
\centering
\caption{Number of best examples based on the characteristics of the configurations.}
\begin{tabular}{c | c | c}
\hline
\textbf{Characteristic} & \textbf{Parameter} & \textbf{\# of examples} \\
\hline
\hline
& DT & 55 \\
\hhline{~--}
& RF & 28 \\
\hhline{~--}
\multirow{-3}{*}{Decision Tool} & NN & 17 \\
\hline\hline
& Select K Best & 48 \\
\hhline{~--}
& Select Percentile & 11 \\
\hhline{~--}
\multirow{-3}{*}{Feature Selection} & None & 41 \\
\hline\hline
& chi2 & 34 \\
\hhline{~--}
& f\_classif & 6 \\
\hhline{~--}
& mutual\_info\_classif & 19 \\
\hhline{~--}
\multirow{-4}{*}{Scoring Function} & None & 41 \\
\hline\hline
& 40\%-20\% & 23 \\
\hhline{~--}
\multirow{-2}{*}{Proportion} & 80\%-20\% & 77 \\
\hline
\end{tabular}
\label{tab:summary}
\end{table}

%% file: Texfiles_from_collaborator/team6/main.tex
To start of with, we read the training pla files using ABC and get the result of $ \& mltest$ directly on the train data. This gives the upper limit of the number of AND gates which are used. Then, in order to learn the unknown Boolean function, we have used the method as mentioned in~\cite{chatterjee2018learning}. We used LUT network as a means to learn from the known set to synthesize on an unknown set.  We use the concept of memorization and construct a logic network. 

In order to construct the LUT network, we use the minterms as input features to construct layers of LUTs with connections starting from the input layer. We then try out two schemes of connections between the layers: `random set of input' and `unique but random set of inputs'. By `random set of inputs', we imply that we just randomly select the outputs of preceding layer and feed it to the next layer. This is the default flow. By `unique but random set of inputs', we mean that we ensure that all outputs from a preceding layer is used before duplication of connection. This obviously makes sense when the number of connections is more than the number of outputs of the preceding layer. 

We have four hyper parameters to experiment with in order to achieve good accuracy-- number of inputs per LUT, number of LUTS per layers, selection of connecting edges from the preceding layer to the next layer and the depth (number of LUT layers) of the model. We carry out experiments with varying number of inputs for each LUT in order to get the maximum accuracy. We notice from our experiments that 4-input LUTs returns the best average numbers across the benchmark suite. We also found that increasing the number of LUTs per layer or number of layers does not directly increases the accuracy. This is due to the fact, because increasing the number of LUTs allows repeatability of connections from the preceding layer. This leads to passing of just zeros/ones to the succeeding layers. Hence, the overall network tends towards a constant zero or constant one.

After coming up with an appropriate set of connections, we create the whole network.
Once we create the network, we convert the network into an SOP form using sympy package in python. This is done from reverse topological order starting from the outputs back to the inputs. We then pass the SOP form of the final output and create a verilog file out of that. Using the verilog file, we convert it into AIG format and then use the ABC command to list out the output of the $\& mltest$. We also report accuracy of our network using sklearn.  

We have also incorporated use of data from validation testcase. For our training model, we have used `0.4' part of the minterms in our training.

%% file: Texfiles_from_collaborator/team7/main.tex
%
\bstctlcite{IEEEexample:BSTcontrol}

Team 7's solution is a mix of conventional machine learning (ML) and pre-defined standard function matching. If the training set matches a pre-defined standard function, a custom AIG of the identified function is written out. Otherwise, an ML model is trained and translated to an AIG.

Team 7 adopts tree-based ML models, considering the straightforward correspondence between tree nodes and SOP terms. The model is either a single decision tree with unlimited depth, or an extreme gradient boosting (XGBoost) \cite{xgboost} of 125 trees with a maximum depth of 5, depending on the results of a 10-fold cross validation on training data.

\begin{figure}
    \centering
    \hfill
    \subfloat[]{\includegraphics[width=0.2\textwidth]{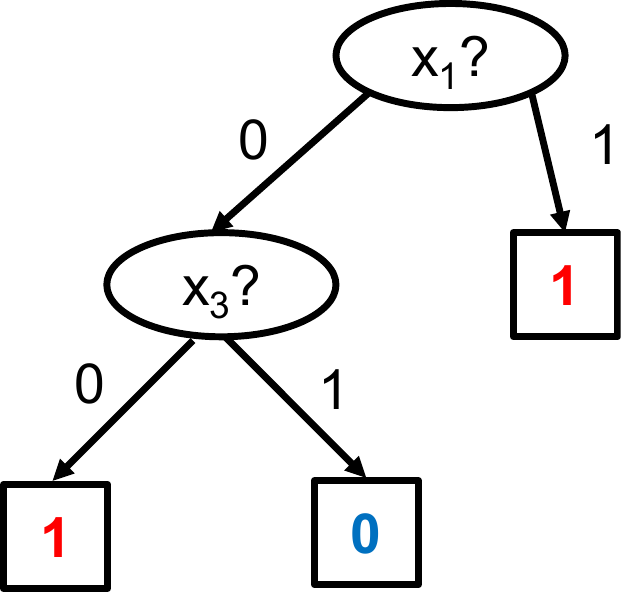}}
    \hfill
    \subfloat[]{\includegraphics[width=0.15\textwidth]{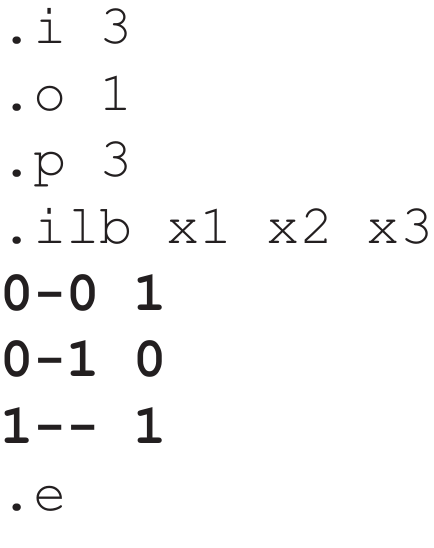}}
    \hfill
    \caption{(a) A decision tree and (b) its corresponding PLA.}
    \label{fig:DT}
\end{figure}

\begin{figure}
    \centering
    \subfloat[]{\includegraphics[width=0.48\textwidth]{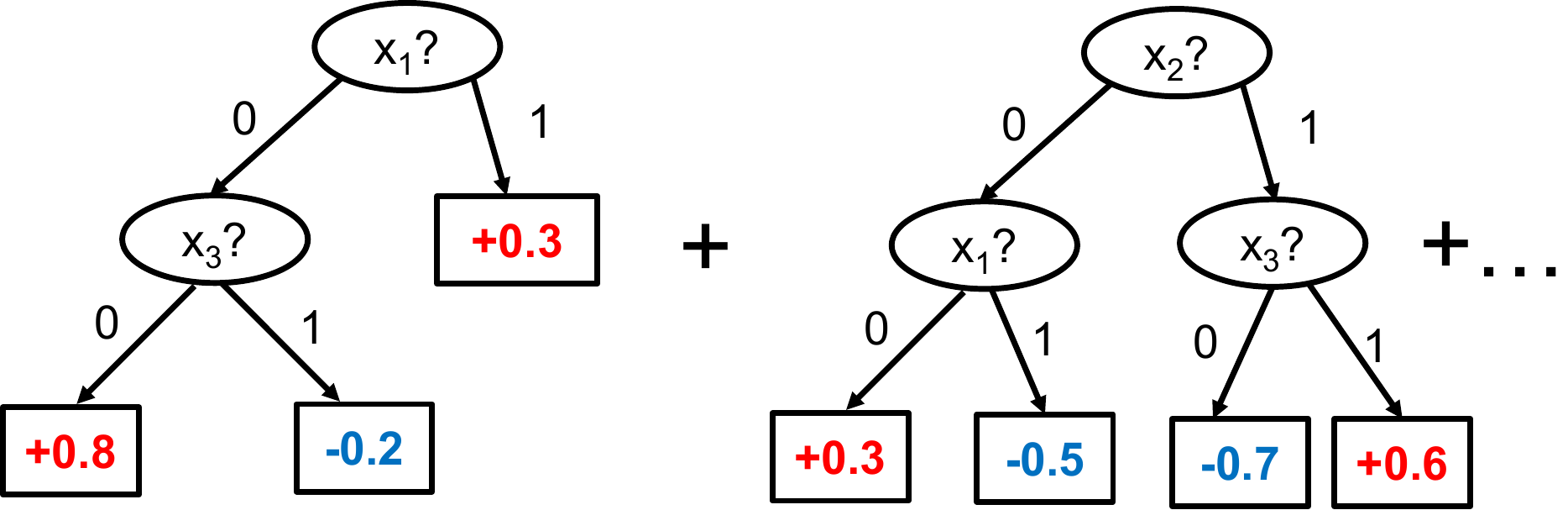}}\\
    \subfloat[]{\includegraphics[width=0.48\textwidth]{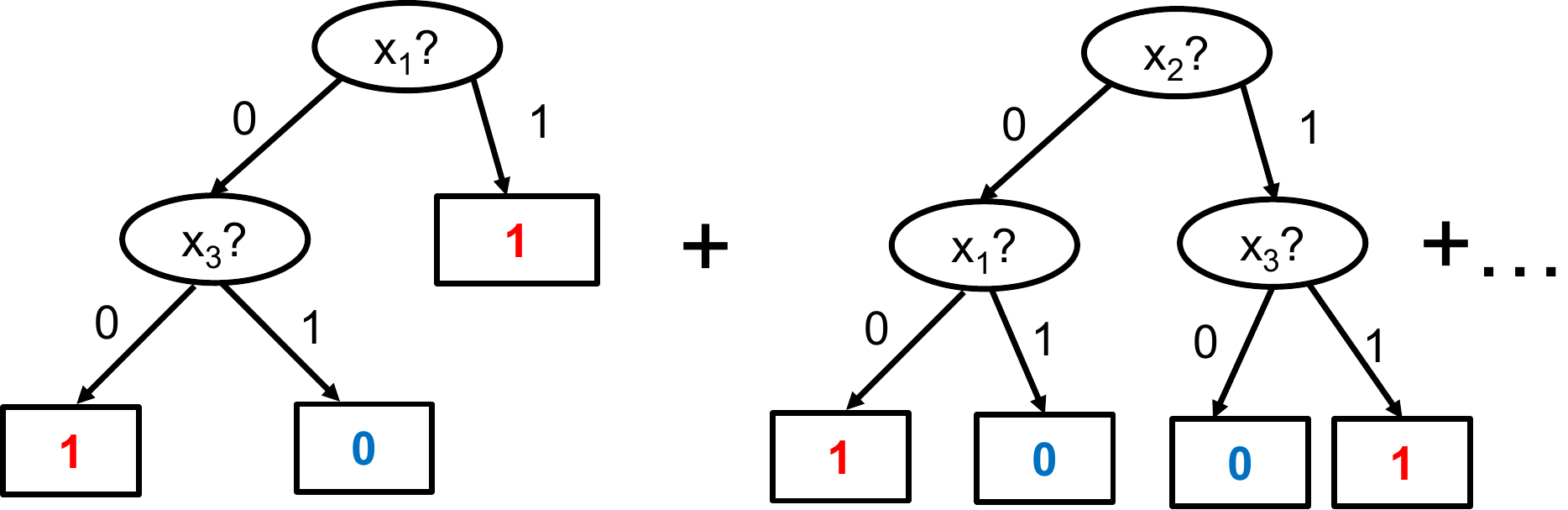}}
    \caption{XGBoost of decision trees, (a) before and (b) after quantization of leaf values. Plus signs mean to add up resulting leaf values, one from each tree.}
    \label{fig:xgboost}
\end{figure}

It is straightforward to convert a decision tree to SOP terms used in PLA.~\figurename{~\ref{fig:DT}} shows a simple example of decision tree and its corresponding SOP terms. Numbers in leaves (rectangular boxes) indicate the predicted output values. XGBoost is a more complex model in two aspects. First, it is a boosting ensemble of many shallow decision trees, where the final output is the sum of all leaf values that a testing data point falls into, expressed in log odds $\log[P(y=1)/P(y=0)]$. Second, since the output is a real number instead of a binary result, a threshold of classification is needed to get the binary prediction (the default value is $0$).~\figurename{~\ref{fig:xgboost}}(a) shows a simple example of a trained XGBoost of trees.

\begin{figure}
    \centering
    \includegraphics[width=0.4\textwidth]{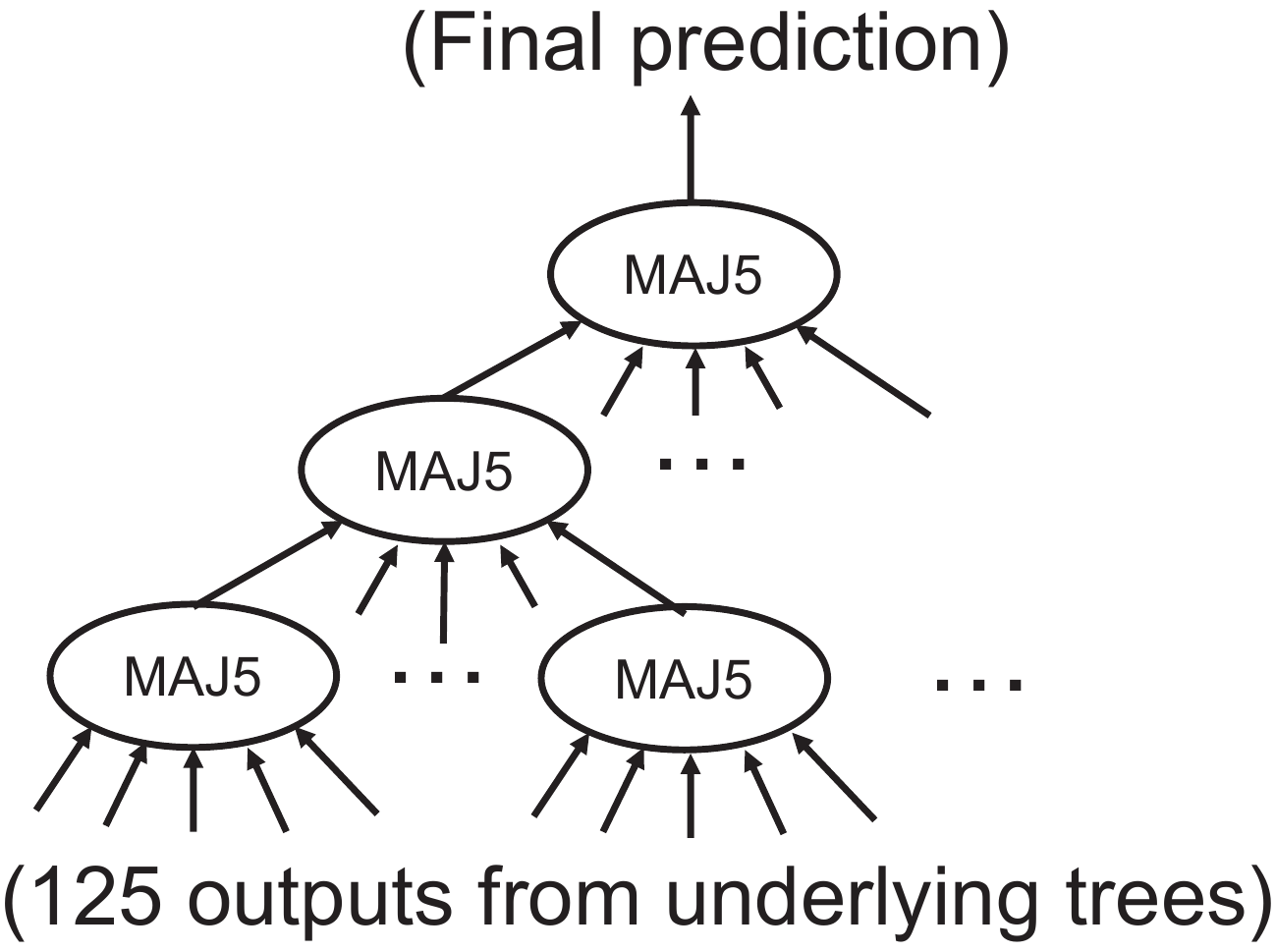}
    \caption{Network of 5-input majority gates as an approximation of a 125-input majority gate.}
    \label{fig:maj}
\end{figure}

With the trained model, each underlying tree is converted to a PLA, where each leaf node in the tree corresponds to a SOP term in the PLA. Each PLA are minimized and compiled to an AIG with the integrated \texttt{espresso} and \texttt{ABC}, respectively. If the function is learned by a single decision tree, the converted AIG is final. If the function is learned by XGBoost of trees, the exact final output of a prediction would be the sum of the 125 leaves (one for each underlying tree) where the testing data point falls into. In order to implement the AIGs efficiently, the model is approximated in the following two steps. First, the value of each underlying tree leaf is quantized to one bit, as shown in~\figurename{~\ref{fig:xgboost}}(b). Each test data point will fall into 125 specific leaves, yielding 125 output bits. To mimic the summation of leaf values and the default threshold of $0$ for classification, a 125-input majority gate could be used to get the final output. However, again for efficient implementation, the 125-input majority gate is further approximated by a 3-layer network of 5-input majority gates as shown in~\figurename{~\ref{fig:maj}}.

\begin{figure}[t]
    \centering
    \begin{subfigure}[b]{\columnwidth}
    \centering
    \scalebox{0.7}{\includegraphics[width=\columnwidth]{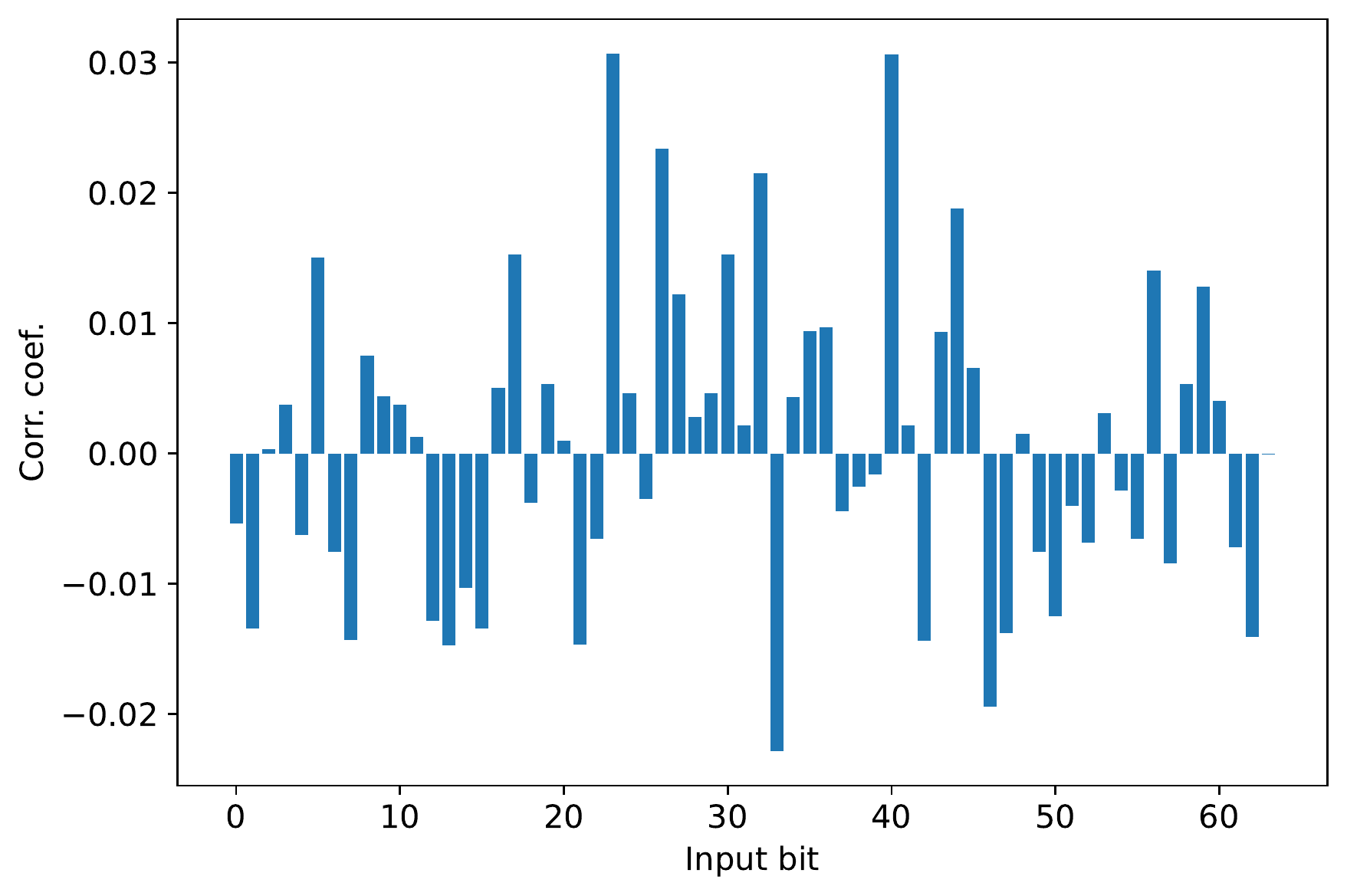}}
    \caption{}
    \end{subfigure}
    \begin{subfigure}[b]{\columnwidth}
    \centering
    \scalebox{0.7}{\includegraphics[width=\columnwidth]{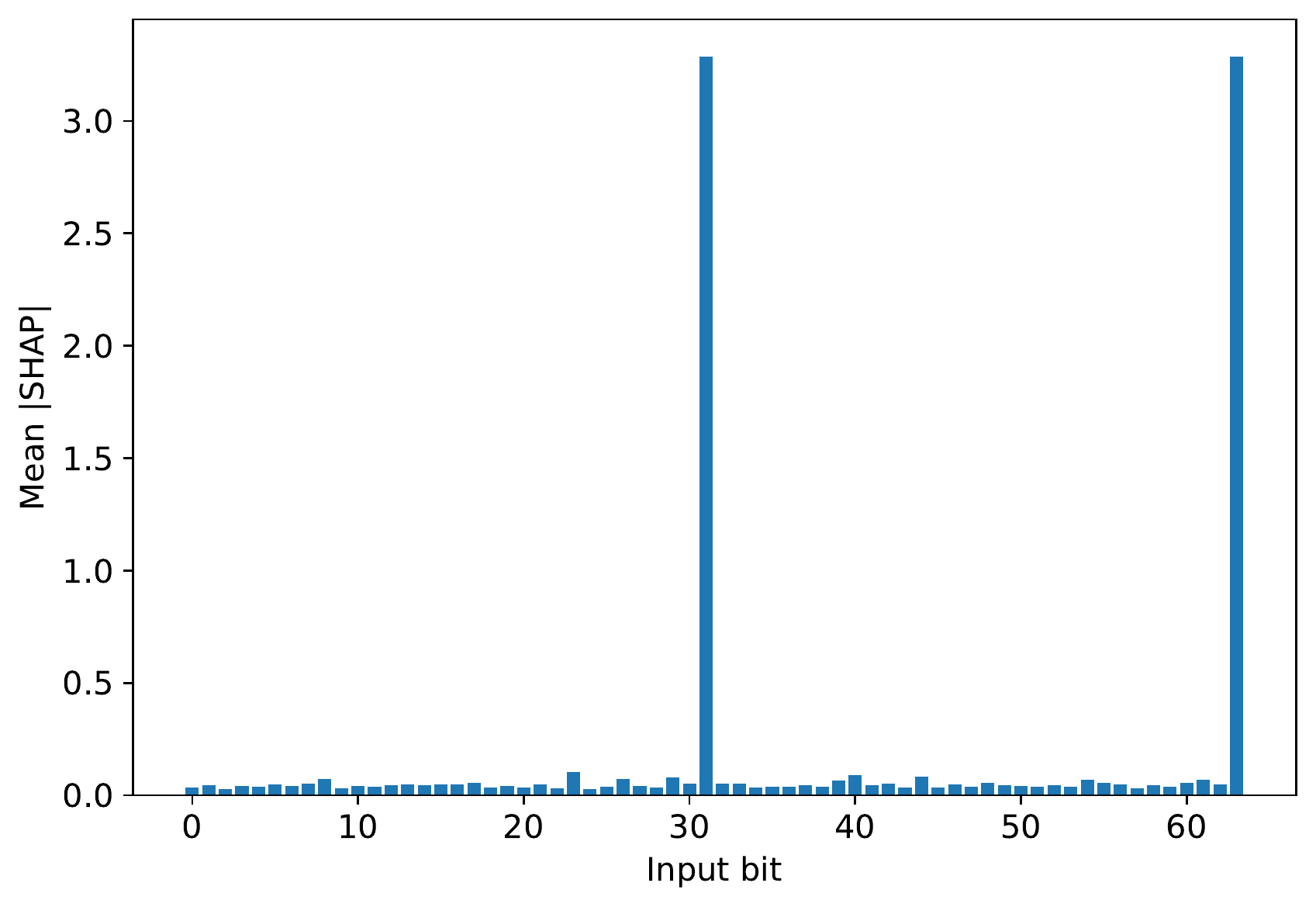}}
    \end{subfigure}
    \caption{Comparison of two importance metrics: (a) correlation coefficient and (b) mean absolute SHAP value in ex25, the MSB of a 32x32 multiplier.}
    \label{fig:ex25}
\end{figure}

\begin{figure}[t]
    \centering
    \includegraphics[width=0.48\textwidth]{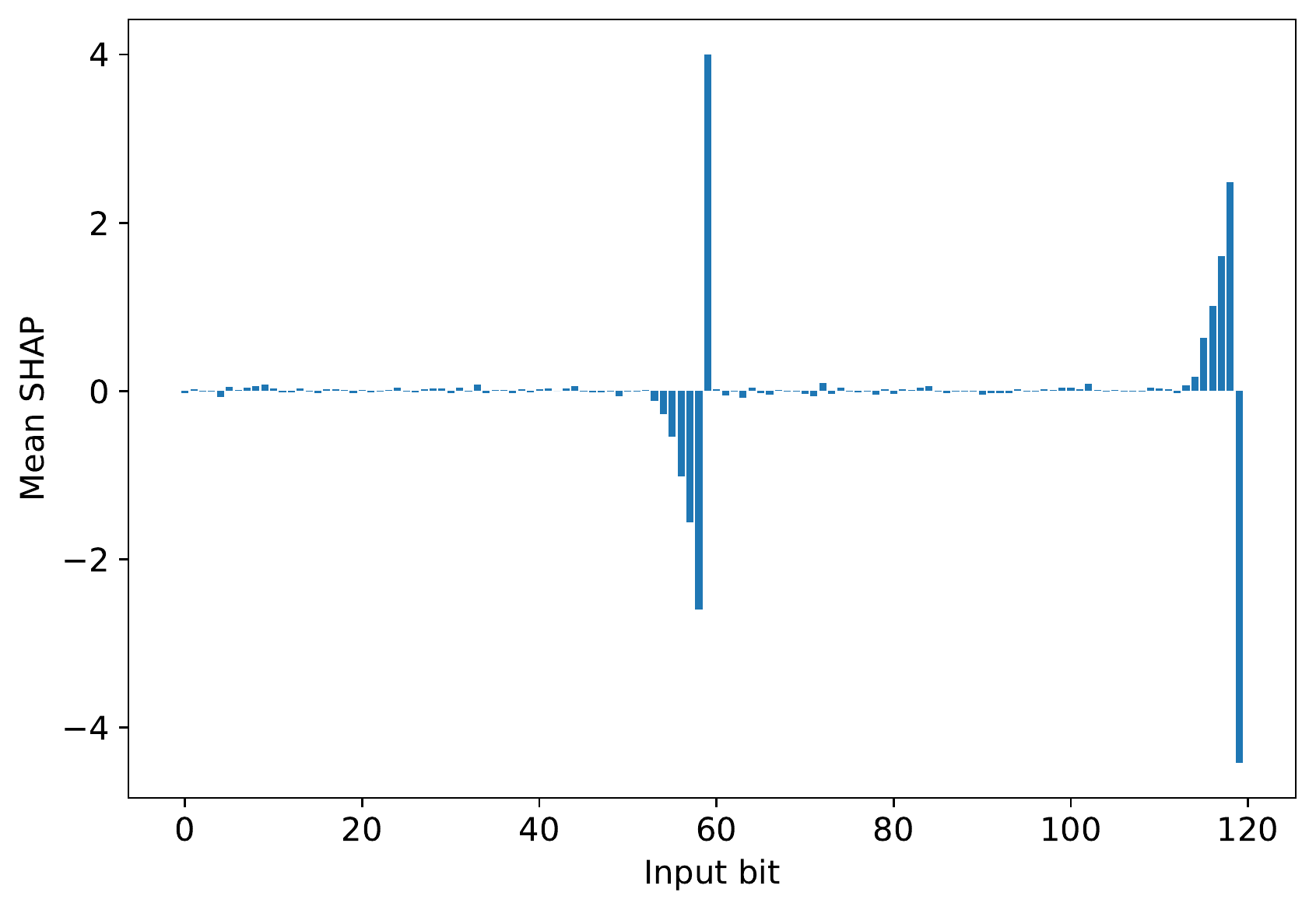}
    \caption{Mean SHAP values of input bits in ex35, comparator of two 60-bit signed integers, showing a pattern of ``weights'' that correspond to two signed integers with opposite polarities.}
    \label{fig:ex35}
\end{figure}

In the above discussion, it is assumed that the gate count does not exceed the limit of 5000. If it is not the case, the maximum depth of the decision tree and the trees in XGBoost, and/or the number of trees in XGBoost, can be reduced at the cost of potential loss of accuracy.

Tree-based models may not perform well in some pre-defined standard functions, especially symmetric functions and complex arithmetic functions. However, symmetric functions are easy to be identified by comparing the number of ones and the output bit. And it can be implemented by adding a side circuit that counts $N_1$, i.e., the number of ones in the input bits, and a decision tree that learns the relationship between $N_1$ and the original output. For arithmetic functions, patterns in the importance of input bits can be observed in some pre-defined standard functions, such as adders, comparators, outputs of XOR or MUX. This is done by training an initial XGBoost of trees and use SHAP tree explainer \cite{shap-tree} to evaluate the importance of each input bit.~\figurename{~\ref{fig:ex25}} shows that SHAP importance shows a pattern in training sets that suggests important input bits, while correlation coefficient fails to show a meaningful pattern.~\figurename{~\ref{fig:ex35}} shows an pattern of mean SHAP values of input bits, which suggests the possible existence of two signed binary coded integers with opposite polarities in the function. Based on these observations, Team 7 checks before ML if the training data come from a symmetric function, and compares training data with each identified pre-defined standard function. In case of a match, an AIG of the standard function is constructed directly without ML. With function matching, all six symmetric functions and 25 out of 40 arithmetic functions can be identified with close to 100\% accuracy.

%% file: Texfiles_from_collaborator/team8/main.tex
%

%
%

\subsection{Machine Learning Model Ensemble}
Model ensemble is a common technique in the machine learning community to improve the performance on a classification or regression task.
For this contest, we use a specific type of ensemble called ``bucket of models", where we train multiple machine learning models for each benchmark, generate the AIGs, and select the model that achieves the best accuracy on the validation set from all models whose AIGs are within the area constraint of 5k AND gates. 
The models we have explored in this contest include decision trees, random forests, and multi-layer perceptrons. 
The decision tree and multi-layer perceptron models are modified to provide better results on the benchmarks, and our enhancements to these models are introduced in Section~\ref{sec-dt} and Section~\ref{sec-mlp}, respectively. 
We perform a grid search to find the best combination of hyper-parameters for each benchmark. 

\subsection{Decision Tree for Logic Learning}
\label{sec-dt}
Using decision trees to learn Boolean functions can be dated back to the 1990s~\cite{oliveira1994learning}.
As a type of light-weight machine learning models, decision trees are very effective when the task's solution space has a ``nearest-neighbor" property and can be divided into 
regions with similar labels by cuts that are parallel to the feature axes. 
Many Boolean functions have these two properties, making them ideal targets for decision trees. 
In the rest of this section we introduce our binary decision tree implementation, and provide some insights on the connection between decision tree and Espresso~\cite{Rudell1987}, a successful logic minimization algorithm. 
\subsubsection{Binary Decision Tree}
\label{sec-dt-base}
We use our own implementation of the C4.5 decision tree~\cite{quinlan2014c4} for the IWLS'20 programming contest. 
The decision trees directly take the Boolean variables as inputs. 
We use mutual information as the impurity function: when evaluating the splitting criterion at each non-leaf node, the feature split that provides the maximum mutual information is selected. 
During training, our decision tree continues splitting at each node unless one of the following three conditions is satisfied:
\begin{enumerate}
    \item All samples at a node have the same label.
    \item All samples at a node have exactly the same features.
    \item The number of samples at a node is smaller than a hyper-parameter $N$ set by the user. 
\end{enumerate}
The hyper-parameter $N$ helps avoid overfitting by allowing the decision tree to tolerate errors in the training set. 
With larger $N$, the decision tree learns a simpler classification function, which may not achieve the best accuracy on the training set but generalizes better to unseen data. 

\begin{figure}[t]
\centering
\includegraphics[width=0.8\columnwidth]{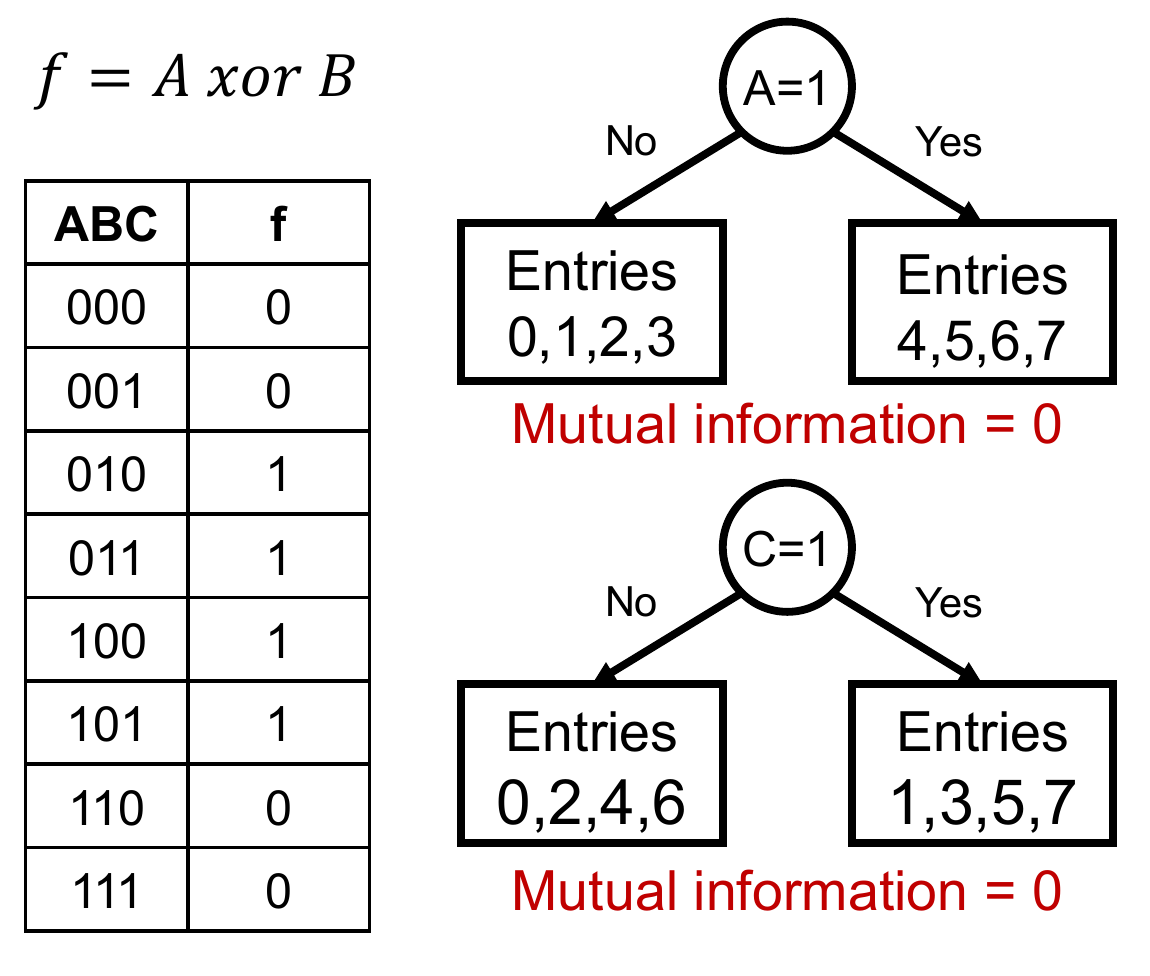}
\caption{Example of learning a two-input XOR with an irrelevant input. The whole truth table is provided for illustration purposes. 
The impurity function based on mutual information cannot distinguish the useful inputs ($A$ and $B$) from the irrelevant input $C$.
This is because mutual information only considers the probability distributions at the current node (root) and its children. }
\label{fig-fdecom-ex}
\end{figure}

\subsubsection{Functional Decomposition for Split Criterion Selection}
\label{sec-dt-decompose}
The splitting criterion selection method described in Section~\ref{sec-dt-base} is completely based on the probability distribution of data samples. 
One drawback of such probability-based metrics in logic learning is that they are very sensitive to the sampling noise, especially when the sampled data is only a very small portion of the complete truth table. 
~\figurename{~\ref{fig-fdecom-ex}} shows an example where we try to learn a three-input function whose output is the XOR of two inputs $A$ and $B$. 
An optimal decision tree for learning this function should split by either variable $A$ or $B$ at the root node. 
However, splitting by the correct variable $A$ or $B$ does not provide higher gain than splitting by the irrelevant variable $C$. 
Depending on the tie-breaking mechanism, the decision tree might choose to split by either of the three variables at the root node. 
For the contest benchmarks, only a small portion of the truth table is available.
As a result, it is possible that the sampling noise will cause a slight imbalance between the samples with zero and one labels. 
In such cases, it is likely that the decision tree will pick an irrelevant feature at the root node. 

If the decision tree selects an irrelevant or incorrect feature at the beginning, it is very difficult for it to "recover" during the rest of the training process due to the lack of a backtracking procedure. 
Finding the best decision tree for any arbitrary dataset requires exponential time. 
While recent works~\cite{hu2019optimal} propose to use smart branch-and-bound methods to find optimal sparse decision trees, it is still impractical to use such approaches for logic learning when the number of inputs is large. 
As a result, we propose to apply single-variable functional decomposition when the maximum mutual information is lower than a set threshold $\tau$ during splitting. 
The value of $\tau$ is a hyper-parameter and is included in the grid search for tuning. 
At each none-leaf node, we test all non-used features to find a feature that satisfies either of the following two requirements:
\begin{enumerate}
    \item At least one branch is constant after splitting by this feature.
    \item One branch is the complement of the other after splitting by this feature. 
\end{enumerate}
In many cases we don't have enough data samples to fully test the second requirement. 
As a result, we take an aggressive approach where we consider the second requirement as being satisfied unless we find a counter example. 
While such an aggressive approach may not find the best feature to split, it provides an opportunity to avoid the effect of sampling noise. 
In our implementation, functional decomposition is triggered more often at the beginning of the training and less often when close to the leaf nodes. 
This is because the number of samples at each non-leaf node decreases exponentially with respect to the depth of the node, and the sampling noise becomes more salient with few data samples. 
When the number of samples is small enough, the mutual information is unlikely to fall below $\tau$, so functional decomposition will no longer be triggered. 

We observed significant accuracy improvement after incorporating functional decomposition into the training process. 
However, we also realized that the improvement in some benchmarks might be because of an implementation detail: if multiple features satisfy the two requirements listed above, we select the last one. 
This implementation decision happens to help for some benchmarks, mostly because our check for the second requirement is too aggressive. 
This finding was also confirmed by team 1 after our discussion with them. 

\subsubsection{Connection with Espresso}
We found an interesting connection between decision tree and the Espresso logic optimization method. 
For the contest benchmarks, only 6.4k entries of the truth table are given and all other entries are don't cares. 
Espresso exploits don't cares by expanding the min-terms in the SOP, which also leverages the ``nearest neighbor" property in the Boolean input space. 
Decision trees also leverage this property, but they exploit don't cares by making cuts in the input space. 
As a result, neither Espresso nor decision tree is able to learn a logic function from an incomplete truth table if the nearest neighbor property does not hold. 
For example, according to our experiments, neither technique can correctly learn a parity function even if 90\% of the truth table has been given. 

\subsubsection{Converting into AIG}
One advantage of using decision trees for logic learning is that converting a decision tree into a logic circuit is trivial.
Each non-leaf node in the tree can be implemented as a two-input multiplexer, and the multiplexers can be easily connected by following the structure of the tree. 
The values at the leaf nodes are simply implemented as constant inputs to the multiplexers. 
We follow this procedure to convert each decision tree into an AIG, and use ABC~\cite{abc} to further optimize the AIG to reduce area. 
After optimization, none of the AIGs generated from decision trees exceeds the area budget of 5k AND gates. 

\subsection{Random Forest}
Random forest can be considered as an ensemble of bagged decision trees. 
Each random forest model contains multiple decision trees, where each tree is trained with a different subset of data samples or features so that the trees won't be very similar to each other. 
The final output of the model is the majority of all the decision trees' predictions. 

We used the random forest implementation from sci-kit learn~\cite{sklearn} and performed grid search to find the best combination of hyper-parameters. 
For all benchmarks we use a collection of seventeen trees with a maximum depth of eight. 
To generate the AIG, we first convert each tree into a separate AIG, and then connect all the AIGs together with a seventeen-input majority gate. 
The generated AIG is further optimized using ABC. None of the AIGs exceeds the 5K gate constraint after optimization. 

\subsection{Multi-Layer Perceptron}
\label{sec-mlp}
Our experiments find that multi-layer perceptrons (MLP) with a variety of activation functions perform well on certain benchmark classes. These models come with a few core challenges. First, it is difficult to translate a trained floating-point model to an AIG without significantly decreasing the accuracy. Thus, we limit the size of the MLP, synthesize it precisely through enumeration of all input-output pairs, and carefully reduce resultant truth table. Second, MLP architectures have difficulties with latent periodic features. To address this, we utilize the sine activation function which can select for periodic features in the input, validating previous work in its limited applicability to learning some classes of functions. 

\subsection{Sine Activation}
We constructed a sine-based MLP that performed as well or better than the equivalent ReLU-based MLP on certain benchmark classes. This work is concurrent with a recent work that applies periodic activations to implicit neural representation \cite{sitzmann2020sirens}, and draws from intermittent work over the last few decades on periodic activation functions \cite{lapedes1987neuralnets}\cite{sopena1999periodicneuralnets}. In our case, we find the sine activation function performs well in certain problems that have significant latent frequency components, e.g. parity circuit. It also performs well in other cases since for sufficiently small activations sine is monotonic, which passes gradients in a similar manner to the standard activation functions, e.g ReLU. We learn the period of the activation function implicitly through the weights inputted into the neuron. However, we find that one disadvantage is the exponential increase in local minima \cite{parascandolo2017taming}, which makes training more unstable.

\subsection{Logic Synthesis}
After training the (sine-based) MLP models, we convert them into AIGs. Some common methods for this conversion are binary neural networks \cite{liu2020reactnet}, quantized neural networks \cite{zhao2019ocs}, or directly learned logic \cite{wang2020lutnet}. Instead, we took advantage of the small input size and generated the full truth tables of the neural network by enumerating all inputs and recording their outputs. Then, we passed this truth table to ABC for minimization. This method is a simpler variant to one described in LogicNets \cite{umuroglu2020logicnets}. It easily produces models within the 5k gate limit for smaller models, which with our constraints corresponds to benchmarks with fewer than 20 inputs.

\subsection{Summary and Future Work}
We have described the ML models we used for the IWLS'20 contest. 
Based on our evaluation on the validation set, different models achieve strong results on different benchmarks. 
Decision trees achieve the best accuracy for simple arithmetic benchmarks (0-39), random forest gives better performance on the binary classification problems (80-99), 
while our special MLP dominates on the symmetric functions (70-79). 
It remains a challenge to learn complex arithmetic functions (40-49) from a small amount of data. 
Another direction for future work is to convert MLP with a large number of inputs into logic.

%% file: Texfiles_from_collaborator/team9/long.tex

%

\subsection{Introduction}
Team 9 uses a population-based optimization method called Cartesian Genetic Programming (CGP) for synthesizing circuits for solving the IWLS'20 contest problems. The flow used includes bootstrapping the initial population with solutions found by other well-performing methods as decision trees and espresso technique, and further improving them. 
The results of this flow are two-fold: (i) it allows to improve further the solutions found by the other techniques used for bootstrapping the evolutionary process, and 
(ii) alternatively, when no good solutions are provided by the bootstrapping methods, CGP starts the search from random (unbiased) individuals seeking optimal circuits.
 This text starts with a brief introduction to the CGP technique, followed by presenting the proposed flow in detail.

CGP is a stochastic search algorithm that uses an evolutionary approach for exploring the solution's parameter space, usually starting from a random population of candidate solutions. Through simple genetic operators such as selection, mutation and recombination, the search process often leads to incrementally improved solutions, despite there is no guarantee of optimality. 

The evolutionary methods seek to maximize a fitness function, which usually is user-defined and problem-related, that measures each candidate solution's performance, also called individual. The parameter space search is conducted by adding variation to the individuals by mutation or recombination and retaining the best performing individuals through the selection operator after evaluation. Mutation means that specific individual's parameters are randomly modified with a certain probability. Recombination means that individuals can be combined to generate new individuals, also with a given probability. This latter process is referred to as \textit{crossover}. However, the CGP variation does not employ the crossover process since it does not improve CGP efficiency \cite{millerStatusFuture}. The genetic operators are applied iteratively, and each iteration is usually called generation.

\begin{figure}[!b]
  \centering
  \includegraphics[width=1\linewidth]{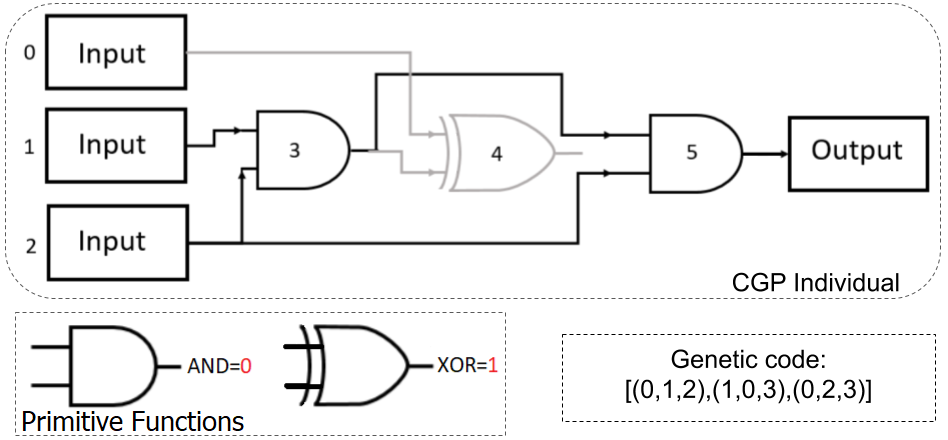}
    \caption{Single Line CGP Example.}
  \label{1line}
\end{figure}

~\figurename{~\ref{1line}} presents an example of a CGP implementation with one individual, its set of primitive logic functions, and the individual's genetic code.
The CGP structure is represented as a logic circuit, and the CGP Cartesian values are also shown. Notice that there is only one line present in the two-dimensional graph, in reference \cite{milano2018scaling} it is shown that such configuration provides faster convergence when executing the CGP search. Recent works also commonly make use of such configuration \cite{manazir2019recent}. The CGP implemented in work presented herein used such configuration of only one line and multiple columns. Please note that one may represent any logic circuit in a single line as one would in a complete two-dimensional graph.

\figurename{~\ref{1line}} also presents the individuals' representation, sometimes called genotype or genetic coding. This code is given by a vector of integers, where a 3-tuple describes each node, the first integer corresponds to the node's logic function, and the other two integers correspond to the fan-ins' Cartesian values. The number of primary inputs, primary outputs, and internal CGP nodes is fixed, i.e., are not under optimization. For the contest specifically, the PIs and POs sizes will always correspond to the input PLA file's values. At the same time, the number of internal CGP nodes are parametrized and set as one of the program's input information.

Notice that the circuit presented in~\figurename{~\ref{1line}} is composed of a functional and a non-functional portion, meaning that a specific section of the circuit does not influence its primary output. The non-functional portion of the circuit is drawn in gray, while the functional portion is black. The functional portion of a CGP individual is said to be its \textit{Phenotype}, in other words, the observable portion with relation to the circuit's primary outputs. At the same time, its complete genetic code representation is known as its \textit{Genotype}.

As demonstrated by \cite{milano2018scaling}, the CGP search has a better convergence if phenotypically larger solutions are considered preferred candidates when analyzing the individual's fitness scores. In other words, if, during the evolution, there are two individuals with equal accuracy but different functional sizes, the larger one will be preferred. The proposed flow makes use of such a technique by always choosing AIGs with larger available sizes as fathers when a tie in fitness values happens.

\begin{figure}[!hbt]
  \centering
  \includegraphics[width=0.78\linewidth]{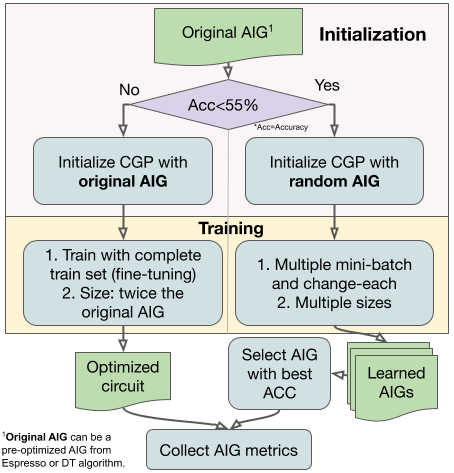}
    \caption{Proposed CGP Flow.}
  \label{flow}
\end{figure}

The proposed CGP flow is presented in~\figurename{~\ref{flow}}. It first tries to start from a well-performing AIG previously found using some other method as, for instance, Decision Trees(DT) or SOP via the Espresso technique. We call this initialization process bootstrap. It is noteworthy that the proposed flow can use any AIG as initialization, making it possible to explore previous AIGs from different syntheses and apply the flow as a fine-tuning process.

 
If the bootstrapped AIG has an accuracy inferior to 55\%, the CGP starts the search from a random (unbiased) population seeking optimal circuits; otherwise, it runs a bootstrapped initialization with the individual generated by previously optimized Sum of Products created by Decision Trees or Espresso. Summarizing, the flow's first step is to evaluate the input AIG provided by another method and check whether it will be used to run the CGP search, or the search will be conducted using a new random AIG.


Due to the CGP training approach, the learning set must be split among the CGP and the other method for using the Bootstrap option. When this is the case, the training set is split in a 40\%-40\%/20\% format, meaning half the training set is used by the initialization method and the other half by the CGP, while leaving 20\% of the dataset for testing with both methods involved in the AIG learning.

An empiric exploration of input hyper-parameters configurations was realized, attempting to look for superior solutions. 
The evolutionary approach used is the (1+4)-ES rule. In this approach, one individual generates four new mutated copies of itself, and the one with the best performance is selected for generating the mutated copies of the next generation. Variation is added to the individuals through mutations. The mutation rate is under optimization according to the $1/5^{th}$ rule \cite{doerr:15}, in which mutation rate value varies jointly with the proportion of individuals being better or worst than their ancestor. 

The CGP was run with three main hyper-parameter: (1) the number of generations with values of 10, 20, 25, 50 and 100 thousand; (2) two logic structures available: AIG and XAIG; and (3) the program had the option to check for all nodes as being a PO during every training generation. The latter is a computationally intensive strategy, but some exemplars demonstrated good results with it. The CGP flow runs exhaustively, combining these three main hyper-parameter options. 

Together with the previously mentioned hyper-parameters, there is a set of four other hyper-parameters dependent on the initialization option, presented in~\tablename{~\ref{tab:parameters}}. For a bootstrap option of a previous AIG, there is a single configuration where the CGP size is twice the Original AIG, meaning that for each node in the Original AIG, a non-functional node is incremented to the learning model, increasing the search space. The training configuration takes half the training set since the other half was already used by the initialization method. This bootstrap option does not execute with \textit{mini-batches} since they add stochasticity for the search and the intention is to fine-tune the AIG solution. 

For a random initialization, multiple hyper-parameters alternatives are explored in parallel. The AIG size is configured to 500 and 5000 nodes. The training set does not have to be shared, meaning that the whole training set is employed for the search. Training mini-batches of size 1024 are used, meaning that the individuals are evaluated using the same examples during a certain number of consecutive generations. The number of generations a mini-batch is kept is determined by the \textit{change each} hyper-parameter, meaning that the examples used for evaluating individuals change each time that number of generations is reached. A previous study showed that this technique could lead to the synthesis of more robust solutions, i. e., solutions that generalize well to unseen instances \cite{carvalho2018}. The random initialization could also use the complete train set as if the batch size was the same as the training set available.

Finally, the last step evaluates the AIGs generated, collecting results of accuracy, number of nodes, and logic levels for the validation sets. All the AIGs generated by the CGP respect the size limit of 5000 AND nodes. The AIG with the highest accuracy is chosen as the final solution for the benchmark. 
The CGP flow shows the potential to optimize previous solutions, where the optimization metric is easily configurable.


\begin{table}[]
\resizebox{1.0\columnwidth}{!}{
\begin{tabular}{ccccc}
\hline
\begin{tabular}[c]{@{}c@{}}Initialization\\ Type\end{tabular} & \begin{tabular}[c]{@{}c@{}}AIG\\ Size\end{tabular}                & \begin{tabular}[c]{@{}c@{}}Train/Test \\ Format\\ (\%)\end{tabular} & \begin{tabular}[c]{@{}c@{}}Batch \\ Size\end{tabular}                  & \begin{tabular}[c]{@{}c@{}}Change \\ Each\end{tabular}    \\ \hline
Bootstrap                                                     & \begin{tabular}[c]{@{}c@{}}Twice the \\ Original AIG\end{tabular} & 40-40/20                                                            & \begin{tabular}[c]{@{}c@{}}Half\\ Train Set\end{tabular}               & \begin{tabular}[c]{@{}c@{}}Not \\ Applicable\end{tabular} \\ \\
Random                                                        & \begin{tabular}[c]{@{}c@{}}500, \\ 5000\end{tabular}              & 80/20                                                               & \begin{tabular}[c]{@{}c@{}}1024, \\ Complete \\ Train Set\end{tabular} & \begin{tabular}[c]{@{}c@{}}1000, \\ 2000\end{tabular}     \\ \hline
\end{tabular}}
\caption{Hyper-Parameters Dependent on Initialization}
\label{tab:parameters}
\end{table}

Experiments were executed on the \textit{Emulab Utah} \cite{white2002integrated} research cluster,
the computers available operate under a Ubuntu 18.04LTS image. The CGP was implemented in C++ with object-oriented paradigm, codes were compiled on GCC 7.5.